\pdfoutput=1

\documentclass[11pt]{article}

\usepackage[preprint]{acl}

\usepackage{times}
\usepackage{latexsym}

\usepackage[T1]{fontenc}

\usepackage[utf8]{inputenc}

\usepackage{microtype}

\usepackage{inconsolata}

\usepackage{graphicx}
\usepackage{tikz}
\usepackage{tipa} 
\usepackage{xspace}
\usepackage{xcolor,pifont}
\usepackage{url}            
\usepackage{booktabs}       
\usepackage{amsfonts}       
\usepackage{nicefrac}       
\usepackage{amsmath}
\usepackage{amssymb}
\usepackage{tablefootnote}
\usepackage{enumitem}
\usepackage{caption}
\usepackage{subcaption}
\usepackage{pifont}
\usepackage{multirow}
\usepackage{makecell} 
\usepackage{ragged2e}
\usepackage{comment}
\usepackage{textgreek}
\usepackage{tipa}
\usepackage{lscape}
\usepackage{longtable}
\usepackage{array}
\usepackage{booktabs}   
\usepackage{listings}  
\usepackage{hyperref}
\usepackage{times}
\usepackage{latexsym}
\usepackage{xspace,mfirstuc,tabulary}
\usepackage{adjustbox}

\usepackage{color, colortbl}
\usepackage{xcolor}
\usepackage{enumitem}
\usepackage{amssymb} 
\usepackage{pifont} 
\usepackage{multirow}
\usepackage{longtable}
\usepackage[utf8]{inputenc}

\usepackage{siunitx}
\usepackage{textgreek}
\usepackage{placeins}

\usepackage[hang,flushmargin]{footmisc}

\usepackage{float}

\restylefloat{table}

\usepackage[T1]{fontenc}
\usepackage[utf8]{inputenc}

\definecolor{Gray}{gray}{0.9}
\definecolor{LightCyan}{rgb}{0.88,1,1}
\definecolor{deeppeach}{rgb}{1.0, 0.8, 0.64}

\newcommand*{\yoruba}{Yor\`ub\'a\xspace}
\newcommand*{\ghomala}{Ghom\'al\'a'\xspace}
\newcommand*{\ewe}{\'Ew\'e\xspace}

\newcommand*{\kabiye}{Kabiy\`e \xspace}

\newcommand*{\afrisenti}{\textsc{AfriSenti} \xspace}
\newcommand*{\nollysenti}{\textsc{NollySenti} \xspace}
\newcommand*{\masakhanews}{\textsc{MasakhaNEWS} \xspace}
\newcommand*{\injongo}{\textsc{Injongo-Intent} \xspace}
\newcommand*{\masakhaner}{\textsc{MasakhaNER} \xspace}
\newcommand*{\masakhanerx}{\textsc{MasakhaNER-X} \xspace}
\newcommand*{\masakhapos}{\textsc{MasakhaPOS} \xspace}
\newcommand*{\mafand}{\textsc{Mafand} \xspace}
\newcommand*{\afrimmlu}{\textsc{AfriMMLU} \xspace}
\newcommand*{\openaimmlu}{\textsc{OpenAI-MMLU} \xspace}
\newcommand*{\afrimgsm}{\textsc{AfriMGSM} \xspace}
\newcommand*{\afrixnli}{\textsc{AfriXNLI} \xspace}
\newcommand*{\uhura}{\textsc{Uhura} \xspace}
\newcommand*{\afriqa}{\textsc{AfriQA} \xspace}
\newcommand*{\belebele}{\textsc{Belebele} \xspace}
\newcommand*{\afrihate}{\textsc{AfriHate} \xspace}
\newcommand*{\naijarc}{\textsc{NaijaRC} \xspace}
\newcommand*{\sib}{\textsc{SIB-200} \xspace}
\newcommand*{\xlsum}{\textsc{XL-SUM} \xspace}
\newcommand*{\afrobench}{\textsc{AfroBench}}
\newcommand*{\afrolite}[0]{\textsc{AfroBench-Lite}}
\newcommand*{\afriadr}{\textsc{AfriADR} \xspace}
\newcommand*{\flores}{\textsc{Flores} \xspace}
\newcommand*{\ntrex}{\textsc{NTREX-128} \xspace}
\newcommand*{\salt}{\textsc{SALT} \xspace}
\newcommand*{\gpt}{\texttt{GPT-4o}}
\newcommand*{\gemini}{\texttt{Gemini-1.5 pro}}
\newcommand*{\gemma}{\texttt{Gemma 2 27B}}
\newcommand*{\gemmaS}{\texttt{Gemma 2 9B} }
\newcommand*{\gemmaO}{\texttt{Gemma 1.1 7B}}
\newcommand*{\llamaTwo}{\texttt{LLaMa 2 7B} \xspace}
\newcommand*{\llamaThree}{\texttt{LLaMa 3 8B} \xspace}
\newcommand*{\llamaThreeOne}{\texttt{LLaMa 3.1 8B \xspace}}
\newcommand*{\llamaL}{\texttt{LlaMa 3.1 70B \xspace}}
\newcommand*{\afrollama}{\texttt{AfroLlama 8B \xspace}}
\newcommand*{\aya}{\texttt{Aya-101 \xspace}}
\newcommand*{\llamaX}{\texttt{LLaMaX 8B \xspace}}

\newcommand{\cmark}{\ding{51}}%
\newcommand{\insertlanguagescovered}{
        \clearpage
        \onecolumn        

        \tabcolsep=3pt
        \begin{longtable}{l|lllcr}
       
        \toprule
            & \textbf{Language} & \textbf{Branch} & \textbf{Region (of Africa)} & \textbf{Script} & \textbf{\# speakers} \\
            \midrule
            \endhead
        
            \midrule
            \multicolumn{6}{r}{\textit{Continued on next page}} \\
            \endfoot
        
            \bottomrule
            \endlastfoot

            \multirow{12}{*}[-8pt]{\rotatebox{90}{Afro-Asiatic}} 
            & Algerian Arabic (arq)  & Semitic & North & Arabic  & 36M \\
            & Amharic (amh)  & Ethio-Semitic & East & Ge'ez  & 57M \\
            & Egyptian Arabic (arz)  & Semitic & North & Arabic  & 41M \\
            & Hausa (hau) & Chadic & West & Latin & 77M \\
            & Kabyle (kab)  & Berber & North & Arabic & 3M \\
            & Oromo (orm) & Cushitic & East & Latin & 37M \\            
            & Moroccan Arabic (ary)  & Semitic & North & Arabic  & 29M \\
           & Somali (som) & Cushitic & East & Latin  & 22M \\
            & Tamasheq (taq) & Berber & East & Latin  & 1M \\
            & Tamazight (tzm) & Berber & East & Latin  & - \\
            & Tigrinya (tig) & Ethio-Semitic & East & Ge'ez & 9M \\
            & Tunisian Arabic (aeb)  & Semitic & North & Arabic  & 12M \\
        \midrule
            \multirow{5}{*}[-4pt]{\rotatebox{90}{Niger-Congo}}
        \multirow{34}{*}[-4pt]{\rotatebox{90}{Niger-Congo}}
            & Akan (aka)  &  Tano & West  & Latin  & 10M \\
            & Bambara (bam)  & Mande & West & Latin  & 14M \\
            & Bemba (bem)  &  Bantu & South, East \& Central  & Latin  & 4M \\
            & Chichewa (nya) &  Bantu & South-East & Latin & 14M \\
            & chiShona (sna) &  Bantu & Southern  & Latin & 11M \\
            & Chokwe (cjk) &  Bantu & South \& Central & Latin & 1M \\
            & Dyula (dyu) & Mande & West & Latin  & 3M \\
            & \ewe (ewe)  &  Kwa & West & Latin  & 7M \\
            & Fon (fon)  &  Volta-Niger & West & Latin  & 14M \\
            & \ghomala (bbj) &  Grassfields & Central & Latin & 1M \\
             & Igbo (ibo) & Volta-Niger & West  & Latin  & 31M \\
            & isiXhosa (xho) &  Bantu & Southern & Latin & 19M \\
            & isiZulu (zul) &  Bantu & Southern & Latin & 27M \\
            & \kabiye (kbp) &  Gur & West & Latin & 1M \\
            & Kamba (kam) &  Bantu & East & Latin & 5M \\
            & Kikongo (kon) &  Bantu & South \& Central & Latin & 5M \\
            & Kikuyu (kik) &  Bantu & East & Latin & 8M \\
            & Kimbundu (kmb) &  Bantu & Southern & Latin & 2M \\
            & Kinyarwanda (kin) & Bantu & East  & Latin  & 10M \\
            & Kiswahili (swa) & Bantu & East \& Central & Latin & 71M-106M \\
            & Lingala (lin) &   Bantu & Central  & Latin  & 40M \\
            & Luba-Kasai (lua) &   Bantu & Central  & Latin  & 6M \\
            & Luganda (lug) &   Bantu & Central  & Latin  & 11M \\
            & Lugbara (lgg) \\
            & Mossi (mos) &  Gur & West & Latin & 8M \\
            & Nigerian Fulfulde (fuv) &  Senegambia & West & Latin & 15M \\
            & N'Ko (nqo) & Mande & West & Latin  & - \\
            & Northern Sotho (nso) &  Bantu & Southern & Latin & 4M \\
            & Rundi (run) &  Bantu & East & Latin & 11M \\
            & Runyankole (nyn) & \\
            & Sango (sag) &   Ubangian & Central  & Latin  & 5M \\
            & Setswana (tsn)  &  Bantu & Southern & Latin & 14M \\
            & Southern Sotho (sot) &  Bantu & Southern & Latin & 7M \\
            & Swati (ssw)  &  Bantu & Southern  & Latin  & 1M \\
             & Twi (twi)  & Kwa & West  & Latin  & 9M \\
            & Tumbuka (tum) &  Bantu & South \& East & Latin & 2M \\
            & Umbundu (umb)  &  Bantu & Southern  & Latin  & 7M \\
            & Xitsonga (tso) &  Bantu & Southern & Latin & 7M \\
            & Wolof (wol) &  Senegambia & West & Latin & 5M \\
            & Yoruba (yor) & Volta-Niger & West  & Latin & 46M \\
        \midrule
            \multirow{5}{*}[-4pt]{\rotatebox{90}{Nilo-Saharan}}
            & Acholi (ach) & Nilotic & East & Latin & 1.5M \\ 
            & Ateso (teo) & Nilotic  & East & Latin & 2.8M\\

            & Dinka (dik)  & Nilotic & Central  & Latin  & 4M \\
            & Kanuri (knc)  & Saharan  & West/Central  & Latin  & 10M \\
            & Kanuri (knc)  & Saharan  & West/Central  & Arabic  & 10M \\
            & Luo (luo)  & Nilotic  & East  & Latin  & 4M \\
            & Neur (nus)  & Nilotic  & Central  & Latin  & 2M \\
        \midrule
            \multirow{4}{*}[-4pt]{\rotatebox{90}{Austronesian}}
            &  &  &  &  &  \\  
            &  &  &  &  &  \\  
            & Malagasy (plt)  & Malayo-Polynesian & Southern & Latin  & 25M \\
            &  &  &  &  &  \\  
            &  &  &  &  &  \\  
            
        \midrule
            \multirow{5}{*}[-7pt]{\rotatebox{90}{Indo-European}}
            &  &  &  &  &  \\  
            &  &  &  &  &  \\  
            & Afrikaans (afr)  & Germanic & Southern & Latin  & 7M \\
            & Mozambican Portuguese (pt-MZ) & Italic & South East & Latin & 13M \\
            &  &  &  &  &  \\  
            &  &  &  &  &  \\  
        \midrule
            \multirow{3}{*}[-7pt]{\rotatebox{90}{Creoles}}
             &  &  &  &  &  \\  
            & Nigerian Pidgin (pcm) & English-based & West & Latin  & 121M \\
            & Kabuverdianu (kea) & Portuguese-based & West & Latin & 1M \\
             &  &  &  &  &  \\  
         \caption{\textbf{Languages covered in each of our evaluation tasks}: language family, region, script, number of  L1 \& L2 speakers}
         \label{tab:languages_covered}
        \end{longtable}

        \clearpage
}

\newcommand{\inserttaskscovered}{
    
    \begin{longtable}{l|ccccccccc|c|cccccc|cccccc|r}
        
        \toprule
        
        & \multicolumn{9}{c|}{Classification} & Reasoning & \multicolumn{6}{c|}{Question} & \multicolumn{6}{c|}{Generation} \\
        &&&&&&&&&&& \multicolumn{6}{c|}{Answering} &&&&&&& \\
        &&&&&&&&&&&&&&&&&&&&&&& \\
        \textbf{Lang.}
        & \rotatebox{90}{\afrihate}
        & \rotatebox{90}{\afrisenti}
        & \rotatebox{90}{\afrixnli}
        & \rotatebox{90}{\injongo}
        & \rotatebox{90}{\nollysenti}
        & \rotatebox{90}{\masakhanews}
        & \rotatebox{90}{\masakhaner}
        & \rotatebox{90}{\masakhapos}
        & \rotatebox{90}{\sib}
        & \rotatebox{90}{\afrimgsm~}
        & \rotatebox{90}{\afrimmlu}
        & \rotatebox{90}{\afriqa}
        & \rotatebox{90}{\belebele}
        & \rotatebox{90}{\naijarc}
        & \rotatebox{90}{\openaimmlu}
        & \rotatebox{90}{\uhura}
        & \rotatebox{90}{\afriadr}
        & \rotatebox{90}{\flores}
        & \rotatebox{90}{\mafand}
        & \rotatebox{90}{\ntrex}
        & \rotatebox{90}{\salt}
        & \rotatebox{90}{\xlsum}
        & \textbf{\# Tasks} \\
        \midrule
        \endhead
        
        \midrule
        \multicolumn{23}{r}{\textit{Continued on next page}} \\
        \endfoot
        
        \bottomrule
        \endlastfoot

        aeb &   & &&& & &&& \cmark & &   & &  &&&&& \cmark &&&&& 2 \\
        ach &   & &&& & &&&& &   & &  &&&&&&&& \cmark && 1 \\
        afr &   & &&& & &&&&  &   & &  \cmark &&&&& \cmark && \cmark &&& 3 \\
        aka &   & &&& & &&& \cmark & &   & &  &&&&& \cmark &&&&& 2 \\
        amh & \cmark  & \cmark & \cmark & \cmark & & \cmark & \cmark && \cmark & \cmark &   \cmark & &  \cmark &&& \cmark & & \cmark & \cmark & \cmark &&& 14 \\
        ara &   & &&& & &&&& &   & &  && \cmark && &&&&& \cmark & 2 \\
        arq & \cmark  & \cmark &&& & &&&& &   & &  &&&&&&&&&& 2 \\
        ary &  \cmark & \cmark &&& & &&& \cmark & &   & &  \cmark &&&&& \cmark &&&&& 5 \\
        arz &   & &&& & && & \cmark & &   & &  \cmark &&&&& \cmark &&&&& 3 \\
        bam &   & &&& & & \cmark & \cmark & \cmark & &   & &  \cmark &&&&& \cmark & \cmark &&&& 6 \\
        bbj &   & &&& & & \cmark & \cmark && &   & &  &&&& \cmark & & \cmark &&&& 4 \\
        bem &   & &&& & &&&\cmark & &   & \cmark &  &&&&& \cmark && \cmark &&& 4 \\
        cjk &   & &&& & &&& \cmark & &   & &  &&&&& \cmark &&&&& 2 \\
        dik &   & &&& & &&& \cmark & &   & &  &&&&& \cmark &&&&& 2 \\
        dyu &   & &&& & &&& \cmark & &   & &  &&&&& \cmark &&&&& 2 \\
        ewe &   & & \cmark & \cmark & & & \cmark & \cmark & \cmark & \cmark &   \cmark & &  &&&&& \cmark & \cmark & \cmark &&& 10 \\
        fon &   & &&& & && \cmark & \cmark & &   & \cmark &  &&&& \cmark & \cmark & \cmark &&&& 6 \\
        fuv &   & &&& & &&& \cmark & &   & &  &&&&&&&&&& 1 \\
        gaz &   & &&& & &&& \cmark & &   & &  &&&&& \cmark &&&&& 2 \\
        hau &  \cmark & \cmark & \cmark & \cmark & \cmark & \cmark & \cmark & \cmark & \cmark & \cmark &   \cmark & \cmark &  \cmark & \cmark && \cmark && \cmark & \cmark & \cmark & & \cmark & 19 \\
        ibo &  \cmark & \cmark & \cmark & \cmark & \cmark & \cmark & \cmark & \cmark & \cmark & \cmark &   \cmark & \cmark &  \cmark & \cmark &&& \cmark & \cmark & \cmark & \cmark & \cmark & \cmark & 19 \\
        kab &   & &&& & &&& \cmark & &   & &  &&&&& \cmark &&&&& 2 \\
        kam &   & &&& & &&& \cmark & &   & &  &&&&& \cmark &&&&& 2 \\
        kbp &   & &&& & &&& \cmark & &   & &  &&&&& \cmark &&&&& 2 \\
        kea &   & &&& & &&& \cmark & &   & &  &&&&& \cmark &&&&& 2 \\
        kik &   & &&& & &&& \cmark & &   & &  &&&&& \cmark &&&&& 2 \\
        kin & \cmark  & \cmark & \cmark & \cmark & & & \cmark & \cmark & \cmark & \cmark &   \cmark & \cmark &  &&&&& \cmark & \cmark & \cmark &&& 13 \\
        kmb &   & &&& & &&& \cmark & &   & &  &&&&& \cmark &&&&& 2 \\
        knc &   & &&& & &&& \cmark & &   & &  &&&&& \cmark &&&&& 2 \\
        kon &   & &&& & &&& \cmark & &   & &  &&&&& \cmark &&&&& 2 \\
        lgg &   & &&& & & &&& &   & &  &&&&&&&& \cmark && 1 \\
        lin &   & & \cmark & \cmark & & \cmark &&& \cmark & \cmark &   \cmark & & \cmark &&&&& \cmark &&&&& 8 \\
        lua &   & &&& & &&& \cmark & &   & &  &&&&& \cmark &&&&& 2 \\
        lug &   & & \cmark & \cmark & & \cmark & \cmark & \cmark & \cmark & \cmark &   \cmark & &  &&&&& \cmark & \cmark && \cmark && 11 \\
        luo &   & &&& & & \cmark & \cmark & \cmark & &   & &  &&&&& \cmark & \cmark &&&& 5 \\
        mos &   & &&& & & \cmark & \cmark & \cmark & &   & &  &&&&& \cmark & \cmark &&&& 5 \\
        nde &   & &&& & &&&& &   & &  &&&&&&& \cmark &&& 1 \\
        nso &   & &&& & &&& \cmark & &   & &  &&&&& \cmark && \cmark &&& 3 \\
        nus &   & &&& & &&& \cmark & &   & &  &&&&& \cmark &&&&& 2 \\
        nya &   & &&& & & \cmark & \cmark & \cmark & &   & &  &&&&& \cmark & \cmark & \cmark &&& 6 \\
        nyn &   & &&& & &&&& &   & &  &&&&&&&& \cmark && 1 \\
        orm & \cmark  & \cmark & \cmark & \cmark & & \cmark &&&& \cmark &   \cmark & &  &&&&&&& \cmark && \cmark & 9 \\
        pcm & \cmark & \cmark &&& & \cmark & \cmark & \cmark && &   & &  &&&&&& \cmark &&& \cmark & 7 \\
        plt &   & &&& & &&& \cmark & &   & &  &&&&& \cmark && \cmark &&& 3 \\
        run &   & &&& & \cmark &&& \cmark & &   & &  &&&&& \cmark &&&&& 3 \\
        sag &   & &&& & &&& \cmark & &   & &  &&&&& \cmark &&&&& 2 \\
        sna &   & & \cmark & \cmark & & \cmark & \cmark & \cmark & \cmark & \cmark &   \cmark & &  \cmark &&&&& \cmark & \cmark & \cmark &&& 12 \\
        som & \cmark  & &&& & \cmark &&& \cmark & &   & &  &&&&& \cmark && \cmark && \cmark & 6 \\
        sot &   & & \cmark & \cmark & & &&&& \cmark &   \cmark & &  &&&&& \cmark &&&&& 5 \\
        ssw &   & &&& & &&& \cmark & &   & &  &&&&& \cmark && \cmark &&& 3 \\
        swa & \cmark  & \cmark & \cmark & \cmark & & \cmark & \cmark & \cmark & \cmark & \cmark &   \cmark & &  \cmark && \cmark & \cmark && \cmark & \cmark & \cmark & \cmark & \cmark & 18 \\
        taq &   & &&& & &&&& &   & &  &&&&& \cmark &&&&& 1 \\
        teo &   & &&& & &&&& &   & &  &&&&&&&& \cmark && 1 \\
        tir & \cmark & \cmark &&& & \cmark &&& \cmark & &   & &   \cmark &&&&& \cmark & & \cmark && \cmark & 8 \\
        tsn &   & &&& & & \cmark & \cmark && &   & &  &&&&& \cmark & \cmark & \cmark &&& 5 \\
        tso &   & \cmark &&& & &&& \cmark & &   & &  \cmark &&&&&&&&&& 3 \\
        tum &   & &&& & &&& \cmark & &   & &  &&&&& \cmark &&&&& 2 \\
        twi &   & \cmark & \cmark & \cmark &  & & \cmark & \cmark & \cmark & \cmark &   \cmark & \cmark &  &&&&& \cmark & \cmark &&&& 11 \\
        tzm &   & &&& & &&& \cmark & &   & &  &&&&& \cmark &&&&& 2 \\
        umb &   & &&& & &&& \cmark & &   & &  &&&&& \cmark &&&&& 2 \\
        ven &   & &&& & &&&& &   & &  &&&&&&& \cmark &&& 1 \\
        wol &   & & \cmark & \cmark & & & \cmark & \cmark & \cmark & &  \cmark & & \cmark &&&& \cmark & \cmark & \cmark & \cmark & \cmark && 12 \\
        xho & \cmark & & \cmark & \cmark & & \cmark & \cmark && \cmark & \cmark &  \cmark  & & \cmark &&&&& \cmark & \cmark & \cmark &&& 13 \\
        yor & \cmark & \cmark & \cmark & \cmark & \cmark & \cmark & \cmark & \cmark & \cmark & \cmark & \cmark & \cmark & \cmark & \cmark & \cmark & \cmark & \cmark & \cmark & \cmark & \cmark &  & \cmark & 21 \\
        zul & \cmark &  & \cmark & \cmark &  &  & \cmark & \cmark & \cmark & \cmark & \cmark & \cmark & \cmark &&& \cmark & & \cmark & \cmark & \cmark &&& 14 \\
        \bottomrule
        \caption{\textbf{Languages covered in each of our evaluation tasks}: check marks (\cmark) indicate that a language is covered by the task in that column. While 13 languages are covered by $\ge$ 10 tasks, 44 languages are covered by $\le$ 5 tasks. \sib and \flores have the broadest coverage of African languages. In general, classification and generation tasks have better coverage of African languages than reasoning and question answering tasks.} 
    \end{longtable}
    \label{tab:tasks_by_languages}
    
    \clearpage
    \twocolumn
}
\newcommand{\insertafrobenchresults}{
    \begin{table*}[t]
    \begin{center}
    \footnotesize
    \setlength{\tabcolsep}{2.8pt}  
    \resizebox{\textwidth}{!}{%
    \begin{tabular}{lrrrrrrr|rr|rr|r|rrrr|cc}
    \toprule
     & \multicolumn{7}{c}{\textit{natural language understanding}}  & \multicolumn{2}{c}{\textit{QA}}  & \multicolumn{2}{c}{\textit{knowledge}}  & \textit{reasoning} & \multicolumn{4}{c}{\textit{text generation}} & &  \\
    \textbf{Tasks} & \textbf{POS} & \textbf{NER} & \textbf{SA} & \textbf{TC} & \textbf{Intent} & \textbf{Hate} & \textbf{NLI} & \textbf{XQA} & \textbf{RC} & \textbf{Arc-E} & \textbf{MMLU} & \textbf{Math} & \multicolumn{2}{c}{\textbf{MT}} & \textbf{Summ} & \textbf{ADR} & \textbf{ALL}  & \textbf{FT.} \\
    \textbf{Metrics} & \textbf{acc} & \textbf{F1} & \textbf{F1} & \textbf{acc} & \textbf{acc} & \textbf{F1} & \textbf{acc} & \textbf{F1} & \textbf{F1} & \textbf{acc} & \textbf{acc} & \textbf{EM} &\multicolumn{2}{c}{\textbf{ChrF}} & \textbf{BertScore} & \textbf{ChrF} &\textbf{AVG} & \textbf{AVG}   \\
    \midrule
    \multicolumn{7}{l}{\textit{Fine-tuned baselines}} & & & & & & & \textit{en/fr-xx} & \textit{xx-en/fr} &  \\
    AfroXLMR & \textbf{89.4} & \textbf{84.6} & \textbf{72.1} & 74.4 & \textbf{93.7} & \textbf{77.2} & 61.4 &  & &  & &  & &  & & & &  \\ 
    mT5/AfriTeVa V2 1B & &  & &  & &  & & 52.5 & N/A& N/A & N/A & N/A &  & & \textbf{72.3} &\textbf{79.4} & &\cellcolor{lime}{\textbf{70.4}} \\ 
    NLLB 3.3B &  &  & &  &  &  &  &  &  & &  & & \textbf{40.4} & \textbf{47.8} & &  & & \\  
    \midrule
    \multicolumn{7}{l}{\textit{Prompt-based baselines}} \\
    \multicolumn{7}{l}{\textit{open models}} \\
    Gemma 1.1 7B &38.6 &27.9 &43.3  &45.3 & 9.4 &24.3 &34.0 &17.4 &38.1 &32.2 &28.6  &4.6  &11.7 &9.7 & 49.1&50.8 & 29.1&29.7 \\ 
    LLaMa 2 7B  &27.9 &15.6 &42.3 &19.4 &  1.5& 21.9 &33.8 & 13.7 &24.3 &23.3 &25.6  &2.0 &10.5 &20.3 & 46.9&30.4 & 22.5 &22.2\\ 
    LLaMa 3 8B  &48.5 &22.7 &43.6 &37.0 &  2.1& 27.8 &35.4 &12.6 &27.6 &32.0 &27.4  &5.1 &15.9 & 27.7& 66.2&26.1  & 28.6 &28.6\\ 
    LLaMaX 8B &41.6 &0.0 &51.9 &49.8 &  5.6& 28.6 &40.8 &2.2 &29.7 &39.9 &28.3  &4.0 &22.7 &35.0 & 50.7&49.4 &30.0 &29.0  \\ 
    LLaMa 3.1 8B  &47.1 &11.5 &50.5 &46.7 &  6.0& 23.6 &36.6 &21.8 &39.5 &32.8 &31.4  &6.8 & 16.4 &28.5 & 43.7&25.9 & 29.3 & 28.1 \\ 
    AfroLLaMa 8B &0.0 &3.5 &43.4 &19.8 & 0.8& 18.4 & 35.9 & 21.8& 24.1 & 37.2 & 25.8  & $3.7$ &8.4 & 9.5& 50.8& 5.2  & 19.3 & 17.6\\  
    Gemma 2 9B &51.9 &40.3 &60.0 &56.0 &29.2 &29.9 &40.3 &45.9 &51.6 &53.4 &37.1  &18.7 &24.8 &29.1 & 66.1 &51.6 & 42.9 & 42.9\\ 
    Aya-101 13B & 0.0 & 0.0 & \underline{63.4} &70.3 & 42.4 & 31.0 & \underline{51.5} & \textbf{62.5} & \underline{60.7} & \underline{59.6} &30.9 & 4.4 & 23.4 & 37.9 & 52.4&50.4  & 40.1 & 37.7\\ 
    \rowcolor{LightCyan}
    Gemma 2 27B & \underline{55.1} & \underline{50.8} & \underline{63.4} &62.4 &33.0 & 45.5 & 42.8 & 50.5 &53.9 &56.3 &40.5  & \underline{27.0} &27.9 &32.9 & 66.4& 55.1 & 47.7 &48.3\\ 
    LlaMa 3.1 70B & 54.1 &14.4 &52.2 &57.7 &  34.0 & \underline{49.0} & 38.0 & 44 & 49.7& 54.9 & \underline{39.9} & 23.2 &25.1 &37.9 & 67.6& \underline{51.7} & 43.3 &42.6\\ 
    \midrule
    \multicolumn{7}{l}{\textit{proprietary models}} \\
    \rowcolor{LightCyan}
    Gemini 1.5 pro & 60.8& 41.8& 68.3& \textbf{76.7}& 74.3 &62.1 &62.0 & 40.5&  52.7& 84.8 &57.6  &\textbf{52.3} &37.6 &41.7 & 66.7&  \textbf{55.6} & 58.5 &58.9 \\
    \rowcolor{LightCyan}
    GPT-4o (Aug) & \underline{62.8} & 40.7 & 68.0& 74.8& 74.0 &63.5 &\textbf{64.3} & 43.4 &  \textbf{69.2}& \textbf{85.7}&\textbf{60.4}  &49.8 &35.1 & 40.7& 66.5& 54.9 & \textbf{59.6} &58.1\\ 
    \bottomrule
     \end{tabular}
    }
    \vspace{-3mm}
    \caption[\textbf{AfroBench Evaluation results on fine-tuned models and LLMs}]{
    \textbf{AfroBench Evaluation Results on Fine-Tuned Models and LLMs.} We cover 15 tasks, 22 datasets, and 64 African languages in the evaluation. The best closed and open LLMs are highlighted in \colorbox{LightCyan}{Cyan}. We \textbf{bolden} the best result per task in each column. We provide average on \textbf{ALL} tasks and on those with fine-tuned baselines (\textbf{FT})}
    \label{tab:afrobench_results}
      \end{center}
    \end{table*}
    
}

\newcommand{\insertdatabreakdown}{
    \begin{table}[h]
    \begin{center}
    \scalebox{0.9}{
        \begin{tabular}{lrrr}
        \toprule
            \multirow{2}{*}{\textbf{Dataset}} & \textbf{\# Sentences} & \multirow{2}{*}{\textbf{\# Languages}} \\
             & \textbf{Evaluated} & \\
            \midrule
            MASAKHANEWS & $6,025$ & $16$ \\
            \sib & $11,628$ & $56$ \\
            AFRISENTI & $34,321$ & $14$ \\
            MASAKHANER-X & $29,901$ & $20$ \\
            AFRIQA & $3,560$ & $9$ \\
            MAFAND-MT & 24,201 & $16$ \\
            \xlsum & $19,410$ & $10$ \\
        \bottomrule
        \end{tabular}
    }
    \vspace{-4mm}
    \caption{\textbf{Dataset Breakdown} We breakdown the total number of sentences we evaluated for each task and the number of languages covered. 
    }
    \label{tab:data_breakdown}
    \end{center}
    \end{table}
}

\newcommand{\insertafrobenchlite}{
    \begin{table}[t]
    \centering
    \scriptsize  
    \setlength{\tabcolsep}{2pt}
    \resizebox{\columnwidth}{!}{
    \begin{tabular}{clccccccc|c}
    \toprule
        & & & & & & & & \textbf{MT} &  \\
        \textbf{Model} & \textbf{Lang} & \textbf{Intent} & \textbf{TC}  & \textbf{NLI} & \textbf{RC}  & \textbf{MMLU}  & \textbf{Math}  & \textit{en/fr-xx} & \textbf{AVG} \\ \midrule
        \rowcolor{Gray}
       \multirow{2}{*}{{\begin{tabular}[c]{@{}c@{}} \cellcolor{white}  Gemma 1.1\\ \cellcolor{white} 
 7B\end{tabular}}} & \textit{eng} & 72.1 & 86.3 & 59.2 & 87.9 & 44.6 & 20.8 & 26.1 & 56.7 \\ 
        & \textit{africa} & 10.2 & 42.0 & 34.6 & 34.1 & 27.3 & 5.1 & 10.9 &23.5 \\
        \midrule
        \rowcolor{Gray}
        \cellcolor{white}\multirow{2}{*}{{\begin{tabular}[c]{@{}c@{}} \cellcolor{white}Gemma 2\\ \cellcolor{white}9B\end{tabular}}} & \textit{eng} & 36.3 & 82.5 & 70.7 & \textbf{93.7} & 69.8 & 68.8 & 67.9 & 70.0 \\ 
       & \textit{africa} & 27.8 & 64.0 & 40.9 & 49.3 & 36.1 & 21.7 & 37.2 & 39.6 \\ 
        \midrule
        \rowcolor{Gray}
      \cellcolor{white} \multirow{2}{*}{{\begin{tabular}[c]{@{}c@{}} \cellcolor{white} Aya-101\\ \cellcolor{white} 13B\end{tabular}}} &  \textit{eng} & 78.0 & 82.8 & 67.0 & 86.1 & 42.8 & 11.6 & 64.2 & 61.8 \\ 
        & \textit{africa} & 40.2 & 76.0 & 52.4 & 59.7 & 30.3 & 4.9 & 31.8 & 42.2 \\ 
        \midrule
        \rowcolor{Gray}
       \cellcolor{white} \multirow{2}{*}{{\begin{tabular}[c]{@{}c@{}}\cellcolor{white} Gemma 2\\\cellcolor{white} 27B\end{tabular}}} &  \textit{eng} & 84.0 & \textbf{89.3} & 67.8 & 93.4 & 75.6 & 85.6 & 68.5 & 80.6\\ 
        & \textit{africa} & 31.4& 66.6 & 43.7 & 52.1 & 40.8 & 30.6 & 39.1 & 43.5\\ 
        \midrule
        \rowcolor{Gray}
        \cellcolor{white}\multirow{2}{*}{{\begin{tabular}[c]{@{}c@{}} \cellcolor{white} LLaMa 3.1 \\ \cellcolor{white} 70B\end{tabular}}} &  \textit{eng} & 84.5 & 88.3 & 59.5 & 93.2 & 76.4 & 86.8 & \textbf{71.6} & 80.0 \\ 
        & \textit{africa} & 36.9 & 61.9 & 38.4 & 45.3 & 40.6 & 26.5 & 29.6 & 39.9 \\ 
        \midrule
        \rowcolor{Gray}
        \cellcolor{white}\multirow{2}{*}{{\begin{tabular}[c]{@{}c@{}} \cellcolor{white} Gemini 1.5 \\ \cellcolor{white}pro\end{tabular}}} & \textit{eng} & \textbf{86.8} & 88.7 & 88.5 & 69.6 & \textbf{88.8} & 86.8 & 69.1 & 82.6 \\ 
        & \textit{africa} & 75.6 & \underline{81.3} & 63.6 & 54.4 & \underline{62.6} & \underline{57.7} & \underline{44.2} & 62.8\\ 
        \midrule
        \rowcolor{Gray}
        \cellcolor{white}\multirow{2}{*}{{\begin{tabular}[c]{@{}c@{}}\cellcolor{white} GPT-4o \\ \cellcolor{white}(Aug)\end{tabular}}} & \textit{eng} & 86.2 & 89.2 & \textbf{89.2} & 84.3 & 88.0 & \textbf{88.8} & 70.2 &  \textbf{85.1} \\ 
        & \textit{africa} & 78.4 & 83.0 & 66.3 & 70.3 & 63.1 & 57.3 & 43.6& \underline{66.0} \\ 
        \bottomrule
    \end{tabular}
    }
    \vspace{-2mm}
    \caption{\textbf{AfroBench-Lite Evaluation}: LLM baselines on 7 datasets spanning 14 African languages. Tasks were selected for broad NLP coverage, prioritizing language consistency. The best score per task is in \textbf{bold}.}
    \label{tab:afrobench-lite}
\end{table}
}

\newcommand{\insertafrobenchlitenew}{
    \begin{table*}[t]
    \centering
    \scriptsize  
    \begin{tabular}{clccccccc|c}
    \toprule
        & & & & & & & & \textbf{MT} &  \\
        \textbf{Model} & \textbf{Lang} & \textbf{Intent} & \textbf{TC}  & \textbf{NLI} & \textbf{RC}  & \textbf{MMLU}  & \textbf{Math}  & \textit{en/fr-xx} & \textbf{AVG} \\ \midrule
        \rowcolor{Gray}
        \cellcolor{white}\multirow{2}{*}{{\begin{tabular}[c]{@{}c@{}} \cellcolor{white} Lugha-Llama \\ \cellcolor{white} 8B\end{tabular}}} &  \textit{eng} & 16.7 & 43.6 & 46.8 & 22.4 & 31.8 & 6.4 & 51.3 & 31.3 \\ 
        & \textit{africa} & 4.1 & 34.1 & 36.7 & 23.0 & 25.2 & 1.8 & 22.1 & 21.0 \\ 
        \midrule
        
        \rowcolor{Gray}
       \multirow{2}{*}{{\begin{tabular}[c]{@{}c@{}} \cellcolor{white}  Gemma 1.1\\ \cellcolor{white} 
 7B\end{tabular}}} & \textit{eng} & 72.1 & 86.3 & 59.2 & 87.9 & 44.6 & 20.8 & 26.1 & 56.7 \\ 
        & \textit{africa} & 10.2 & 42.0 & 34.6 & 34.1 & 27.3 & 5.1 & 10.9 &23.5 \\
        \midrule
        
        \rowcolor{Gray}
        \cellcolor{white}\multirow{2}{*}{{\begin{tabular}[c]{@{}c@{}} \cellcolor{white}Gemma 2\\ \cellcolor{white}9B\end{tabular}}} & \textit{eng} & 36.3 & 82.5 & 70.7 & \underline{93.7} & 69.8 & 68.8 & 67.9 & 70.0 \\ 
       & \textit{africa} & 27.8 & 64.0 & 40.9 & 49.3 & 36.1 & 21.7 & 37.2 & 39.6 \\ 
        \midrule
        
        \rowcolor{Gray}
        \cellcolor{white}\multirow{2}{*}{{\begin{tabular}[c]{@{}c@{}} \cellcolor{white} LLaMa 3.1 \\ \cellcolor{white} 70B\end{tabular}}} &  \textit{eng} & 84.5 & 88.3 & 59.5 & 93.2 & 76.4 & 86.8 & 71.6 & 80.0 \\ 
        & \textit{africa} & 36.9 & 61.9 & 38.4 & 45.3 & 40.6 & 26.5 & 29.6 & 39.9 \\ 
        \midrule
        
        \rowcolor{Gray}
      \cellcolor{white} \multirow{2}{*}{{\begin{tabular}[c]{@{}c@{}} \cellcolor{white} Aya-101\\ \cellcolor{white} 13B\end{tabular}}} &  \textit{eng} & 78.0 & 82.8 & 67.0 & 86.1 & 42.8 & 11.6 & 64.2 & 61.8 \\ 
        & \textit{africa} & 40.2 & 76.0 & 52.4 & 59.7 & 30.3 & 4.9 & 31.8 & 42.2 \\
        \midrule
        
        \rowcolor{Gray}
       \cellcolor{white} \multirow{2}{*}{{\begin{tabular}[c]{@{}c@{}}\cellcolor{white} Gemma 2\\\cellcolor{white} 27B\end{tabular}}} &  \textit{eng} & 84.0 & 89.3 & 67.8 & 93.4 & 75.6 & 85.6 & 68.5 & 80.6\\ 
        & \textit{africa} & 31.4& 66.6 & 43.7 & 52.1 & 40.8 & 30.6 & 39.1 & 43.5\\ 
        \midrule

        \rowcolor{Gray}
        \cellcolor{white}\multirow{2}{*}{{\begin{tabular}[c]{@{}c@{}} \cellcolor{white} LLaMa 4 \\ \cellcolor{white} 405B\end{tabular}}} &  \textit{eng} & \underline{88.9} & 84.8 & 49.2 & 25 & 11.2 & 97.6 & 73 & 61.4 \\ 
        & \textit{africa} & 73.9 & 80.6 & 45.5 & 24.6 & 15.8 & 65.0 & 42.8 & 49.7 \\ 
        \midrule
        
        \rowcolor{Gray}
       \cellcolor{white} \multirow{2}{*}{{\begin{tabular}[c]{@{}c@{}}\cellcolor{white} Gemma 3\\\cellcolor{white} 27B\end{tabular}}} &  \textit{eng} & 79.6 & 87.3 & 65.5 & 93.4 & 74.2 & 87.6 & 68.9 & 79.5\\ 
        & \textit{africa} & 55.2 & 74.2 & 51.2  & 62.4 & 44.4 & 47.5 & 33.1 & 52.6\\ 
        \midrule
        
        \rowcolor{Gray}
        \cellcolor{white}\multirow{2}{*}{{\begin{tabular}[c]{@{}c@{}} \cellcolor{white} Gemini 1.5 \\ \cellcolor{white}pro\end{tabular}}} & \textit{eng} & 86.8 & 88.7 & 88.5 & 69.6 & \underline{88.8} & 86.8 & 69.1 & 82.6 \\ 
        & \textit{africa} & 75.6 & 81.3 & 63.6 & 54.4 & 62.6 & 57.7 & 44.2 & 62.8\\ 
        \midrule
        
        \rowcolor{Gray}
        \cellcolor{white}\multirow{2}{*}{{\begin{tabular}[c]{@{}c@{}}\cellcolor{white} GPT-4o \\ \cellcolor{white}(Aug)\end{tabular}}} & \textit{eng} & 86.2 & \underline{89.2} & 89.2 & 84.3 & 88.0 & 88.8 & 70.2 & \underline{85.1} \\ 
        & \textit{africa} & 78.4 & 83.0 & 66.3 & \textbf{70.3} & \textbf{63.1} & 57.3 & 43.6& 66.0 \\ 
        \midrule

        \rowcolor{Gray}
        \cellcolor{white}\multirow{2}{*}{{\begin{tabular}[c]{@{}c@{}} \cellcolor{white} Gemini 2.0\\ \cellcolor{white}Flash\end{tabular}}} & \textit{eng} & 87.6 & 86.8 & 87 & 63 & 80.8 & 92.8 & \underline{73.1} & 79.7 \\ 
        & \textit{africa} & 82.5 & \textbf{84.9} & 66.5 & 56.8 & 57.8 & \textbf{67.5} & \textbf{49.6} & 66.5\\
        \midrule
        
        \rowcolor{Gray}
        \cellcolor{white}\multirow{2}{*}{{\begin{tabular}[c]{@{}c@{}}\cellcolor{white} GPT-4.1 \\ \cellcolor{white}(April)\end{tabular}}} & \textit{eng} & 87.8 & \underline{89.7} & 88.5 & 73.9 & 71.4 & 82.4 & \underline{73.1} &  81.0 \\ 
        & \textit{africa} & \textbf{84.4} & 84.8 & \textbf{67.5} & 64.8 & 60.2 & 59.9 & 47.3 & \textbf{67.0} \\ 
        \bottomrule
    \end{tabular}

    \caption{\textbf{AfroBench-Lite Evaluation (NEW)}: LLM baselines on 7 datasets spanning 14 African languages (sorted by performance on African languages). Tasks were selected for broad NLP coverage, prioritizing language consistency. The best score per task is in \textbf{bold}.}
    \label{tab:afrobench-lite-new}
\end{table*}
}

\newcommand{\insertfewshot}{
    \begin{table*}[t]
    \centering
    \resizebox{\textwidth}{!}{
    \begin{tabular}{cl|ccccccccccccccc|c}
    \toprule
        & & & & & & & && && & & \textbf{MT} & \textbf{MT} & & \\
        \textbf{Tasks} & \textbf{\# shots}& \textbf{POS} & \textbf{NER} & \textbf{SA} & \textbf{TC} & \textbf{Intent} & \textbf{Hate} & \textbf{NLI} & \textbf{XQA} &\textbf{RC} & \textbf{MMLU} & \textbf{Math} & \textit{en/fr-xx} & \textit{xx-en/fr} & \textbf{SUMM} & \textbf{ADR} & \textbf{AVG} \\ \midrule
        \multirow{2}{*}{{\begin{tabular}[c]{@{}c@{}}Gemma 2  27B\\  \\ \end{tabular}}}
        & 0-shot & \underline{55.1} & \textbf{\underline{50.8}} & 58.6 & 57.3 & 35.2 &45.5 & 42.8 & 50.5 &53.6 & 39.9 & \underline{27.0} & \underline{32.4} & 32.4 & \underline{66.4} & \underline{55.1} & \underline{46.8}  \\
        \rowcolor{Gray}
        \cellcolor{white}& 5-shot & 43.9 & 14.5 & \underline{59.7} & \underline{62.5} & \underline{56.7} & \underline{57.3} & \underline{56.0} & \textbf{52.4} & \underline{58.3} & \underline{44.8} & 27.5 & 22.7 & \underline{34.9} & 55.5& 31.2 & 45.2\\ \hline
        \multirow{2}{*}{{\begin{tabular}[c]{@{}c@{}}Gemini 1.5 pro \\ pro \\ \end{tabular}}}
        & 0-shot & \underline{60.8} & \underline{41.8} & 62.6 & 74.5 & \textbf{74.3}& 62.1 & 62.0 & \underline{40.5} & \underline{53.0} & \textbf{60.2} &52.3  & 35.4 & 41.7 & 66.7 & 55.6 & \underline{56.2}\\
        \rowcolor{Gray}
         \cellcolor{white}& 5-shot & 33.2 & 37.4 & \textbf{64.5} & \textbf{77.3} &73.4 & \underline{64.1} & \underline{35.9} &28.7& 24.4 & 46.0 & \underline{49.0} & \underline{37.4} & \underline{43.1} & \textbf{70.4} & \textbf{63.4} & 49.9 \\ \hline
         \multirow{2}{*}{{\begin{tabular}[c]{@{}c@{}}GPT-4o (Aug) \\ (Aug) \\ \end{tabular}}}
        & 0-shot & \textbf{\underline{62.8}} & 40.7 & \underline{62.6} & 72.5 &\underline{74.0} & 63.5 & \textbf{64.3} & \underline{43.4} & 69.1 & \underline{60.0} & 49.8 & 31.5 & 41.0 & 66.5 & 54.9 & 57.1 \\
        \rowcolor{Gray}
          \cellcolor{white}& 5-shot& 62.4 & \underline{45.0} & 62.3 & \underline{72.9} &71.6 & \textbf{69.3} & 64.2 & 40.0&\textbf{71.9} & 59.7 &  \textbf{54.7} & \textbf{33.9} & \textbf{43.3} & \underline{67.9} & \underline{62.7} & \textbf{58.8}\\ \bottomrule
    \end{tabular}
    }
        \vspace{-2mm}
    \caption{\textbf{Few-shot Evaluation. } 
    The better score between each model's 0-shot and few-shot is in \underline{underlined}.} 
    \label{tab:few-shot-results}
\end{table*}
\vspace{-3mm}
}
\newcommand{\insertrelatedworks}{
    \begin{table*}[h!]
    \small\centering
    \resizebox{\textwidth}{!}{%
      \begin{tabular}{lrrrrll}
      \toprule
      \textbf{Benchmark} & \textbf{\# Tasks} & \textbf{\# Datasets}  & \textbf{\# African Lang.} & \textbf{\# LLMs} & \textbf{Closed LLMs evaluated} & \textbf{Dominant task(s)} \\
      \midrule
      ChatGPT-MT~\cite{robinson-etal-2023-chatgpt} & 1 & 1 & 57 & 1 & GPT-3.5 & MT \\
      Mega~\cite{ahuja-etal-2023-mega} & 10 & 16 & 11 & 4 & GPT-3, GPT-3.5-Turbo, GPT-4  & POS, NER \\
      Megaverse~\cite{ahuja-etal-2024-megaverse} & 16 & 22 & 16 & 8 & PaLM, GPT-3.5, GPT-4, Gemini Pro & POS, NER, XQA \\
      SIB-200~\cite{adelani-etal-2024-sib} & 1 & 1 & 57 & 2 & GPT-3.5, GPT-4 & Topic classification \\
      Belebele~\cite{bandarkar-etal-2024-belebele} & 1 & 1 & 28 & 6 & GPT-3.5-Turbo & QA \\
      Uhura~\cite{Bayes2024UhuraAB} & 1 & 2 & 6 & 6 & Claude-3.5-Sonnet, GPT-4, 4o, o1-preview & QA \\
      IrokoBench~\cite{Adelani2024IrokoBenchAN} & 3 & 3 & 16 & 16 & GPT-3.5,4,4o, Gemini-1.5-Pro, Claude OPUS & NLI, MMLU, Math. \\
    
      \rowcolor{Gray}
      \afrobench (Ours) & 15 & 22 & 60 & 12 & Gemini-1.5-Pro, GPT-4o & several \\
      \bottomrule
      \end{tabular}
    }
      \vspace{-3mm}
      \caption{\textbf{Overview of Related works that evaluated  on African languages}. We included the number of tasks, datasets, African languages, LLMs evaluated, and the dominant tasks covering at least three African languages. }
      \label{tab:works}
    \end{table*}
}
\newcommand{\insertqualitative}{
\begin{table}[t]
 \resizebox{\columnwidth}{!}{
\centering
\begin{tabular}{lcp{83mm}}
    \toprule
    \textbf{Lang.} & \textbf{Shot} & \textbf{Input / Output} \\
    \midrule
    \multicolumn{3}{l}{\textbf{\texttt{ADR}: }\gemini} \\ 
    \multirow{3}{*}{\texttt{bbj}} & \textbf{Input:} & A jw\textschwa \ gu\textipa{N} ts\textschwa \ aw$\varepsilon$ a l\textschwa \ nə\textipa{N} kwit\textschwa \ \\
    & \textbf{Target:} &  \^{A} jw\'{ə} gu\textipa{N} ts\'{ə} \ aw{$\acute{\varepsilon}$} a l\textschwa \ n\'{ə}\textipa{N} \ kw\'{i}t\'{ə}  \\
    &  \textbf{0-shot} & \textcolor{red}{\`a} jw\textcolor{red}{$\check{\varepsilon}$} g\textcolor{red}{\^{u}}\textipa{N} ts\textcolor{red}{\textschwa} \ \textcolor{red}{\'{a}}w{$\acute{\varepsilon}$} \textcolor{red}{\`a} l\textschwa \ n\textcolor{red}{\^{ə}}\textipa{N} \ \textcolor{red}{k\"{w}it\textschwa} (ChrF: 21.4) \\  
    &  \textbf{5-shot} & \textcolor{red}{a} jw\'{ə} gu\textipa{N} ts\'{ə} \ aw{$\acute{\varepsilon}$} a l\textschwa \ n\'{ə}\textipa{N} \ kw\textcolor{red}{i}t\textcolor{red}{ə} (ChrF: 81.6) \\
    \midrule
    \multicolumn{3}{l}{\textbf{\texttt{Hate speech}: } \gemini} \\ 
    \multirow{3}{*}{\texttt{yor}} & \textbf{Input:} & \textcolor{red}{oloriburuku} ni pupo ninu awon \textcolor{blue}{igbo} yii \\
    & \textbf{English:} & these \textcolor{blue}{igbo} people are  \textcolor{red}{brainless}  \\
    & \textbf{Label:} & hate \\
    &\textbf{0-shot} & abuse \\
     &\textbf{5-shot} & hate \\
    \midrule
   \multicolumn{3}{l}{\textbf{\texttt{Math reasoning}: }\gpt} \\ 
    \multirow{3}{*}{\texttt{yor}} & \textbf{Input:} & Ryan gbin \`{o}d\`{o}d\'{o} 2 n\'{i} oj\'{u}m{\d{\'o}} s\'{i} in\'{u} \d{o}gba\`{a} r\d{\`{e}}. L\d{\'{e}}y\`{i}n \d{o}j\d{\'{o}} 15, \`{o}d\`{o}d\'{o} m\'{e}l\`{o}\'{o} n\'{i} \'{o} n\'{i} t\'{i} 5 \`{o} b\'{a} w\`{u}?\\
    & \textbf{English:} & Ryan plants 2 flowers a day in his garden. After 15 days, how many flowers does he have if 5 did not grow? \\
    & \textbf{Answer:} & 25 \\
    &\textbf{0-shot} & ryan n\'{i} \`{o}d\`{o}d\'{o} \textcolor{red}{30} t\'{i} \'{o} b\'{a} \'{n} gbin 2 n\'{i} oj\'{u}m{\d{\'o}} \\
     &\textbf{8-shot} & \`{i}d\'{a}h\`{u}n: ryan gbin \`{o}d\`{o}d\'{o} 2 n\'{i} oj\'{u}m{\d{\'o}}. l{\d{\'e}}{y}\`{i}n {\d{o}}j{\d{\'o}} 15, \'{o} m\'{a}a gbin \`{o}d\`{o}d\'{o} 2 * 15 = 30. t\'{i} 5 \`{o} b\'{a} w\`{u}, \'{o} n\'{i} \`{o}d\`{o}d\'{o} 30 - 5 = 25. \`{i}d\'{a}h\`{u}n n\'{a}\`{a} ni 25. \\
    \bottomrule
    
\end{tabular}
}
\vspace{-3mm}
    \caption{\textbf{Qualitative Analysis} comparison of the 0-shot and 5-shot samples on ADR, Hate speech and Math. }
    \label{tab:qualitative-analysis}
\end{table}
}
\newcommand{\insertbestprompts}{
    \begin{table*}[t]
    \centering
    \resizebox{\textwidth}{!}{
    \begin{tabular}{cl|cccccccccccc}
    \toprule
    \textbf{Task} & \textbf{Dataset} & \textbf{AfroLLaMa 8B} & \textbf{LLaMAX3 8B} & \textbf{LLaMa2 7b} & \textbf{LLaMa3 8B} & \textbf{LLaMa3.1 8B} & \textbf{LLaMa3.1 70B} & \textbf{Aya-101 13B} & \textbf{Gemma1.1 7b} & \textbf{Gemma2 9b} & \textbf{Gemma2 27b} & \textbf{Gemini 1.5 Pro} & \textbf{GPT-4o (Aug)} \\
    \midrule
    \multirow{2}{*}{{\begin{tabular}[c]{@{}c@{}}\textbf{SA} \\ \end{tabular}}}
    & AfriSenti & T4 & T3 & T5 & T5 & T5 & T5 & T3 & T5 & T5 & T5 & T3 & T2 \\
     & NollySenti & T5 & T3 & T4 & T3 & T4 & T4 & T5 & T4 & T4 & T4 & T1 & T3 \\
    \midrule
    \multirow{2}{*}{{\begin{tabular}[c]{@{}c@{}}\textbf{TC} \\ \end{tabular}}}
    & Masakhanews & T3 & T3 & T4 & T3 & T3 & T3 & T2 & T2 & T3 & T3 & T2 & T2 \\
    & SIB & T3 & T3 & T5 & T2 & T2 & T3 & T4 & T4 & T5 & T3 & T5 & T3 \\
    \midrule
    \multirow{2}{*}{{\begin{tabular}[c]{@{}c@{}}\textbf{TokC} \\ \end{tabular}}}
    & MasakhaNER & T4 & T1 & T5 & T3 & T3 & T5 & T1 & T5 & T2 & T1 & T3 & T3 \\
     & MasakhaPOS & T1 & T5 & T1 & T2 & T2 & T2 & T1 & T2 & T2 & T3 & T3 & T3 \\
    \midrule
    \textbf{Intent} & InjongoIntent & T1 & T5 & T4 & T5 & T3 & T4 & T5 & T5 & T5 & T5 & T5 & T4 \\
    \midrule
    \textbf{Hate} & AfriHATE & T5 & T4 & T3 & T1 & T4 & T4 & T1 & T4 & T1 & T4 & T1 & T4 \\
    \midrule
    \textbf{NLI} & AfriXNLI & T2 & T1 & T3 & T2 & T2 & T2 & T4 & T2 & T2 & T2 & T3 & T3 \\
    \midrule
    \textbf{XQA} & AfriQA & T5 & T4 & T5 & T5 & T5 & T2 & T2 & T2 & T2 & T2 & T2 & T2 \\
    \midrule
    \multirow{2}{*}{{\begin{tabular}[c]{@{}c@{}}\textbf{RC} \\ \end{tabular}}}
    & NaijaRC & T4 & T5 & T4 & T1 & T5 & T5 & T4 & T3 & T4 & T5 & T3 & T2 \\
    & Belebele & T2 & T5 & T4 & T1 & T5 & T5 & T5 & T3 & T5 & T5 & T1 & T1 \\
    \midrule
    \textbf{Arc-E} & Uhura-Arc Easy & T1 & T4 & T1 & T5 & T3 & T2 & T2 & T1 & T4 & T4 & T1 & T3 \\
    \midrule
    \multirow{2}{*}{{\begin{tabular}[c]{@{}c@{}}\textbf{MMLU} \\ \end{tabular}}}
    & AfriMMLU & T5 & T5 & T1 & T4 & T3 & T1 & T2 & T1 & T1 & T1 & T1 & T1 \\
    & Openai-MMLU & T5 & T4 & T3 & T5 & T5 & T5 & T5 & T5 & T3 & T5 & T1 & T1 \\
    \midrule
    \textbf{Math} & AfriMGSM & T1 & T4 & T3 & T4 & T4 & T4 & T1 & T1 & T2 & T4 & T5 & T2 \\
    \midrule
    \multirow{2}{*}{{\begin{tabular}[c]{@{}c@{}}\\ \\ \\ \textbf{MT} \\ \end{tabular}}}
    & Flores en\_xx & T3 & T2 & T1 & T1 & T2 & T2 & T1 & T2 & T2 & T2 & T2 & T3 \\
    & Flores xx\_en & T1 & T3 & T3 & T1 & T3 & T2 & T3 & T1 & T2 & T2 & T1 & T2\\
    
    & Mafand en\_xx & T1 & T2 & T2 & T2 & T2 & T2 & T1 & T1 & T2 & T2 & T3 & T1 \\
    & Mafand xx\_en & T3 & T2 & T2 & T2 & T2 & T2 & T2 & T1 & T2 & T1 & T3 & T1\\
    
    & NTREX en\_xx & T3 & T2 & T1 & T1 & T2 & T2 & T2 & T1 & T1 & T3 & T2 & T2 \\
    & NTREX xx\_en & T1 & T2 & T3 & T1 & T2 & T2 & T2 & T1 & T2 & T2 & T2 & T2 \\
    
    & Salt en\_xx & T2 & T1 & T1 & T2 & T1 & T2 & T3 & T1 & T3 & T2 & T2 & T3 \\
    & Salt xx\_en & T1 & T2 & T3 & T1 & T3 & T3 & T3 & T1 & T2 & T2 & T2 & T2 \\
    \midrule
    \textbf{Summ} & XLSUM & T3 & T3 & T3 & T3 & T3 & T3 & T1 & T3 & T1 & T1 & T3 & T1 \\
    \midrule
    \textbf{ADR} & ADR & T4 & T4 & T2 & T4 & T4 & T3 & T1 & T3 & T4 & T3 & T2 & T1 \\
    \midrule
    \end{tabular}}
    \vspace{-2mm}
    \caption{Best-performing prompts per model for each dataset. These prompts achieved the highest scores reported in the paper}
    \label{tab:best-prompts}
    \end{table*}
    \vspace{-3mm}
}

\title{AfroBench: How Good are Large Language Models on African Languages?}

\author{%
Jessica Ojo $^{1,3*}$, Odunayo Ogundepo $^{4,5*}$, Akintunde Oladipo $^{4,5*}$, Kelechi Ogueji$^{5*}$,\\ \textbf{Jimmy Lin $^{5}$, Pontus Stenetorp $^{6}$, David Ifeoluwa Adelani $^{1,2*}$}\\
\footnotesize
$^*$Masakhane NLP, 
$^1$Mila - Quebec AI Institute \& McGill University, $^2$Canada CIFAR AI Chair, 
$^3$Lelapa AI,\\
\footnotesize
$^4$The African Research Collective
$^5$University of Waterloo,
$^6$University College London\\
\footnotesize
\texttt{Correspondence:\{jessica.ojo, david.adelani\}@mila.quebec}}
\begin{document}

\maketitle

\begin{abstract}

Large-scale multilingual evaluations, such as MEGA, often include only a handful of African languages due to the scarcity of high-quality evaluation data and the limited discoverability of existing African datasets.
This lack of representation hinders comprehensive LLM evaluation across a diverse range of languages and tasks.
To address these challenges, we introduce \textsc{AfroBench}---a multi-task benchmark for evaluating the performance of LLMs across 64 African languages, 15 tasks and 22 datasets.
\afrobench{} consists of nine natural language understanding datasets, six text generation datasets, six knowledge and question answering tasks, and one mathematical reasoning task.
We present results comparing the performance of prompting LLMs to fine-tuned baselines based on BERT and T5-style models.
Our results suggest large gaps in performance between high-resource languages, such as English, and African languages across most tasks; but performance also varies based on the availability of monolingual data resources.
Our findings confirm that performance on African languages continues to remain a hurdle for current LLMs, underscoring the need for additional efforts to close this gap.\footnote{\url{https://mcgill-nlp.github.io/AfroBench/}}

\end{abstract}

\section{Introduction}

Large language models (LLMs) have risen to the fore of natural language processing (NLP) and also become increasingly commercially viable.
These models have empirically demonstrated strong performance across a variety of NLP tasks and languages~\citep{NEURIPS2020_1457c0d6,xglm,chowdhery2022palm,scaling-iflm}.
However, their performance on low-resource languages (LRLs), such as African languages, is largely understudied.
This is problematic because there is a great disparity in the coverage of languages by NLP technologies. \citet{joshi-etal-2020-state} note that over 90\% of the world’s 7000+ languages are under-studied by the NLP community.
Ideally, approaches to enhance language understanding should be applicable to all languages.

\begin{figure}
    \centering
    \includegraphics[width=1\linewidth]{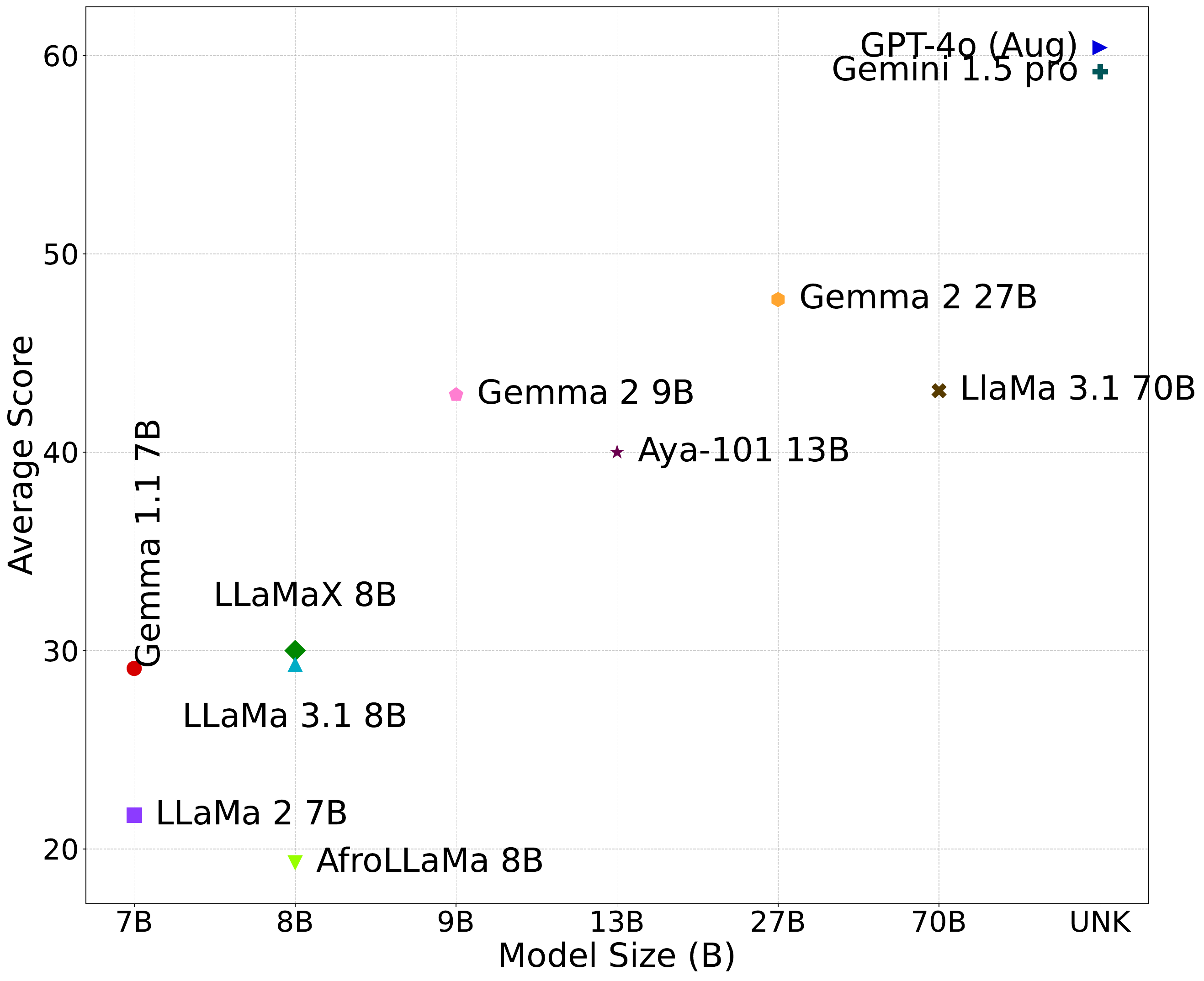}
    \vspace{-6mm}
    \caption{\afrobench{} average score on various LLMs}
    \label{fig:llm-size-vs-score}
    \vspace{-1mm}
\end{figure}

\insertrelatedworks

While there have been some recent evaluation of the performance of LLMs on several languages~\citep{ahuja-etal-2023-mega,lai-etal-2023-chatgpt, robinson-etal-2023-chatgpt}, the evaluation is focused on \textit{closed models} like GPT-3.5 \cite{ouyang2022training} and GPT-4 \cite{OpenAI2023GPT4TR}. 
Megaverse~\citep{ahuja2023megaverse} extended the evaluation to more models such as PaLM 2~\citep{anil2023palm} and LLaMa 2~\citep{Touvron2023Llama2O}, Mistral~\citep{Jiang2023Mistral7}, Gemma~\citep{Mesnard2024GemmaOM} and Gemini Pro~\citep{team2023gemini}. However, previous evaluation faces two main issues: (1) they cover only few tasks for African languages, for example, Megaverse only evaluated on part-of-speech, named entity recognition, and cross-lingual question answering for African languages, primarily due to \textit{poor discoverability} of African languages benchmarks, \textit{limited available evaluation data}, and \textit{bias in the selection} of languages covered in the evaluation.~\footnote{Belebele \cite{bandarkar-etal-2024-belebele} covers over 29 African languages, but Megaverse did not include any in their evaluation. }
(2) Evaluation of LLMs needs to be continuous since many new LLMs have been released with improved multilingual abilities, but a comprehensive evaluation is not available for African languages.


In this paper, we address the challenges of previous large-scale LLM evaluation by introducing a new carefully curated benchmark known as \textbf{\afrobench{} which comprises 15 tasks, 22 evaluation data, and 64 indigeneous African languages.} 
\afrobench{} consists of nine natural language understanding tasks, six text generation tasks, six knowledge and question answering tasks, and one mathematical reasoning task.
Finally, we created a \textbf{new evaluation datasets}, \afriadr for 
diacritic restoration of tonal marks and accents on African language texts. 
Leveraging \textsc{AfroBench}, we conduct an extensive analysis of the performance of LLMs for African languages from different language families and geographical locations.

For our evaluation, we compute the average performance score over the 15 tasks covered in \textsc{AfroBench}. Additionally, we introduce \afrolite{} that only cover a subset of seven tasks and 14 diverse languages in \afrobench{} which reduces the evaluation cost for a newly introduced LLM on our leaderboard. 
\autoref{fig:llm-size-vs-score} shows our evaluation on \afrobench{}, we find that proprietary models such as \gpt{} and \gemini{} achieve $+13$ score improvement over \gemma{}, our best-performing open model. We also compared the performance of English language to 14 African languages, finding that \gpt{} and \gemma{} achieve better performance than African languages by more than $+25$ and $+40$ score improvements respectively. This shows that the gap in the multilingual abilities of open models is wider than that of proprietary models. Finally, we compare the performance of LLMs to fine-tuned models based on AfroXLMR \cite{alabi-etal-2022-adapting}, AfriTeVa V2 T5 model \cite{oladipo-etal-2023-better} and NLLB \cite{nllb2022} whenever training data is present. Results show that prompting LLMs often yield lower average performance than the fine-tuned baselines.  Our findings show that more effort is needed to close the gap between the performance of LLMs for high-resource languages and African languages. 

\begin{figure*}[h!tbp]
    \centering
    \begin{center}
        \includegraphics[trim=0 8 0 0,clip,width=0.9\textwidth]{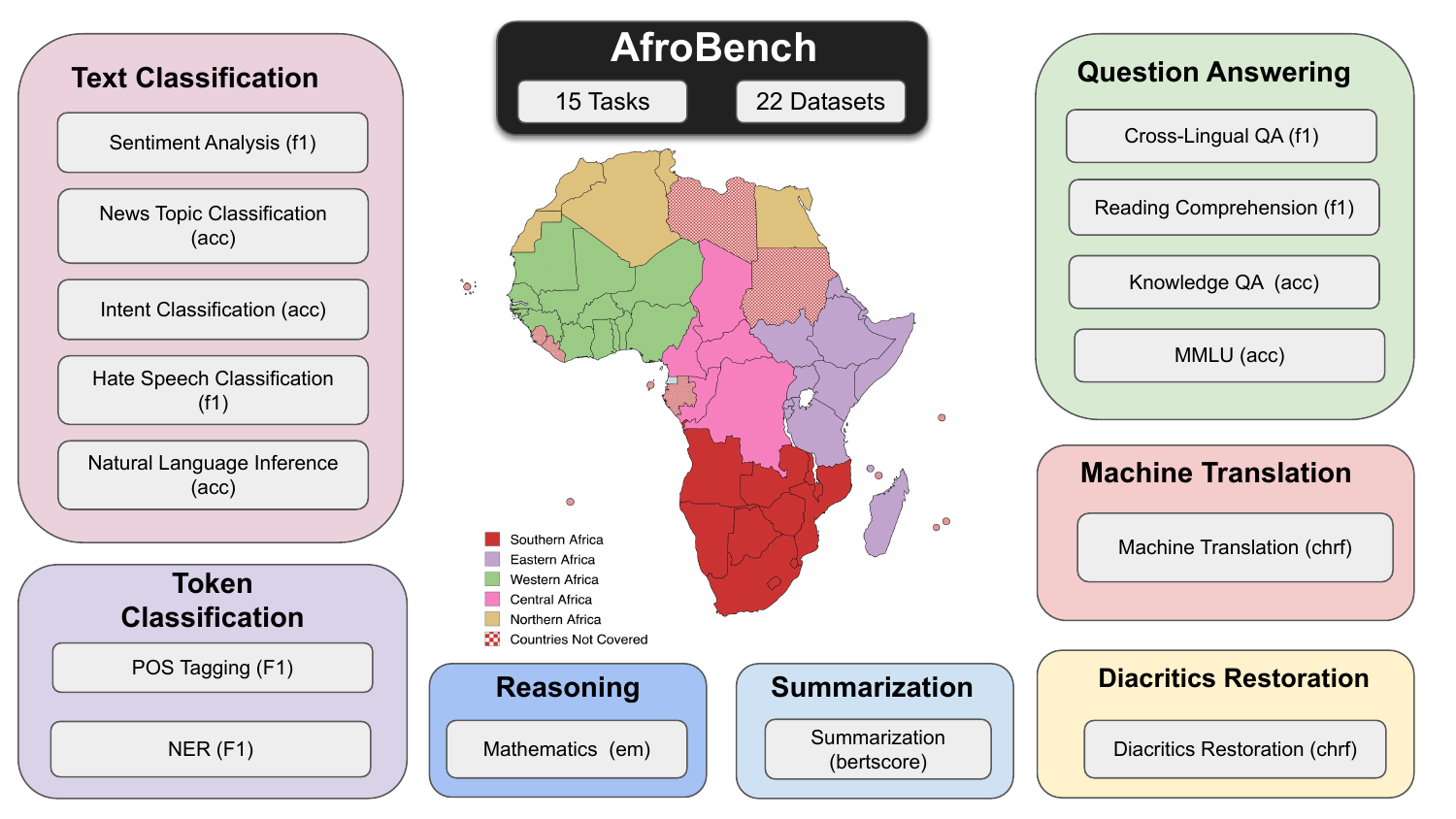}
       \vspace{-2mm}
        \caption{\textbf{\afrobench{}: A comprehensive benchmark for evaluating performance of LLMs on African Language tasks}. The benchmark features 15 distinct tasks across 22 datasets and 64 indigeneous African languages. The benchmark covers diverse tasks with geographical coverage spanning different regions in Africa.}
        \label{fig:afrobench-overview}
    \end{center}
    \vspace*{-\baselineskip}
\end{figure*}

\section{Related Work}

\noindent\textbf{Large Language Model Evaluation:} Accurate and reproducible evaluation of language models is important as more and more models are being released.
As these models are integrated into various applications, developing robust evaluation frameworks becomes paramount for understanding their true capabilities and limitations.
As a result, the community has worked on developing evaluation frameworks \citep{eval-harness, lighteval, liang2023holistic}, leaderboards \citep{10.5555/3692070.3692401, srivastava2023beyond, open-llm-leaderboard-v2} and benchmarks \citep{Adelani2024IrokoBenchAN, zhou2023instructionfollowingevaluationlargelanguage, hendrycks2021measuringmathematicalproblemsolving}.
While each of these evaluation tools focuses on assessing specific aspects of language model capabilities - from basic linguistic understanding to complex reasoning tasks - the development of truly comprehensive benchmarks remains a significant challenge \citep{ruder2021benchmarking, biderman2024lmevaluation}.
These challenges stem from complex nature of language understanding and the stochastic nature of language models 

\paragraph{Multilingual LLM Benchmarks:}
Benchmarks serve as a standard for measuring how systems have improved over time on across specific tasks and metrics.
In the context of LLMs, multilingual benchmarks are crucial to assessing both the quality and practical utility of these models across diverse languages and tasks.
Our primary focus lies in understanding LLM performance specifically for African languages, with several notable benchmarks having emerged in recent years to address this need. ChatGPT-MT~\cite{robinson-etal-2023-chatgpt} evaluated the translation capability of GPT-4 and they find that it's demonstrates strong performance on high-resource languages, the performance on low-resource languages is subpar.
Belebele~\cite{bandarkar-etal-2024-belebele} is a question answering task in 122 languages including 28 African languages for assessing reading comprehension abilities of LLMS.
Mega~\cite{ahuja-etal-2023-mega} and Megaverse~\cite{ahuja-etal-2024-megaverse} are multi-task multilingual and multimodal benchmarks in 83 languages including 16 African languages. \autoref{tab:works} provides a summary of the related works. 

While these existing benchmarks have provided valuable insights, they collectively highlight a pressing need for more comprehensive evaluation that encompass a broader range of African languages and diverse tasks.
Our research, through the development of \afrobench{}, addresses this gap by building upon and complementing existing work. 
We create a robust evaluation framework that assesses LLM performance across 64 African languages, evaluating capabilities across 15 distinct tasks. This expanded scope allows for a more nuanced and thorough understanding of LLM capabilities in African language contexts.

\section{AfroBench}

\afrobench{} is a comprehensive LLM evaluation benchmark  designed to assess both proprietary and open LLMs across diverse Natural Language Processing (NLP) tasks in African languages.
As shown in \autoref{fig:afrobench-overview}, the benchmark encompasses 15 distinct tasks, spanning Natural Language Generation (NLG) and Natural Language Understanding (NLU), incorporating 22 curated datasets in 64 African languages.
These evaluation tasks extend beyond traditional NLP benchmarks, such as text classification and named entity recognition, to include more challenging benchmarks such as mathematical reasoning and knowledge QA. 

Each task within \afrobench{} has been carefully selected to assess different aspects of language model capabilities, from basic linguistic competency to more complex reasoning abilities.
\afrobench{} also provides valuable insights into model behavior across different African language families and their unique linguistic features.
All tasks and sub-tasks within \afrobench{} are evaluated using both zero-shot and few-shot prompting  to guide model responses.
To ensure consistent and reliable evaluation, we implement task-specific response constraints to facilitate systematic extraction and analysis of model outputs.
For completion, we compare against existing SoTA encoder-only and encoder-decoder architectures that have previously demonstrated superior performance on individual tasks within the benchmark.
This enables us to directly compare the performance of specialized models to general-purpose LLMs.

\autoref{tab:dataset_summary} summarizes the tasks, the dataset used, number of languages covered, and total size.

\subsection{Languages}

We cover 64 African languages from seven language families~(Afro-Asiatic, Atlantic-Congo, Austronesian, Indo-European, Mande, Nilotic, and English-Creole).
40 languages are from the Atlantic-Congo family, 12 from the Afro-Asiatic family,  seven from Nilotic family, 2 Indo-European, 2 Creole languages, and 1 Austronesian language. 
\autoref{fig:afrobench-overview} shows the geographical distribution of the languages covered in \afrobench{} and the full list of languages can be found in \autoref{appendix:languages}.

\subsection{Evaluation tasks}
Our evaluation spans multiple datasets across 15 NLP tasks.
While some of these multilingual datasets cover languages across several continents, we focus specifically on the African language subsets, along with select high-resource languages (English, French, Portuguese and Arabic), due to their widespread use across different African regions. \autoref{tab:dataset_summary} details the testsize and number of languages evaluated per task per dataset. We present a breakdown of the tasks, sub-tasks and specific datasets contained in \afrobench{}.

\begin{table}[t]
\centering
\footnotesize
\resizebox{\columnwidth}{!}{%
\begin{tabular}{llrrcc}
\toprule
 &  & \textbf{Total} & \textbf{No. of} & \textbf{Per. Lang.} \\
\textbf{Task} & \textbf{Dataset} & \textbf{Size} & \textbf{Lang.} & \textbf{size} \\

\midrule
POS & MasakhaPOS & 12,190 & 20 & 500--700 \\
NER & MasakhaNER-X & 18,192 & 20 & 900--1000\textsuperscript{*} \\
\rowcolor{Gray}
TC & SIB-200 & 11,220 & 55 & 204 \\
 & MasakhaNEWS & 6,242 & 16 & 200--948 \\ 
SA & AfriSenti & 37,670 & 15 & 950--4500\textsuperscript{‡} \\
 & NollySenti & 2,500 & 5 & 500 \\
\rowcolor{Gray}
Intent & Injongo-Intent & 10,880 & 17 & 640 \\
Hate & AfriHate & 14,250 & 15 & 323--1600 \\ 
\rowcolor{Gray}
NLI & AfriXNLI & 9,600 & 16 & 600 \\
XQA & AfriQA & 3,107 & 9 & 250--500 \\
\rowcolor{Gray}
RC & Belebele & 27,900 & 31 & 900 \\
 & NaijaRC & 357 & 3 & 80--190 \\
QA & Uhura-Arc-Easy & 3,257 & 7 & 300--500 \\
\rowcolor{Gray}
MMLU & AfriMMLU & 8,500 & 17 & 500 \\
 & MMMLU & 42,126 & 3 & 14042 \\
\rowcolor{Gray}
Math & AfriMGSM & 4,500 & 18 & 250 \\
\rowcolor{Gray}
MT & Flores-200 & 58,696 & 58 & 1012 \\
 & MAFAND & 29,155 & 21 & 1000--2000 \\
 & NTREX & 48,000 & 24 & 2000 \\
 & Salt & 3,500 & 7 & 500 \\
Summ & XLSum & 25,769 & 12 & 500--1300\textsuperscript{¶} \\
ADR & AfriADR & 7,567 & 5 & 1400--1600 \\
\bottomrule
\end{tabular}
}
\caption{\textbf{AfroBench data statistics:} We detail the dataset evaluated per task, test set size and number of languages for each dataset as well as the range of sample per language, 
\textsuperscript{*}excl. amh: 500 \& luo: 185 (in MasakhaNER-X), 
\textsuperscript{‡}excl. tso: 254 (in AfriSenti), and 
\textsuperscript{¶}excl. arb: 4689 \& eng: 11,535. (in XLSum). The tasks covered in the \textbf{Lite} version is highlighted in \colorbox{Gray}{Grey}.}
\label{tab:dataset_summary}
\end{table}

\subsubsection{Text Classification}

\paragraph{Sentiment Classification (SA):} 
We evaluate \nollysenti~\citep{shode-etal-2023-nollysenti} and \afrisenti~\citep{muhammad2023afrisenti}.
\afrisenti evaluates sentiment analysis of tweets across 14 African languages, while \nollysenti focuses on movie review sentiment 
in four African languages. 

\paragraph{Topic Classification (TC):} We evaluate \sib and \masakhanews~\citep{adelani2023masakhanews} that cover 53 and 14 African languages, respectively. 
The topic categories could be general topic such as  \textit{business}, \textit{entertainment}, and \textit{health}.

\paragraph{Intent Classification:}
\injongo~\citep{Yu2025INJONGOAM} is an intent classification task in 16 African languages. 
The goal is to classify an utterance into one of 40 intent types from different domains such as \textit{Banking} (e.g. ``freeze account''),  \textit{Home} (e.g. ``play music''),  \textit{Kitchen and Dining} (e.g. ``cook time''), and \textit{Travel} (e.g. ``plug type''). 

\paragraph{Hate Speech detection:} \afrihate~\citep{muhammad2025afrihatemultilingualcollectionhate} is a multilingual 
hate speech and abusive language datasets in 15 African languages for tweets. Each tweet can be categorized into one of \textit{abusive}, \textit{hate} or \textit{neural} label.


\paragraph{Natural Language Inference (NLI):}
\afrixnli\citep{Adelani2024IrokoBenchAN} is a dataset collection in 16 African languages where each sample is a pair of sentences (a premise and a hypothesis) and the task is to classify each pair as an \textit{entailment}, \textit{contradictor} or \textit{neural} pair.

\subsubsection{Token Classification}

\paragraph{Named Entity Recognition~(NER):}
We evaluate entity recognition for 20 African languages on
\masakhanerx~\citep{Ruder2023XTREMEUPAU}---an extension of \masakhaner dataset~\citep{adelani-etal-2021-masakhaner, adelani-etal-2022-masakhaner}  that converts NER tags from CoNLL format into a text generation task of predicting entities with a delimiter, ``$\$$'' between them.

\paragraph{Part-of-Speech Tagging (POS) :}
\masakhapos\citep{dione-etal-2023-masakhapos} is a part-of-speech tagging dataset in 20 African languages created from news articles. Each token is categorized into one of the 17 POS tags.

\subsubsection{Reasoning:}
\paragraph{Mathematical reasoning (Math)} We evaluate on \afrimgsm~\citep{Adelani2024IrokoBenchAN}, an extension of the MGSM dataset to 17 African languages. The question is a grade school level question, and a single digit answer.

\subsubsection{Question Answering}

\paragraph{Cross-Lingual Question Answering~(XQA):} \afriqa~\citep{ogundepo2023afriqa} is a cross-lingual QA task with questions in 10 African languages and context passages in English or French. The goal is to extract the span with the right answer from the text, similar to a cross-lingual reading comprehension. 

\paragraph{Reading Comprehension (RC):} We evaluate on \naijarc~\citep{aremu2024naijarc}, a multi-choice reading comprehension dataset in three African languages
and \belebele~\citep{bandarkar-etal-2024-belebele}, a  multi-choice reading comprehension task for 122 languages including 29 African languages. 

\paragraph{Knowledge QA:} We focus on two human-translated \textbf{MMLU} datasets: 
\openaimmlu~\footnote{https://huggingface.co/datasets/openai/MMMLU} and \afrimmlu~\citep{Adelani2024IrokoBenchAN} that covers 3 and 16 African languages respectively. 
Both tasks span multiple subjects and follow a four-option multiple-choice format. Although, the subjects covered by \afrimmlu are only five. We also extend our evaluation to the human translation of \textit{scientific} \textbf{Arc-Easy} benchmark in six African languages~\uhura~\citep{Bayes2024UhuraAB}. 

\subsubsection{Text Generation}

\paragraph{Machine translation (MT):} Our MT benchmark includes the following datasets: \flores\citep{goyal-etal-2022-flores}, \mafand\citep{adelani-etal-2022-thousand}, \ntrex\citep{federmann-etal-2022-ntrex} and \salt\citep{akera2022machine} covering $57$, $21$, $23$ and $7$ translation direction to African languages. All translations are from English except for the \mafand benchmark with a few languages whose source is French. 

\paragraph{Summarization (Summ):} Given a news article, our goal is to generate its summary based on the popular
\xlsum dataset~\citep{hasan-etal-2021-xl} covering $10$ African languages. 

\paragraph{Automatic Diacritics Restoration (ADR):} This is a \textbf{new benchmark} we introduce called \textbf{\afriadr}. Given a sentence in a language, say ``\textit{Sugbon sibesibe, Mama o gbagbo}'' (in \yoruba), the model's goal is to add the missing tonal marks and accents, say ``\textit{{\d S}{\` u}gb{\d {\' o}}n s{\' i}b{\d {\` e}}s{\' i}b{\d {\` e}}, M{\` a}m{\' a} {\` o} gb{\` a}gb{\d {\' o}}}''. We cover five African languages for this task: \textit{\ghomala}, \textit{Fon}, \textit{Igbo}, \textit{Wolof}, and \textit{\yoruba}. To create \afriadr, we selected the five languages with extensive use of diacritics from \mafand MT dataset, then, we strip all accents and diacritics on each sentence, and use it as the ``source'' text, while the ``target'' has the fully diacritized texts. \autoref{tab:adr_examples} shows details of data size and example sentence for each language in \afriadr. 

\subsection{\textbf{AfroBench-Lite}: A cost-effective bench}
Following the idea of Global-MMLU-Lite~\citep{Singh2024GlobalMU} in creating a cost-effective benchmark with fewer languages and samples. We introduce \afrolite, a subset of \afrobench{} covering 14 languages and seven datasets (and tasks):  \sib(TC), \injongo(Intent), \afrixnli (NLI), \belebele (RC), AfriMMLU (MMLU), AfriMGSM (Math), and Flores (MT). The languages covered are very typologically-diverse, and have different resource-level~\citep{kudugunta2023madlad}, they include: \textit{English}, \textit{Kiswahili}, \textit{Kinyarwanda}, \textit{Hausa}, \textit{Amharic}, \textit{isiXhosa}, \textit{chiShona}, \textit{isiZulu}, \textit{Igbo}, \textit{\yoruba}, \textit{Sesotho}, \textit{Lingala}, \textit{Oromo}, \textit{Luganda}, and \textit{Wolof}.

\begin{table}[t]
 \resizebox{\columnwidth}{!}{
\centering
\begin{tabular}{lrl}
    \toprule
    \textbf{Lang.} & \textbf{Size} & \textbf{Example sentence } \\
    \midrule
    \multirow{2}{*}{\texttt{\ghomala}} & \multirow{2}{*}{1430}  & \textbf{Input:} A jw\textschwa \ gu\textipa{N} ts\textschwa \ aw$\varepsilon$ a l\textschwa \ nə\textipa{N} kwit\textschwa \ \\
     & &  \textbf{Target:} \^{A} jw\'{ə} gu\textipa{N} ts\'{ə} \ aw{$\acute{\varepsilon}$} a l\textschwa \ n\'{ə}\textipa{N} \ kw\'{i}t\'{ə}  \\
    
    \midrule
     \multirow{2}{*}{\texttt{Fon}} & \multirow{2}{*}{1579}  & \textbf{Input:} Din \textopeno‚ nu l$\varepsilon\varepsilon$ bi j\textepsilon wexo.   \ \\
     & &  \textbf{Target:} Din \textopeno‚ n\'{u} l$\acute{\varepsilon}\varepsilon$ \  b\v{i} j$\varepsilon$ \ wexo.  \\

    \midrule
     \multirow{2}{*}{\texttt{Igbo}} & \multirow{2}{*}{1500}  & \textbf{Input:} Akuko ndi ga-amasi gi: \\
     & &  \textbf{Target:} Ak\d{u}k\d{o} nd\d{i} ga-amas\d{i} g\d{i}:
 \\
    
    \midrule
     \multirow{2}{*}{\texttt{Wolof}} & \multirow{2}{*}{1500}  & \textbf{Input:} Naari taggatkat lanu yu xaran lu kawe.\\
     & &  \textbf{Target:}  Ñaari tàggatkat lañu yu xarañ lu kawe. \\
    
    \midrule
     \multirow{2}{*}{\texttt{\yoruba}} & \multirow{2}{*}{1558}  & \textbf{Input:} Isokan awon Oniroyin naa fe oro naa loju: \\
     & &  \textbf{Target:} \'{I}\d{s}\d{\`o}kan \`{a}w\d{o}n On\'{i}r\`{o}y\`{i}n n\'{a}\`{a} f\d{e} \d{\`{o}}r\d{\`{o}} n\'{a}\`{a} l\'{o}j\'{u}: \\
    \bottomrule
    
\end{tabular}
}
 \vspace{-3mm}
 \caption{\textbf{AfriADR dataset}: Language, test size, and Example sentence}
    \label{tab:adr_examples}
\end{table}

\section{Experimental setup}
\label{method}

\subsection{Evaluation Framework}


We model all tasks as text-generation problems, where we combine inputs with prompts to guide language models in generating outputs under specific constraints.
To ensure robust evaluation, we employ multiple prompts for each task with few- and zero-shot examples, which helps maintain consistency and minimize potential biases across different models.

Our evaluation framework is fully integrated with Eleuther LM Evaluation Harness \citep{eval-harness}\footnote{\url{https://github.com/EleutherAI/lm-evaluation-harness/tree/main/lm_eval/tasks/afrobench}} with custom evaluation scripts to run open-source models. However, for the proprietary models, we developed a 
custom framework for prompting various LLMs via API including open models available on TogetherAI API. \footnote{\url{https://github.com/McGill-NLP/AfroBench/tree/main/prompt_with_API}}  These tools are open source, easily accessible, and reproducible. Details of custom framework and Eleuther LM Evaluation Harness integration in Appendix \ref{sec:tool_integration}

\subsection{Fine-tuned baselines}
For the tasks with available training data, we use available task-specific trained models, such as NLLB-200 3.3B for MT, and fine-tuned multilingual encoders or encoder-decoder T5 models on applicable datasets. We fine-tune AfroXLMR~\citep{alabi-etal-2022-adapting} --- one of the SoTA BERT-style encoders for African languages on each of the NLU tasks. For summarization and ADR, we fine-tune AfriTeVa V2 Large~\citep{oladipo-etal-2023-better} on the available training data of each task. While AfriTeVa V2 outperformed mT5~\citep{xue-etal-2021-mt5} overall, its tokenization failed for Fon language, so we fine-tune mT5-large, which as a more diverse tokenizer, for the language. 

\subsection{LLMs Evaluated}
We evaluate two broad categories of Large Language Models (LLMs): \textbf{Open Models} and \textbf{Closed Models}. We evaluate 10 open models: \llamaTwo~\cite{Touvron2023Llama2O}, \gemmaO{}~\cite{Mesnard2024GemmaOM}, \textsc{LLaMa 3} series (3 8B, 3.1 8B and 3.1 70B)~\cite{Dubey2024TheL3}, \llamaX~\cite{lu-etal-2024-llamax} (an adapted LLaMa 3 8B to 100 languages), \afrollama \footnote{\nolinkurl{https://huggingface.co/Jacaranda/AfroLlama_V1}} (an adapted LLaMa 3 8B to Swahili, Xhosa, Zulu, Yoruba, Hausa and English languages), \textsc{Gemma 2} (9B \& 27B)~\cite{Riviere2024Gemma2I}, and \aya (an instruction-tuned mT5 encoder-decoder model on massively multilingual prompted dataset). Finally, we evaluate on two popular proprietary models: \gpt{} and \gemini{}~\cite{Reid2024Gemini1U}. We provide full description of the LLMs in Appendix~\ref{sec:llm_evaluated}.

\paragraph{Prompts used for evaluation}
We make use of \textit{five} different prompts in the evaluation of each task except the text generation tasks, and we report the best prompt in the paper. For the text generation tasks, we reduce the number of prompts to \textit{three} since the generation is often time consuming and expensive especially for summarization tasks. Moreover, we find that performance is less sensitive to prompt templates, unlike the NLU tasks. The prompt templates are provided in Appendix \ref{sec:prompt-bank}. 

\paragraph{Few shot evaluation}
We restrict the few shot evaluation to the best closed and open models. We fixed the number of examples to \textit{five}, except for AfriMGSM whose number of examples is \textit{eight} \footnote{8-shot samples is the standard setting for MGSM datasets}. 

\insertafrobenchresults

\section{Results}
\label{results}

\subsection{AfroBench Evaluation}
\autoref{tab:afrobench_results} shows the overall results across all the 15 tasks and 22 datasets. We report only the best prompt results. The average results across all the five prompts and confidence interval is provided in \autoref{sec:afrobench_ci}.   

Our \textbf{first} observation is that closed models such as \gpt{} and \gemini{} achieve better performance than the best open model, \gemma{} with differences of $+12$ or more points on average performance. This shows that the gap in performance is wider for low-resource African languages than for high-resource languages, such as English, when using open models. \textbf{Secondly}, we find that performance gap varies across different tasks. Knowledge intensive and reasoning tasks such as \textsc{Arc-Easy}, \textsc{MMLU}, \textsc{Math} have the largest gaps of $+29.4$, $+19.9$, $+22.6$ respectively, when we compare the performance of \gpt{} to \texttt{Gemma 2 27B}. In general, performance gets better with newer versions of LLMs (e.g. \gemmaO{} vs. \gemmaS{} and \llamaTwo vs. \llamaThreeOne) and model sizes (\gemmaS{} and \gemma{}). This suggests that newer iterations of models are getting better on low-resource languages, although with limited improvements on knowledge intensive tasks. \textbf{Finally}, while LLMs have made significant progress, they still fall behind their \textit{fine-tuned baselines} (\textbf{FT. AVG}) when training data is available for a task. The gap in performance is around $+11.5$ on average, showing that curating annotated datasets for low-resource languages is still beneficial since the capabilities of LLMs lags behind. We provide task and per-language results in Appendix \ref{sec:task_results} and \ref{sec:detailed_lang_results}.






\subsection{AfroBench-Lite Evaluation}
In the \afrolite{}~ evaluation, we restrict the evaluation to seven LLMs, and seven tasks, and compare performance gap to English. 

\paragraph{Large gap in performance when compared to English} One striking observation is that open models such as \llamaL and \gemma{} have competitive performance to closed models on English language with $-5$ to $-2$ performance gap. However, when compared to African languages, \gpt{} and \gemini{} achieves an average score better than \gemma{} by more than $20$ points on \afrolite{}. These results suggest that current LLMs especially the open models, are more biased towards \textit{English} and a few high-resource languages. Adapting LLMs for a region of African languages could help bridge the gap. For instance, we see that continually pre-training LLaMa 3 8B, that resulted in \llamaX shows slight overall performance of $+1.4$ or more over vanilla \llamaThree 
in \autoref{tab:afrobench_results}. However, to further boost performance, better adaptation techniques are needed. 

\insertafrobenchlite

\begin{figure*}
    \centering
    \includegraphics[width=0.95\linewidth]{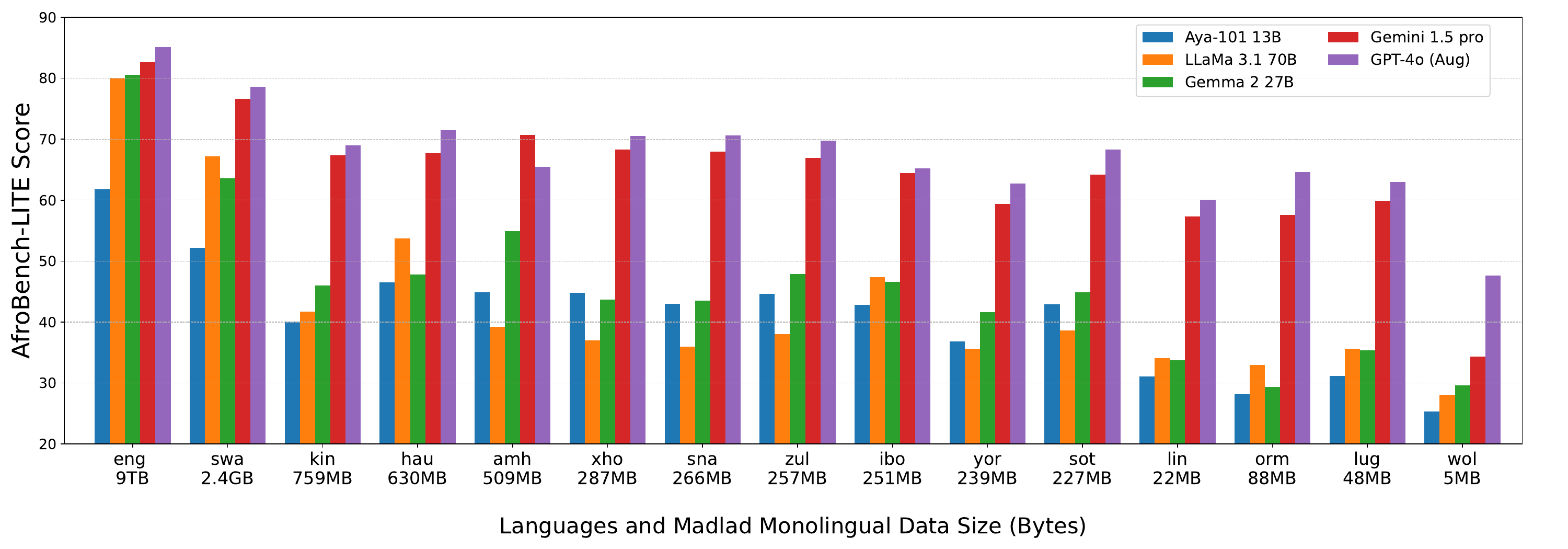}
    \vspace{-3mm}
    \caption{AfroBench-Lite performance of various models across African languages, plotted against the availability of monolingual data (MADLAD byte size).}
    \label{fig:results_per_language}
\end{figure*}

\insertfewshot

\smallskip
\paragraph{Performance varies across languages} \autoref{fig:results_per_language} shows the results for per-language performance scores of 14 languages in \afrolite{}. Our result shows that performance correlates with the available monolingual text on the web~\citep{kudugunta2023madlad}. We find that Swahili (\texttt{swa}) with over 2.4GB of monolingual text has the highest performance among the African languages, while Wolof with the smallest monolingual data (5MB) has the lowest performance. While this data size estimates are approximate, it shows that there is a need to invest more on developing language texts for many African languages for them to benefit in the LLM age. For most languages, \gpt{} gives the best overall results except for Amharic (\texttt{amh}) where \gemini{} was better. For the open models, \gemma{} achieves better performance on eight out of the 14 languages, even better than \llamaL that is more than twice its number of parameters. Although \aya covers 100 languages in its pre-training and often achieves better performance on NLU tasks in \afrolite{}, it often struggles with math reasoning and MMLU, leading to worse overall results.






\begin{figure*}[t]
    \centering
    \includegraphics[width=\linewidth]{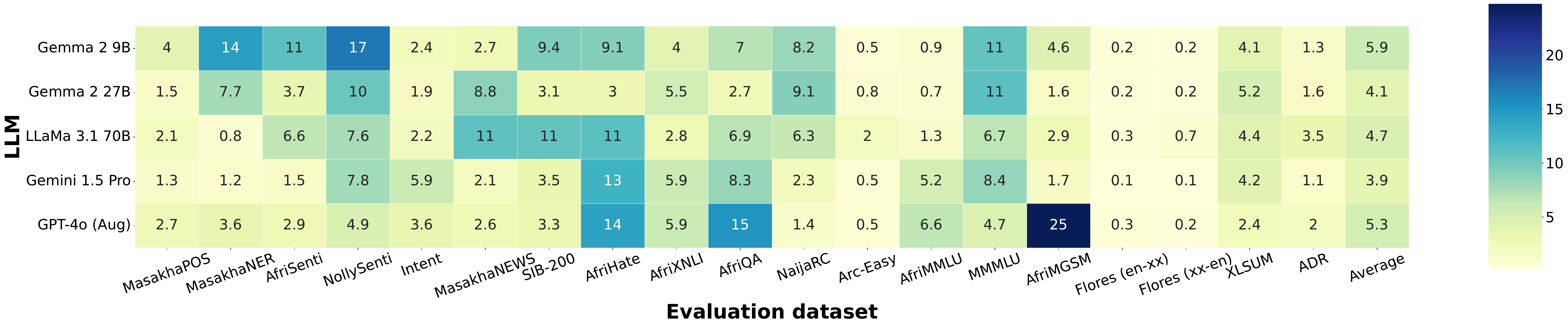}
    \caption{\textbf{Prompt Variability}: Heatmap of the difference between the Best and Average prompt results. }
    \label{fig:variability}
\end{figure*}

\subsection{Few-shot results}
\autoref{tab:few-shot-results} shows the result of zero-shot and few-shot evaluation on three LLMs: \gemma{}, \gemini{} and \gpt{}. The benefit of few-shot varies for different LLMs and tasks. For \gpt{}, we find that across all tasks, there is an average improvement of $+1.8$ while the other LLMs dropped in performance on average. The tasks that benefits the most from the few-shot examples are math reasoning, hate speech detection and ADR with $+4.9$, $+5.8$,  and $+7.8$ respective points improvement. The result shows that few-shot examples are important for teaching LLM a new task it is unfamiliar with such as ADR since the rules of adding diacritics are not provided during the zero-shot, therefore, 5-examples, provides some demonstration to the LLMs on how to perform the task especially for low-resource languages such as \ghomala and Fon with small monolingual data on the web.  These two languages improved by $+16.4$ and $7.2$ respectively, while the other languages such as Igbo, Wolof and \yoruba achieved more than $+5.0$ boost in chrF scores. Similarly, for \gemini, we observed consistent performance boost for ADR with 5 demonstration examples. 

For both \gpt{} and \gemini, there is a significant boost in performance across all the text generation tasks we evaluated, which shows that the current model's have weaker generative capabilities in these low-resource languages, except provided with few shots examples. 
For Hate speech, we provided detailed explaination on the distinction between ``abusive'' content and ``hate'' in the prompt, but this is often confusing even for native speakers of the language, who often need examples of such sentences to improve annotation. We found that LLMs also require such additional examples to be able to better predict if a tweet is offensive. In general,  \gemma{} improved for several NLU but did not benefit from additional examples for the token classification, math reasoning, summarization and ADR tasks. 



\section{Discussion}
\subsection{Prompt variability}
In our evaluation, we present results for the \texttt{Best prompt} rather than the \texttt{Average results} over several prompts to ensure no LLM is at a disadvantage due to their sensitivity to prompt templates. Here, we analyze the difference in the performance scores between the Best prompt and the average over five prompts (or three prompts for the NLG tasks). 

\autoref{fig:variability} shows the result of our analysis across 18 tasks. Our \textbf{first observation} is that LLMs are not sensitive to different prompts when evaluating text generation tasks, all LLMs have lower than $6$ point difference, and the task that is the least sensitive is machine translation (\flores). The \textbf{second observation} is that \gemini{} is the least sensitive LLM to different prompt templates on average. The gap in performance across different prompts is often small for several NLU tasks. Interestingly, we find that \gpt{} is very sensitive to prompts for a few tasks such as hate speech, cross-lingual QA and math reasoning---which explains the large difference in performance scores. This analysis shows the benefit of using several prompts in evaluation, although, the benefit for text generation tasks are limited. Finally, we find that the largest variability is by a small sized \gemmaS, which shows that, smaller LLMs requires more prompt template search than bigger models as shown that \gemma{} is less sensitive.

\subsection{Qualitative Analysis}
\insertqualitative

\autoref{tab:qualitative-analysis} shows the benefit of few-shot examples on ADR, hate speech and math reasoning---the three tasks that improved the most with few-shot examples. For the ADR evaluation on \ghomala, we saw more than $60.0$ chrF point improvement, and noticed that only few characters have the wrong diacritics unlike the zero-shot setting. Similarly, for hate speech, without the few-shot example, the LLM focused on the abusive word ``oloriburuku'' (i.e. brainless), however, when we consider the \textit{target} to tweet, it is obvious that it was referring to an entire tribe in Nigeria, which is ``hate''. In the definition of ``hate'' provided in the prompt, and some examples provided, this is clearer to the model than without any demonstration examples. Finally, for the math reasoning, in zero-shot, the LLM often has \textit{incorrect} and \textit{short} reasoning steps about the \yoruba question which leads to an incorrect answer  . However, when provided with few-shot in the language, \gpt{} came up with more appropriate reasoning steps,  leading to the \textit{correct} answer. 
This observation is particularly exciting for many low-resource languages. 


\section{Conclusion} 
In this paper, we introduce a new benchmark, \afrobench{}, that aggregates existing evaluation datasets for African languages, and added a \textit{new} dataset focused on diacritics restoration. \afrobench{} comprises 15 NLP tasks, 22 datasets, and 64 African languages under-represented in NLP. We evaluate the performance of several closed and open LLMs on these tasks, showing that they all fall behind the fine-tuned baselines. 
We also show large performance gap compared to English, although we notice the gap is smaller for closed models such as \gpt{} and \gemini{}. Through this benchmark, we have created a leaderboard focusing on LLM evaluation for African languages, which will be maintained going forward with additional tasks, LLMs and languages. We will be releasing our prompts and tasks configurations to Eleuther \textit{lm-eval}. We hope this encourages the development of more African-centric LLMs for African languages.
\footnote{Our evaluation suite is available at: \\ 
\href{https://github.com/The-African-Research-Collective/afrobench-eval-suite}{The-African-Research-Collective/afrobench-eval-suite}.} Our aim is to continuously add newer LLMs to the leaderboard, we demonstrate this by adding the following LLMs to the \afrolite{}: Lugha-LLaMa (an African-centric LLM)~\citep{Buzaaba2025LughaLlamaAL}, GPT-4.1, Gemini-2.0-Flash, and LLaMa 4 400B (Maverick) as shown in \autoref{sec:afrobench_lite}. 

\section{Acknowledgement}
This research was supported in part by the Natural Sciences and Engineering Research Council (NSERC) of Canada. 
Additional funding is provided by Microsoft via the Accelerating Foundation Models
Research program. 
We are grateful for the funding of IVADO and the Canada First Research Excellence Fund. 
We would also like to thank Google Cloud for the GCP credits Award through the Gemma 2 Academic Program, and OpenAI for providing us access for providing API
credits through their Researcher Access API Program. Finally, we thank Israel Abebe for contributing the inference results of GPT-4.1 and Lugha-LLaMa to the \afrolite{} leaderboard.


\section{Limitation}

In today's NLP landscape, large language models are generalist models that are capable of performing multiple NLP tasks without the need for special traininig on these tasks.
These models are often multilingual and are able to perform tasks in multiple languages.
Our research examines how these models perform specifically with African languages, revealing performance disparities when compared to more resourced languages.
In this section, we discuss some of the limitations of our research methodology and findings.

\smallskip
\noindent\textbf{1. Training Data Transparency and Contamination:} One of the challenges in evaluating large language models lies in the limited visibility into their training data composition. 
While organizations frequently publish training documentation, many reports lack comprehensive details about data mixtures and language distributions across different training stages.
There are multiple ways that this lack of transparency impacts the findings of our research.
Without knowledge of the data mixture, we cannot determine whether or by how much or evaluation sets overlap with the training dataset.
Thus, we cannot conclude that superior performance on certain tasks is a true demonstration of generalization or merely the models exposure to similar content during training.
In the context of African languages,  knowledge about the training data helps us access other factors such as cross-lingual transfer that might help us understand and better analyze evaluation results.
A clear understanding of training data composition serves as a crucial foundation for meaningful model evaluation. It helps establish the validity of performance metrics and provides essential context for interpreting results across different languages and tasks.

\smallskip
\noindent\textbf{2. Limited Selection of LLMs and Evaluation Costs: }
We are only able to evaluate a limited set of LLMs due to the computational and financial costs associated with model access and inference.
Language models are accessed using two primary methods; loading the pretrained checkpoints directly or via an API service.
While providers like Together AI offer access to open-source models and companies like OpenAI provide proprietary model access, both approaches incur considerable costs that directly impact the scope of evaluation studies.
In our evaluation, the costs were substantial, requiring approximately \$2,500 each for \gemini{} and \gpt{} model access, with an additional \$1,200 for utilizing the Together.AI platform.
The total evaluation costs manifests in two key dimensions; First when running the models locally, the GPU requirements for larger models is substantial and secondly while utilizing API services, the cost scales directly with the size of the evaluation dataset and number of models.
These cost implication impose a limitation on the breadth and depth of our evaluation studies.
We had to make strategic decisions about which models to include in our benchmark and how extensively to test them.
This financial constraint introduces a selection bias on which models and tasks to prioritize which limits the scope of our evaluation

\noindent\textbf{3. Long-tail Distribution of Languages Across Tasks \& Datasets: }
Another limitation of \afrobench{} is the uneven distribution of languages across tasks and datasets.
While our evaluation covers 64 languages in total, the coverage across tasks and datasets exhibits a long-tail distribution.
As shown in \autoref{tab:tasks_by_languages}, 60\% of the languages appear in fewer than 5 of the 21 datasets.
This poses two challenges; first, it limits our ability to properly access the performance of LLMs across these underrepresented languages.
Secondly, it highlights the gap in the availability of evaluation datasets even among low-resource languages.
Without extensive dataset coverage for these languages, conclusions about LLM capabilities across these languages remains tentative.

\noindent\textbf{4. Contraints in Machine Translation Metrics: }
Machine translation is often evaluated using BLEU and ROUGE, which rely on word-level recall and precision, and chrF, which operates at the character level.
Research has shown these metrics sometimes demonstrate poor correlation with human judgments of translation quality.
Other evaluation metrics that utilize embedding similarity, such as BERTScore~\citep{Zhang*2020BERTScore:} and COMET~\citep{rei-etal-2020-comet} / AfriCOMET~\citep{wang-etal-2024-afrimte}, which leverage pretrained encoder models to generate scores by comparing translations against reference texts, are promising alternatives.
However, these neural evaluation models have limited language coverage, making them unsuitable for many of the languages in our study.
As a result, we rely on chrF++, which combines unigram and character n-gram overlap measurements. While this metric provides broader language coverage, it is a compromise between evaluation quality and practical applicability.


\bibliography{custom}

\begin{thebibliography}{66}
\providecommand{\natexlab}[1]{#1}

\bibitem[{Adelani et~al.(2022{\natexlab{a}})Adelani, Alabi, Fan, Kreutzer, Shen, Reid, Ruiter, Klakow, Nabende, Chang, Gwadabe et~al.}]{adelani-etal-2022-thousand}
David Adelani, Jesujoba Alabi, Angela Fan, Julia Kreutzer, Xiaoyu Shen, Machel Reid, Dana Ruiter, Dietrich Klakow, Peter Nabende, Ernie Chang, Tajuddeen Gwadabe, et~al. 2022{\natexlab{a}}.
\newblock \href {https://doi.org/10.18653/v1/2022.naacl-main.223} {A few thousand translations go a long way! leveraging pre-trained models for {A}frican news translation}.
\newblock In \emph{Proceedings of the 2022 Conference of the North American Chapter of the Association for Computational Linguistics: Human Language Technologies}, pages 3053--3070, Seattle, United States. Association for Computational Linguistics.

\bibitem[{Adelani et~al.(2024{\natexlab{a}})Adelani, Liu, Shen, Vassilyev, Alabi, Mao, Gao, and Lee}]{adelani-etal-2024-sib}
David Adelani, Hannah Liu, Xiaoyu Shen, Nikita Vassilyev, Jesujoba Alabi, Yanke Mao, Haonan Gao, and En-Shiun Lee. 2024{\natexlab{a}}.
\newblock \href {https://aclanthology.org/2024.eacl-long.14} {{SIB}-200: A simple, inclusive, and big evaluation dataset for topic classification in 200+ languages and dialects}.
\newblock In \emph{Proceedings of the 18th Conference of the European Chapter of the Association for Computational Linguistics (Volume 1: Long Papers)}, pages 226--245, St. Julian{'}s, Malta. Association for Computational Linguistics.

\bibitem[{Adelani et~al.(2022{\natexlab{b}})Adelani, Neubig, Ruder, Rijhwani, Beukman, Palen-Michel, Lignos, Alabi, Muhammad, Nabende, Dione et~al.}]{adelani-etal-2022-masakhaner}
David Adelani, Graham Neubig, Sebastian Ruder, Shruti Rijhwani, Michael Beukman, Chester Palen-Michel, Constantine Lignos, Jesujoba Alabi, Shamsuddeen Muhammad, Peter Nabende, Cheikh M.~Bamba Dione, et~al. 2022{\natexlab{b}}.
\newblock \href {https://doi.org/10.18653/v1/2022.emnlp-main.298} {{M}asakha{NER} 2.0: {A}frica-centric transfer learning for named entity recognition}.
\newblock In \emph{Proceedings of the 2022 Conference on Empirical Methods in Natural Language Processing}, pages 4488--4508, Abu Dhabi, United Arab Emirates. Association for Computational Linguistics.

\bibitem[{Adelani et~al.(2021)Adelani, Abbott, Neubig, D{'}souza, Kreutzer, Lignos, Palen-Michel, Buzaaba, Rijhwani, Ruder, Mayhew, Azime, Muhammad, Emezue, Nakatumba-Nabende, Ogayo, Anuoluwapo et~al.}]{adelani-etal-2021-masakhaner}
David~Ifeoluwa Adelani, Jade Abbott, Graham Neubig, Daniel D{'}souza, Julia Kreutzer, Constantine Lignos, Chester Palen-Michel, Happy Buzaaba, Shruti Rijhwani, Sebastian Ruder, Stephen Mayhew, Israel~Abebe Azime, Shamsuddeen~H. Muhammad, Chris~Chinenye Emezue, Joyce Nakatumba-Nabende, Perez Ogayo, Aremu Anuoluwapo, et~al. 2021.
\newblock \href {https://doi.org/10.1162/tacl_a_00416} {{M}asakha{NER}: Named entity recognition for {A}frican languages}.
\newblock \emph{Transactions of the Association for Computational Linguistics}, 9:1116--1131.

\bibitem[{Adelani et~al.(2023)Adelani, Masiak, Azime, Alabi, Tonja, Mwase, Ogundepo, Dossou, Oladipo et~al.}]{adelani2023masakhanews}
David~Ifeoluwa Adelani, Marek Masiak, Israel~Abebe Azime, Jesujoba Alabi, Atnafu~Lambebo Tonja, Christine Mwase, Odunayo Ogundepo, Bonaventure F.~P. Dossou, Akintunde Oladipo, et~al. 2023.
\newblock \href {https://arxiv.org/abs/2304.09972} {Masakhanews: News topic classification for african languages}.
\newblock \emph{Preprint}, arXiv:2304.09972.

\bibitem[{Adelani et~al.(2024{\natexlab{b}})Adelani, Ojo, Azime, Jian, Alabi, He, Ochieng, Hooker, Bukula, Lee, Chukwuneke, Buzaaba et~al.}]{Adelani2024IrokoBenchAN}
David~Ifeoluwa Adelani, Jessica Ojo, Israel~Abebe Azime, Zhuang~Yun Jian, Jesujoba~Oluwadara Alabi, Xuanli He, Millicent Ochieng, Sara Hooker, Andiswa Bukula, En-Shiun~Annie Lee, Chiamaka Chukwuneke, Happy Buzaaba, et~al. 2024{\natexlab{b}}.
\newblock \href {https://api.semanticscholar.org/CorpusID:270258352} {Irokobench: A new benchmark for african languages in the age of large language models}.
\newblock \emph{ArXiv}, abs/2406.03368.

\bibitem[{Ahuja et~al.(2023{\natexlab{a}})Ahuja, Diddee, Hada, Ochieng, Ramesh, Jain, Nambi, Ganu, Segal, Ahmed, Bali, and Sitaram}]{ahuja-etal-2023-mega}
Kabir Ahuja, Harshita Diddee, Rishav Hada, Millicent Ochieng, Krithika Ramesh, Prachi Jain, Akshay Nambi, Tanuja Ganu, Sameer Segal, Mohamed Ahmed, Kalika Bali, and Sunayana Sitaram. 2023{\natexlab{a}}.
\newblock \href {https://doi.org/10.18653/v1/2023.emnlp-main.258} {{MEGA}: Multilingual evaluation of generative {AI}}.
\newblock In \emph{Proceedings of the 2023 Conference on Empirical Methods in Natural Language Processing}, pages 4232--4267, Singapore. Association for Computational Linguistics.

\bibitem[{Ahuja et~al.(2024)Ahuja, Aggarwal, Gumma, Watts, Sathe, Ochieng, Hada, Jain, Ahmed, Bali, and Sitaram}]{ahuja-etal-2024-megaverse}
Sanchit Ahuja, Divyanshu Aggarwal, Varun Gumma, Ishaan Watts, Ashutosh Sathe, Millicent Ochieng, Rishav Hada, Prachi Jain, Mohamed Ahmed, Kalika Bali, and Sunayana Sitaram. 2024.
\newblock \href {https://doi.org/10.18653/v1/2024.naacl-long.143} {{MEGAVERSE}: Benchmarking large language models across languages, modalities, models and tasks}.
\newblock In \emph{Proceedings of the 2024 Conference of the North American Chapter of the Association for Computational Linguistics: Human Language Technologies (Volume 1: Long Papers)}, pages 2598--2637, Mexico City, Mexico. Association for Computational Linguistics.

\bibitem[{Ahuja et~al.(2023{\natexlab{b}})Ahuja, Aggarwal, Gumma, Watts, Sathe, Ochieng, Hada, Jain, Axmed, Bali, and Sitaram}]{ahuja2023megaverse}
Sanchit Ahuja, Divyanshu Aggarwal, Varun Gumma, Ishaan Watts, Ashutosh Sathe, Millicent Ochieng, Rishav Hada, Prachi Jain, Maxamed Axmed, Kalika Bali, and Sunayana Sitaram. 2023{\natexlab{b}}.
\newblock \href {https://arxiv.org/abs/2311.07463} {Megaverse: Benchmarking large language models across languages, modalities, models and tasks}.
\newblock \emph{Preprint}, arXiv:2311.07463.

\bibitem[{AI(2024)}]{Llama3.1Meta2024}
Meta AI. 2024.
\newblock \href {https://ai.meta.com/blog/meta-llama-3-1/} {Meta ai announces llama 3.1}.
\newblock Accessed: Feb 1, 2025.

\bibitem[{Akera et~al.(2022)Akera, Mukiibi, Naggayi, Babirye, Owomugisha, Nsumba, Nakatumba-Nabende, Bainomugisha, Mwebaze, and Quinn}]{akera2022machine}
Benjamin Akera, Jonathan Mukiibi, Lydia~Sanyu Naggayi, Claire Babirye, Isaac Owomugisha, Solomon Nsumba, Joyce Nakatumba-Nabende, Engineer Bainomugisha, Ernest Mwebaze, and John Quinn. 2022.
\newblock \href {https://openreview.net/forum?id=BK-z5qzEU-9} {Machine translation for african languages: Community creation of datasets and models in uganda}.
\newblock In \emph{3rd Workshop on African Natural Language Processing}.

\bibitem[{Alabi et~al.(2022)Alabi, Adelani, Mosbach, and Klakow}]{alabi-etal-2022-adapting}
Jesujoba~O. Alabi, David~Ifeoluwa Adelani, Marius Mosbach, and Dietrich Klakow. 2022.
\newblock \href {https://aclanthology.org/2022.coling-1.382} {Adapting pre-trained language models to {A}frican languages via multilingual adaptive fine-tuning}.
\newblock In \emph{Proceedings of the 29th International Conference on Computational Linguistics}, pages 4336--4349, Gyeongju, Republic of Korea. International Committee on Computational Linguistics.

\bibitem[{Anil et~al.(2023)Anil, Dai, Firat, Johnson, Lepikhin, Passos, Shakeri, Taropa, Bailey, Chen, Chu, Clark, Shafey, Huang, Meier-Hellstern, Mishra, Moreira, Omernick, Robinson, Ruder, Tay, Xiao, Xu, Zhang, Abrego, Ahn, Austin, Barham, Botha, Bradbury, Brahma, Brooks, Catasta, Cheng, Cherry, Choquette-Choo, Chowdhery, Crepy, Dave, Dehghani, Dev, Devlin, Díaz, Du, Dyer, Feinberg, Feng, Fienber, Freitag, Garcia, Gehrmann, Gonzalez, Gur-Ari, Hand, Hashemi, Hou, Howland, Hu, Hui, Hurwitz, Isard, Ittycheriah, Jagielski, Jia, Kenealy, Krikun, Kudugunta, Lan, Lee, Lee, Li, Li, Li, Li, Li, Lim, Lin, Liu, Liu, Maggioni, Mahendru, Maynez, Misra, Moussalem, Nado, Nham, Ni, Nystrom, Parrish, Pellat, Polacek, Polozov, Pope, Qiao, Reif, Richter, Riley, Ros, Roy, Saeta, Samuel, Shelby, Slone, Smilkov, So, Sohn, Tokumine, Valter, Vasudevan, Vodrahalli, Wang, Wang, Wang, Wang, Wieting, Wu, Xu, Xu, Xue, Yin, Yu, Zhang, Zheng, Zheng, Zhou, Zhou, Petrov, and Wu}]{anil2023palm}
Rohan Anil, Andrew~M. Dai, Orhan Firat, Melvin Johnson, Dmitry Lepikhin, Alexandre Passos, Siamak Shakeri, Emanuel Taropa, Paige Bailey, Zhifeng Chen, Eric Chu, Jonathan~H. Clark, Laurent~El Shafey, Yanping Huang, Kathy Meier-Hellstern, Gaurav Mishra, Erica Moreira, Mark Omernick, Kevin Robinson, Sebastian Ruder, Yi~Tay, Kefan Xiao, Yuanzhong Xu, Yujing Zhang, Gustavo~Hernandez Abrego, Junwhan Ahn, Jacob Austin, Paul Barham, Jan Botha, James Bradbury, Siddhartha Brahma, Kevin Brooks, Michele Catasta, Yong Cheng, Colin Cherry, Christopher~A. Choquette-Choo, Aakanksha Chowdhery, Clément Crepy, Shachi Dave, Mostafa Dehghani, Sunipa Dev, Jacob Devlin, Mark Díaz, Nan Du, Ethan Dyer, Vlad Feinberg, Fangxiaoyu Feng, Vlad Fienber, Markus Freitag, Xavier Garcia, Sebastian Gehrmann, Lucas Gonzalez, Guy Gur-Ari, Steven Hand, Hadi Hashemi, Le~Hou, Joshua Howland, Andrea Hu, Jeffrey Hui, Jeremy Hurwitz, Michael Isard, Abe Ittycheriah, Matthew Jagielski, Wenhao Jia, Kathleen Kenealy, Maxim Krikun, Sneha Kudugunta, Chang
  Lan, Katherine Lee, Benjamin Lee, Eric Li, Music Li, Wei Li, YaGuang Li, Jian Li, Hyeontaek Lim, Hanzhao Lin, Zhongtao Liu, Frederick Liu, Marcello Maggioni, Aroma Mahendru, Joshua Maynez, Vedant Misra, Maysam Moussalem, Zachary Nado, John Nham, Eric Ni, Andrew Nystrom, Alicia Parrish, Marie Pellat, Martin Polacek, Alex Polozov, Reiner Pope, Siyuan Qiao, Emily Reif, Bryan Richter, Parker Riley, Alex~Castro Ros, Aurko Roy, Brennan Saeta, Rajkumar Samuel, Renee Shelby, Ambrose Slone, Daniel Smilkov, David~R. So, Daniel Sohn, Simon Tokumine, Dasha Valter, Vijay Vasudevan, Kiran Vodrahalli, Xuezhi Wang, Pidong Wang, Zirui Wang, Tao Wang, John Wieting, Yuhuai Wu, Kelvin Xu, Yunhan Xu, Linting Xue, Pengcheng Yin, Jiahui Yu, Qiao Zhang, Steven Zheng, Ce~Zheng, Weikang Zhou, Denny Zhou, Slav Petrov, and Yonghui Wu. 2023.
\newblock \href {https://arxiv.org/abs/2305.10403} {Palm 2 technical report}.
\newblock \emph{Preprint}, arXiv:2305.10403.

\bibitem[{Aremu et~al.(2024)Aremu, Alabi, Abolade, Aguobi, Muhammad, and Adelani}]{aremu2024naijarc}
Anuoluwapo Aremu, Jesujoba~Oluwadara Alabi, Daud Abolade, Nkechinyere~Faith Aguobi, Shamsuddeen~Hassan Muhammad, and David~Ifeoluwa Adelani. 2024.
\newblock \href {https://openreview.net/forum?id=xzi7gDeq5r} {Naija{RC}: A multi-choice reading comprehension dataset for nigerian languages}.
\newblock In \emph{5th Workshop on African Natural Language Processing}.

\bibitem[{Bandarkar et~al.(2024)Bandarkar, Liang, Muller, Artetxe, Shukla, Husa, Goyal, Krishnan, Zettlemoyer, and Khabsa}]{bandarkar-etal-2024-belebele}
Lucas Bandarkar, Davis Liang, Benjamin Muller, Mikel Artetxe, Satya~Narayan Shukla, Donald Husa, Naman Goyal, Abhinandan Krishnan, Luke Zettlemoyer, and Madian Khabsa. 2024.
\newblock \href {https://doi.org/10.18653/v1/2024.acl-long.44} {The belebele benchmark: a parallel reading comprehension dataset in 122 language variants}.
\newblock In \emph{Proceedings of the 62nd Annual Meeting of the Association for Computational Linguistics (Volume 1: Long Papers)}, pages 749--775, Bangkok, Thailand. Association for Computational Linguistics.

\bibitem[{Bayes et~al.(2024)Bayes, Azime, Alabi, Kgomo, Eloundou, Proehl, Chen, Khadir, Etori, Muhammad, Mpanza, Thete, Klakow, and Adelani}]{Bayes2024UhuraAB}
Edward Bayes, Israel~Abebe Azime, Jesujoba~Oluwadara Alabi, Jonas Kgomo, Tyna Eloundou, Elizabeth Proehl, Kai Chen, Imaan Khadir, Naome~A. Etori, Shamsuddeen~Hassan Muhammad, Choice Mpanza, Igneciah~Pocia Thete, Dietrich Klakow, and David~Ifeoluwa Adelani. 2024.
\newblock \href {https://api.semanticscholar.org/CorpusID:274436231} {Uhura: A benchmark for evaluating scientific question answering and truthfulness in low-resource african languages}.
\newblock \emph{ArXiv}, abs/2412.00948.

\bibitem[{bench authors(2023)}]{srivastava2023beyond}
BIG bench authors. 2023.
\newblock \href {https://openreview.net/forum?id=uyTL5Bvosj} {Beyond the imitation game: Quantifying and extrapolating the capabilities of language models}.
\newblock \emph{Transactions on Machine Learning Research}.

\bibitem[{Biderman et~al.(2024)Biderman, Schoelkopf, Sutawika, Gao, Tow, Abbasi, Aji, Ammanamanchi, Black, Clive, DiPofi, Etxaniz, Fattori, Forde, Foster, Hsu, Jaiswal, Lee, Li, Lovering, Muennighoff, Pavlick, Phang, Skowron, Tan, Tang, Wang, Winata, Yvon, and Zou}]{biderman2024lmevaluation}
Stella Biderman, Hailey Schoelkopf, Lintang Sutawika, Leo Gao, Jonathan Tow, Baber Abbasi, Alham~Fikri Aji, Pawan~Sasanka Ammanamanchi, Sidney Black, Jordan Clive, Anthony DiPofi, Julen Etxaniz, Benjamin Fattori, Jessica~Zosa Forde, Charles Foster, Jeffrey Hsu, Mimansa Jaiswal, Wilson~Y. Lee, Haonan Li, Charles Lovering, Niklas Muennighoff, Ellie Pavlick, Jason Phang, Aviya Skowron, Samson Tan, Xiangru Tang, Kevin~A. Wang, Genta~Indra Winata, François Yvon, and Andy Zou. 2024.
\newblock \href {https://arxiv.org/abs/2405.14782} {Lessons from the trenches on reproducible evaluation of language models}.

\bibitem[{Brown et~al.(2020)Brown, Mann, Ryder, Subbiah, Kaplan, Dhariwal, Neelakantan, Shyam, Sastry, Askell, Agarwal, Herbert-Voss, Krueger, Henighan, Child, Ramesh, Ziegler, Wu, Winter, Hesse, Chen, Sigler, Litwin, Gray, Chess, Clark, Berner, McCandlish, Radford, Sutskever, and Amodei}]{NEURIPS2020_1457c0d6}
Tom Brown, Benjamin Mann, Nick Ryder, Melanie Subbiah, Jared~D Kaplan, Prafulla Dhariwal, Arvind Neelakantan, Pranav Shyam, Girish Sastry, Amanda Askell, Sandhini Agarwal, Ariel Herbert-Voss, Gretchen Krueger, Tom Henighan, Rewon Child, Aditya Ramesh, Daniel Ziegler, Jeffrey Wu, Clemens Winter, Chris Hesse, Mark Chen, Eric Sigler, Mateusz Litwin, Scott Gray, Benjamin Chess, Jack Clark, Christopher Berner, Sam McCandlish, Alec Radford, Ilya Sutskever, and Dario Amodei. 2020.
\newblock \href {https://proceedings.neurips.cc/paper/2020/file/1457c0d6bfcb4967418bfb8ac142f64a-Paper.pdf} {Language models are few-shot learners}.
\newblock In \emph{Advances in Neural Information Processing Systems}, volume~33, pages 1877--1901. Curran Associates, Inc.

\bibitem[{Buzaaba et~al.(2025)Buzaaba, Wettig, Adelani, and Fellbaum}]{Buzaaba2025LughaLlamaAL}
Happy Buzaaba, Alexander Wettig, David~Ifeoluwa Adelani, and Christiane Fellbaum. 2025.
\newblock \href {https://api.semanticscholar.org/CorpusID:277634577} {Lugha-llama: Adapting large language models for african languages}.
\newblock \emph{ArXiv}, abs/2504.06536.

\bibitem[{Chiang et~al.(2024)Chiang, Zheng, Sheng, Angelopoulos, Li, Li, Zhu, Zhang, Jordan, Gonzalez, and Stoica}]{10.5555/3692070.3692401}
Wei-Lin Chiang, Lianmin Zheng, Ying Sheng, Anastasios~N. Angelopoulos, Tianle Li, Dacheng Li, Banghua Zhu, Hao Zhang, Michael~I. Jordan, Joseph~E. Gonzalez, and Ion Stoica. 2024.
\newblock Chatbot arena: an open platform for evaluating llms by human preference.
\newblock In \emph{Proceedings of the 41st International Conference on Machine Learning}, ICML'24. JMLR.org.

\bibitem[{Chowdhery et~al.(2022)Chowdhery, Narang, Devlin, Bosma, Mishra, Roberts, Barham, Chung, Sutton, Gehrmann et~al.}]{chowdhery2022palm}
Aakanksha Chowdhery, Sharan Narang, Jacob Devlin, Maarten Bosma, Gaurav Mishra, Adam Roberts, Paul Barham, Hyung~Won Chung, Charles Sutton, Sebastian Gehrmann, et~al. 2022.
\newblock Palm: Scaling language modeling with pathways.
\newblock \emph{arXiv preprint arXiv:2204.02311}.

\bibitem[{Chung et~al.(2022)Chung, Hou, Longpre, Zoph, Tay, Fedus, Li, Wang, Dehghani, Brahma, Webson, Gu, Dai, Suzgun, Chen, Chowdhery, Castro-Ros, Pellat, Robinson, Valter, Narang, Mishra, Yu, Zhao, Huang, Dai, Yu, Petrov, Chi, Dean, Devlin, Roberts, Zhou, Le, and Wei}]{scaling-iflm}
Hyung~Won Chung, Le~Hou, Shayne Longpre, Barret Zoph, Yi~Tay, William Fedus, Yunxuan Li, Xuezhi Wang, Mostafa Dehghani, Siddhartha Brahma, Albert Webson, Shixiang~Shane Gu, Zhuyun Dai, Mirac Suzgun, Xinyun Chen, Aakanksha Chowdhery, Alex Castro-Ros, Marie Pellat, Kevin Robinson, Dasha Valter, Sharan Narang, Gaurav Mishra, Adams Yu, Vincent Zhao, Yanping Huang, Andrew Dai, Hongkun Yu, Slav Petrov, Ed~H. Chi, Jeff Dean, Jacob Devlin, Adam Roberts, Denny Zhou, Quoc~V. Le, and Jason Wei. 2022.
\newblock \href {https://doi.org/10.48550/ARXIV.2210.11416} {Scaling instruction-finetuned language models}.
\newblock \emph{arXiv preprint}.

\bibitem[{Dione et~al.(2023)Dione, Adelani, Nabende, Alabi, Sindane, Buzaaba, Muhammad, Emezue, Ogayo, Aremu, Gitau, Mbaye, Mukiibi, Sibanda, Dossou, Bukula, Mabuya, Tapo, Munkoh-Buabeng, Memdjokam~Koagne, Ouoba~Kabore, Taylor, Kalipe, Macucwa, Marivate, Gwadabe, Elvis, Onyenwe, Atindogbe, Adelani, Akinade, Samuel, Nahimana, Musabeyezu, Niyomutabazi, Chimhenga, Gotosa, Mizha, Agbolo, Traore, Uchechukwu, Yusuf, Abdullahi, and Klakow}]{dione-etal-2023-masakhapos}
Cheikh M.~Bamba Dione, David~Ifeoluwa Adelani, Peter Nabende, Jesujoba Alabi, Thapelo Sindane, Happy Buzaaba, Shamsuddeen~Hassan Muhammad, Chris~Chinenye Emezue, Perez Ogayo, Anuoluwapo Aremu, Catherine Gitau, Derguene Mbaye, Jonathan Mukiibi, Blessing Sibanda, Bonaventure F.~P. Dossou, Andiswa Bukula, Rooweither Mabuya, Allahsera~Auguste Tapo, Edwin Munkoh-Buabeng, Victoire Memdjokam~Koagne, Fatoumata Ouoba~Kabore, Amelia Taylor, Godson Kalipe, Tebogo Macucwa, Vukosi Marivate, Tajuddeen Gwadabe, Mboning~Tchiaze Elvis, Ikechukwu Onyenwe, Gratien Atindogbe, Tolulope Adelani, Idris Akinade, Olanrewaju Samuel, Marien Nahimana, Th{\'e}og{\`e}ne Musabeyezu, Emile Niyomutabazi, Ester Chimhenga, Kudzai Gotosa, Patrick Mizha, Apelete Agbolo, Seydou Traore, Chinedu Uchechukwu, Aliyu Yusuf, Muhammad Abdullahi, and Dietrich Klakow. 2023.
\newblock \href {https://doi.org/10.18653/v1/2023.acl-long.609} {{M}asakha{POS}: Part-of-speech tagging for typologically diverse {A}frican languages}.
\newblock In \emph{Proceedings of the 61st Annual Meeting of the Association for Computational Linguistics (Volume 1: Long Papers)}, pages 10883--10900, Toronto, Canada. Association for Computational Linguistics.

\bibitem[{Dubey et~al.(2024)Dubey, Jauhri, Pandey, Kadian, Al-Dahle, Letman, Mathur, Schelten, Yang, Fan, Goyal, Hartshorn, Yang, Mitra, Sravankumar, Korenev, and et~al.}]{Dubey2024TheL3}
Abhimanyu Dubey, Abhinav Jauhri, Abhinav Pandey, Abhishek Kadian, Ahmad Al-Dahle, Aiesha Letman, Akhil Mathur, Alan Schelten, Amy Yang, Angela Fan, Anirudh Goyal, Anthony~S. Hartshorn, Aobo Yang, Archi Mitra, Archie Sravankumar, Artem Korenev, and et~al. 2024.
\newblock \href {https://api.semanticscholar.org/CorpusID:271571434} {The llama 3 herd of models}.
\newblock \emph{ArXiv}, abs/2407.21783.

\bibitem[{Federmann et~al.(2022)Federmann, Kocmi, and Xin}]{federmann-etal-2022-ntrex}
Christian Federmann, Tom Kocmi, and Ying Xin. 2022.
\newblock \href {https://aclanthology.org/2022.sumeval-1.4} {{NTREX}-128 {--} news test references for {MT} evaluation of 128 languages}.
\newblock In \emph{Proceedings of the First Workshop on Scaling Up Multilingual Evaluation}, pages 21--24, Online. Association for Computational Linguistics.

\bibitem[{Fourrier et~al.(2023)Fourrier, Habib, Kydlíček, Wolf, and Tunstall}]{lighteval}
Clémentine Fourrier, Nathan Habib, Hynek Kydlíček, Thomas Wolf, and Lewis Tunstall. 2023.
\newblock \href {https://github.com/huggingface/lighteval} {Lighteval: A lightweight framework for llm evaluation}.

\bibitem[{Fourrier et~al.(2024)Fourrier, Habib, Lozovskaya, Szafer, and Wolf}]{open-llm-leaderboard-v2}
Clémentine Fourrier, Nathan Habib, Alina Lozovskaya, Konrad Szafer, and Thomas Wolf. 2024.
\newblock Open llm leaderboard v2.
\newblock \url{https://huggingface.co/spaces/open-llm-leaderboard/open_llm_leaderboard}.

\bibitem[{Gao et~al.(2024)Gao, Tow, Abbasi, Biderman, Black, DiPofi, Foster, Golding, Hsu, Le~Noac'h, Li, McDonell, Muennighoff, Ociepa, Phang, Reynolds, Schoelkopf, Skowron, Sutawika, Tang, Thite, Wang, Wang, and Zou}]{eval-harness}
Leo Gao, Jonathan Tow, Baber Abbasi, Stella Biderman, Sid Black, Anthony DiPofi, Charles Foster, Laurence Golding, Jeffrey Hsu, Alain Le~Noac'h, Haonan Li, Kyle McDonell, Niklas Muennighoff, Chris Ociepa, Jason Phang, Laria Reynolds, Hailey Schoelkopf, Aviya Skowron, Lintang Sutawika, Eric Tang, Anish Thite, Ben Wang, Kevin Wang, and Andy Zou. 2024.
\newblock \href {https://doi.org/10.5281/zenodo.12608602} {A framework for few-shot language model evaluation}.

\bibitem[{Goyal et~al.(2022)Goyal, Gao, Chaudhary, Chen, Wenzek, Ju, Krishnan, Ranzato, Guzm{\'a}n, and Fan}]{goyal-etal-2022-flores}
Naman Goyal, Cynthia Gao, Vishrav Chaudhary, Peng-Jen Chen, Guillaume Wenzek, Da~Ju, Sanjana Krishnan, Marc{'}Aurelio Ranzato, Francisco Guzm{\'a}n, and Angela Fan. 2022.
\newblock \href {https://doi.org/10.1162/tacl_a_00474} {The {F}lores-101 evaluation benchmark for low-resource and multilingual machine translation}.
\newblock \emph{Transactions of the Association for Computational Linguistics}, 10:522--538.

\bibitem[{Hasan et~al.(2021)Hasan, Bhattacharjee, Islam, Mubasshir, Li, Kang, Rahman, and Shahriyar}]{hasan-etal-2021-xl}
Tahmid Hasan, Abhik Bhattacharjee, Md.~Saiful Islam, Kazi Mubasshir, Yuan-Fang Li, Yong-Bin Kang, M.~Sohel Rahman, and Rifat Shahriyar. 2021.
\newblock \href {https://aclanthology.org/2021.findings-acl.413} {{XL}-sum: Large-scale multilingual abstractive summarization for 44 languages}.
\newblock In \emph{Findings of the Association for Computational Linguistics: ACL-IJCNLP 2021}, pages 4693--4703, Online. Association for Computational Linguistics.

\bibitem[{Health et~al.(2024)Health, Stanslaus, Jay, Sathy~Rajasekharan, Francesco, Mfoniso, and Ellen}]{AfroLlamaV1}
Jacaranda Health, Mwongela Stanslaus, Patel Jay, Lusiji Sathy~Rajasekharan, Lyvia, Piccino Francesco, Ukwak Mfoniso, and Sebastian Ellen. 2024.
\newblock \href {https://huggingface.co/Jacaranda/AfroLlama_V1} {Afrollama v1: An instruction-tuned llama model for african languages}.
\newblock Accessed: Feb 12, 2025.

\bibitem[{Hendrycks et~al.(2021)Hendrycks, Burns, Kadavath, Arora, Basart, Tang, Song, and Steinhardt}]{hendrycks2021measuringmathematicalproblemsolving}
Dan Hendrycks, Collin Burns, Saurav Kadavath, Akul Arora, Steven Basart, Eric Tang, Dawn Song, and Jacob Steinhardt. 2021.
\newblock \href {https://arxiv.org/abs/2103.03874} {Measuring mathematical problem solving with the math dataset}.
\newblock \emph{Preprint}, arXiv:2103.03874.

\bibitem[{Jiang et~al.(2023)Jiang, Sablayrolles, Mensch, Bamford, Chaplot, de~Las~Casas, Bressand, Lengyel, Lample, Saulnier, Lavaud, Lachaux, Stock, Scao, Lavril, Wang, Lacroix, and Sayed}]{Jiang2023Mistral7}
Albert~Qiaochu Jiang, Alexandre Sablayrolles, Arthur Mensch, Chris Bamford, Devendra~Singh Chaplot, Diego de~Las~Casas, Florian Bressand, Gianna Lengyel, Guillaume Lample, Lucile Saulnier, L'elio~Renard Lavaud, Marie-Anne Lachaux, Pierre Stock, Teven~Le Scao, Thibaut Lavril, Thomas Wang, Timoth{\'e}e Lacroix, and William~El Sayed. 2023.
\newblock \href {https://api.semanticscholar.org/CorpusID:263830494} {Mistral 7b}.
\newblock \emph{ArXiv}, abs/2310.06825.

\bibitem[{Joshi et~al.(2020)Joshi, Santy, Budhiraja, Bali, and Choudhury}]{joshi-etal-2020-state}
Pratik Joshi, Sebastin Santy, Amar Budhiraja, Kalika Bali, and Monojit Choudhury. 2020.
\newblock \href {https://doi.org/10.18653/v1/2020.acl-main.560} {The state and fate of linguistic diversity and inclusion in the {NLP} world}.
\newblock In \emph{Proceedings of the 58th Annual Meeting of the Association for Computational Linguistics}, pages 6282--6293, Online. Association for Computational Linguistics.

\bibitem[{Kudugunta et~al.(2023)Kudugunta, Caswell, Zhang, Garcia, Xin, Kusupati, Stella, Bapna, and Firat}]{kudugunta2023madlad}
Sneha Kudugunta, Isaac~Rayburn Caswell, Biao Zhang, Xavier Garcia, Derrick Xin, Aditya Kusupati, Romi Stella, Ankur Bapna, and Orhan Firat. 2023.
\newblock \href {https://openreview.net/forum?id=Y45ZCxslFx} {{MADLAD}-400: A multilingual and document-level large audited dataset}.
\newblock In \emph{Thirty-seventh Conference on Neural Information Processing Systems Datasets and Benchmarks Track}.

\bibitem[{Lai et~al.(2023)Lai, Ngo, Pouran Ben~Veyseh, Man, Dernoncourt, Bui, and Nguyen}]{lai-etal-2023-chatgpt}
Viet Lai, Nghia Ngo, Amir Pouran Ben~Veyseh, Hieu Man, Franck Dernoncourt, Trung Bui, and Thien Nguyen. 2023.
\newblock \href {https://doi.org/10.18653/v1/2023.findings-emnlp.878} {{C}hat{GPT} beyond {E}nglish: Towards a comprehensive evaluation of large language models in multilingual learning}.
\newblock In \emph{Findings of the Association for Computational Linguistics: EMNLP 2023}, pages 13171--13189, Singapore. Association for Computational Linguistics.

\bibitem[{Liang et~al.(2023)Liang, Bommasani, Lee, Tsipras, Soylu, Yasunaga, Zhang, Narayanan, Wu, Kumar, Newman, Yuan, Yan, Zhang, Cosgrove, Manning, Re, Acosta-Navas, Hudson, Zelikman, Durmus, Ladhak, Rong, Ren, Yao, WANG, Santhanam, Orr, Zheng, Yuksekgonul, Suzgun, Kim, Guha, Chatterji, Khattab, Henderson, Huang, Chi, Xie, Santurkar, Ganguli, Hashimoto, Icard, Zhang, Chaudhary, Wang, Li, Mai, Zhang, and Koreeda}]{liang2023holistic}
Percy Liang, Rishi Bommasani, Tony Lee, Dimitris Tsipras, Dilara Soylu, Michihiro Yasunaga, Yian Zhang, Deepak Narayanan, Yuhuai Wu, Ananya Kumar, Benjamin Newman, Binhang Yuan, Bobby Yan, Ce~Zhang, Christian~Alexander Cosgrove, Christopher~D Manning, Christopher Re, Diana Acosta-Navas, Drew~Arad Hudson, Eric Zelikman, Esin Durmus, Faisal Ladhak, Frieda Rong, Hongyu Ren, Huaxiu Yao, Jue WANG, Keshav Santhanam, Laurel Orr, Lucia Zheng, Mert Yuksekgonul, Mirac Suzgun, Nathan Kim, Neel Guha, Niladri~S. Chatterji, Omar Khattab, Peter Henderson, Qian Huang, Ryan~Andrew Chi, Sang~Michael Xie, Shibani Santurkar, Surya Ganguli, Tatsunori Hashimoto, Thomas Icard, Tianyi Zhang, Vishrav Chaudhary, William Wang, Xuechen Li, Yifan Mai, Yuhui Zhang, and Yuta Koreeda. 2023.
\newblock \href {https://openreview.net/forum?id=iO4LZibEqW} {Holistic evaluation of language models}.
\newblock \emph{Transactions on Machine Learning Research}.
\newblock Featured Certification, Expert Certification.

\bibitem[{Lin et~al.(2021)Lin, Mihaylov, Artetxe, Wang, Chen, Simig, Ott, Goyal, Bhosale, Du, Pasunuru, Shleifer, Koura, Chaudhary, O'Horo, Wang, Zettlemoyer, Kozareva, Diab, Stoyanov, and Li}]{xglm}
Xi~Victoria Lin, Todor Mihaylov, Mikel Artetxe, Tianlu Wang, Shuohui Chen, Daniel Simig, Myle Ott, Naman Goyal, Shruti Bhosale, Jingfei Du, Ramakanth Pasunuru, Sam Shleifer, Punit~Singh Koura, Vishrav Chaudhary, Brian O'Horo, Jeff Wang, Luke Zettlemoyer, Zornitsa Kozareva, Mona~T. Diab, Veselin Stoyanov, and Xian Li. 2021.
\newblock \href {https://arxiv.org/abs/2112.10668} {Few-shot learning with multilingual language models}.
\newblock \emph{CoRR}, abs/2112.10668.

\bibitem[{Lu et~al.(2024)Lu, Zhu, Li, Qiao, and Yuan}]{lu-etal-2024-llamax}
Yinquan Lu, Wenhao Zhu, Lei Li, Yu~Qiao, and Fei Yuan. 2024.
\newblock \href {https://doi.org/10.18653/v1/2024.findings-emnlp.631} {{LL}a{MAX}: Scaling linguistic horizons of {LLM} by enhancing translation capabilities beyond 100 languages}.
\newblock In \emph{Findings of the Association for Computational Linguistics: EMNLP 2024}, pages 10748--10772, Miami, Florida, USA. Association for Computational Linguistics.

\bibitem[{Mesnard et~al.(2024)Mesnard, Hardin, Dadashi, Bhupatiraju, Pathak, Sifre, Rivi{\`e}re, Kale, Love, Tafti, and et~al.}]{Mesnard2024GemmaOM}
Gemma Team~Thomas Mesnard, Cassidy Hardin, Robert Dadashi, Surya Bhupatiraju, Shreya Pathak, L.~Sifre, Morgane Rivi{\`e}re, Mihir Kale, J~Christopher Love, Pouya~Dehghani Tafti, and et~al. 2024.
\newblock \href {https://api.semanticscholar.org/CorpusID:268379206} {Gemma: Open models based on gemini research and technology}.
\newblock \emph{ArXiv}, abs/2403.08295.

\bibitem[{Muhammad et~al.(2025)Muhammad, Abdulmumin, Ayele, Adelani, Ahmad, Aliyu, Onyango, Wanzare, Rutunda, Aliyu, Alemneh, Hourrane, Gebremichael, Ismail, Beloucif, Jibril, Bukula, Mabuya, Osei, Oppong, Belay, Guge, Asfaw, Chukwuneke, Röttger, Yimam, and Ousidhoum}]{muhammad2025afrihatemultilingualcollectionhate}
Shamsuddeen~Hassan Muhammad, Idris Abdulmumin, Abinew~Ali Ayele, David~Ifeoluwa Adelani, Ibrahim~Said Ahmad, Saminu~Mohammad Aliyu, Nelson~Odhiambo Onyango, Lilian D.~A. Wanzare, Samuel Rutunda, Lukman~Jibril Aliyu, Esubalew Alemneh, Oumaima Hourrane, Hagos~Tesfahun Gebremichael, Elyas~Abdi Ismail, Meriem Beloucif, Ebrahim~Chekol Jibril, Andiswa Bukula, Rooweither Mabuya, Salomey Osei, Abigail Oppong, Tadesse~Destaw Belay, Tadesse~Kebede Guge, Tesfa~Tegegne Asfaw, Chiamaka~Ijeoma Chukwuneke, Paul Röttger, Seid~Muhie Yimam, and Nedjma Ousidhoum. 2025.
\newblock \href {https://arxiv.org/abs/2501.08284} {Afrihate: A multilingual collection of hate speech and abusive language datasets for african languages}.
\newblock \emph{Preprint}, arXiv:2501.08284.

\bibitem[{Muhammad et~al.(2023)Muhammad, Abdulmumin, Ayele, Ousidhoum, Adelani, Yimam, Ahmad, Beloucif, Mohammad, Ruder, Hourrane, Brazdil, Ali, Davis, Osei, Bello, Ibrahim, Gwadabe, Rutunda, Belay, Messelle, Balcha, Chala, Gebremichael, Opoku, and Arthur}]{muhammad2023afrisenti}
Shamsuddeen~Hassan Muhammad, Idris Abdulmumin, Abinew~Ali Ayele, Nedjma Ousidhoum, David~Ifeoluwa Adelani, Seid~Muhie Yimam, Ibrahim~Sa'id Ahmad, Meriem Beloucif, Saif Mohammad, Sebastian Ruder, Oumaima Hourrane, Pavel Brazdil, Felermino Dário Mário~António Ali, Davis Davis, Salomey Osei, Bello~Shehu Bello, Falalu Ibrahim, Tajuddeen Gwadabe, Samuel Rutunda, Tadesse Belay, Wendimu~Baye Messelle, Hailu~Beshada Balcha, Sisay~Adugna Chala, Hagos~Tesfahun Gebremichael, Bernard Opoku, and Steven Arthur. 2023.
\newblock \href {https://doi.org/10.48550/arXiv.2302.08956} {{AfriSenti: A Twitter Sentiment Analysis Benchmark for African Languages}}.

\bibitem[{{NLLB Team} et~al.(2022){NLLB Team}, Costa-jussà, Cross, Çelebi, Elbayad, Heafield, Heffernan, Kalbassi, Lam, Licht, Maillard, Sun, Wang, Wenzek, Youngblood, Akula, Barrault, Mejia-Gonzalez, Hansanti, Hoffman, Jarrett, Sadagopan, Rowe, Spruit, Tran, Andrews, Ayan, Bhosale, Edunov, Fan, Gao, Goswami, Guzmán, Koehn, Mourachko, Ropers, Saleem, Schwenk, and Wang}]{nllb2022}
{NLLB Team}, Marta~R. Costa-jussà, James Cross, Onur Çelebi, Maha Elbayad, Kenneth Heafield, Kevin Heffernan, Elahe Kalbassi, Janice Lam, Daniel Licht, Jean Maillard, Anna Sun, Skyler Wang, Guillaume Wenzek, Al~Youngblood, Bapi Akula, Loic Barrault, Gabriel Mejia-Gonzalez, Prangthip Hansanti, John Hoffman, Semarley Jarrett, Kaushik~Ram Sadagopan, Dirk Rowe, Shannon Spruit, Chau Tran, Pierre Andrews, Necip~Fazil Ayan, Shruti Bhosale, Sergey Edunov, Angela Fan, Cynthia Gao, Vedanuj Goswami, Francisco Guzmán, Philipp Koehn, Alexandre Mourachko, Christophe Ropers, Safiyyah Saleem, Holger Schwenk, and Jeff Wang. 2022.
\newblock No language left behind: Scaling human-centered machine translation.

\bibitem[{Ogundepo et~al.(2023)Ogundepo, Gwadabe, Rivera, Clark, Ruder, Adelani, Dossou, DIOP, Sikasote, Hacheme, Buzaaba, Ezeani et~al.}]{ogundepo2023afriqa}
Odunayo Ogundepo, Tajuddeen~R. Gwadabe, Clara~E. Rivera, Jonathan~H. Clark, Sebastian Ruder, David~Ifeoluwa Adelani, Bonaventure F.~P. Dossou, Abdou~Aziz DIOP, Claytone Sikasote, Gilles Hacheme, Happy Buzaaba, Ignatius Ezeani, et~al. 2023.
\newblock \href {https://arxiv.org/abs/2305.06897} {Afriqa: Cross-lingual open-retrieval question answering for african languages}.
\newblock \emph{Preprint}, arXiv:2305.06897.

\bibitem[{Oladipo et~al.(2023)Oladipo, Adeyemi, Ahia, Owodunni, Ogundepo, Adelani, and Lin}]{oladipo-etal-2023-better}
Akintunde Oladipo, Mofetoluwa Adeyemi, Orevaoghene Ahia, Abraham~Toluwalase Owodunni, Odunayo Ogundepo, David~Ifeoluwa Adelani, and Jimmy Lin. 2023.
\newblock \href {https://doi.org/10.18653/v1/2023.emnlp-main.11} {Better quality pre-training data and t5 models for {A}frican languages}.
\newblock In \emph{Proceedings of the 2023 Conference on Empirical Methods in Natural Language Processing}, pages 158--168, Singapore. Association for Computational Linguistics.

\bibitem[{OpenAI(2023)}]{OpenAI2023GPT4TR}
OpenAI. 2023.
\newblock \href {https://api.semanticscholar.org/CorpusID:257532815} {Gpt-4 technical report}.
\newblock \emph{ArXiv}, abs/2303.08774.

\bibitem[{OpenAI(2024)}]{openai2024gpt4o}
OpenAI. 2024.
\newblock \href {https://cdn.openai.com/gpt-4o-system-card.pdf} {Gpt-4o system card}.
\newblock Accessed: February 13, 2025.

\bibitem[{Ouyang et~al.(2022)Ouyang, Wu, Jiang, Almeida, Wainwright, Mishkin, Zhang, Agarwal, Slama, Gray, Schulman, Hilton, Kelton, Miller, Simens, Askell, Welinder, Christiano, Leike, and Lowe}]{ouyang2022training}
Long Ouyang, Jeffrey Wu, Xu~Jiang, Diogo Almeida, Carroll Wainwright, Pamela Mishkin, Chong Zhang, Sandhini Agarwal, Katarina Slama, Alex Gray, John Schulman, Jacob Hilton, Fraser Kelton, Luke Miller, Maddie Simens, Amanda Askell, Peter Welinder, Paul Christiano, Jan Leike, and Ryan Lowe. 2022.
\newblock \href {https://openreview.net/forum?id=TG8KACxEON} {Training language models to follow instructions with human feedback}.
\newblock In \emph{Advances in Neural Information Processing Systems}.

\bibitem[{Rei et~al.(2020)Rei, Stewart, Farinha, and Lavie}]{rei-etal-2020-comet}
Ricardo Rei, Craig Stewart, Ana~C Farinha, and Alon Lavie. 2020.
\newblock \href {https://doi.org/10.18653/v1/2020.emnlp-main.213} {{COMET}: A neural framework for {MT} evaluation}.
\newblock In \emph{Proceedings of the 2020 Conference on Empirical Methods in Natural Language Processing (EMNLP)}, pages 2685--2702, Online. Association for Computational Linguistics.

\bibitem[{Reid et~al.(2024)Reid, Savinov, Teplyashin, Lepikhin, Lillicrap, Alayrac, Soricut, Lazaridou, Firat, Schrittwieser, Antonoglou, Anil, Borgeaud, Dai, Millican, Dyer, Glaese, Sottiaux, jamin Lee, Viola, Reynolds, Xu, and et~al.}]{Reid2024Gemini1U}
Machel Reid, Nikolay Savinov, Denis Teplyashin, Dmitry Lepikhin, Timothy~P. Lillicrap, Jean-Baptiste Alayrac, Radu Soricut, Angeliki Lazaridou, Orhan Firat, Julian Schrittwieser, Ioannis Antonoglou, Rohan Anil, Sebastian Borgeaud, Andrew~M. Dai, Katie Millican, Ethan Dyer, Mia Glaese, Thibault Sottiaux, Ben jamin Lee, Fabio Viola, Malcolm Reynolds, Yuanzhong Xu, and et~al. 2024.
\newblock \href {https://api.semanticscholar.org/CorpusID:268297180} {Gemini 1.5: Unlocking multimodal understanding across millions of tokens of context}.
\newblock \emph{ArXiv}, abs/2403.05530.

\bibitem[{Riviere et~al.(2024)Riviere, Pathak, Sessa, Hardin, Bhupatiraju, Hussenot, Mesnard, Shahriari, Ram'e, Ferret, Liu, Tafti, Friesen, Casbon, Ramos, Kumar, Lan, and et~al.}]{Riviere2024Gemma2I}
Gemma Team~Morgane Riviere, Shreya Pathak, Pier~Giuseppe Sessa, Cassidy Hardin, Surya Bhupatiraju, L'eonard Hussenot, Thomas Mesnard, Bobak Shahriari, Alexandre Ram'e, Johan Ferret, Peter Liu, Pouya~Dehghani Tafti, Abe Friesen, Michelle Casbon, Sabela Ramos, Ravin Kumar, Charline~Le Lan, and et~al. 2024.
\newblock \href {https://api.semanticscholar.org/CorpusID:270843326} {Gemma 2: Improving open language models at a practical size}.
\newblock \emph{ArXiv}, abs/2408.00118.

\bibitem[{Robinson et~al.(2023)Robinson, Ogayo, Mortensen, and Neubig}]{robinson-etal-2023-chatgpt}
Nathaniel Robinson, Perez Ogayo, David~R. Mortensen, and Graham Neubig. 2023.
\newblock \href {https://doi.org/10.18653/v1/2023.wmt-1.40} {{C}hat{GPT} {MT}: Competitive for high- (but not low-) resource languages}.
\newblock In \emph{Proceedings of the Eighth Conference on Machine Translation}, pages 392--418, Singapore. Association for Computational Linguistics.

\bibitem[{Ruder(2021)}]{ruder2021benchmarking}
Sebastian Ruder. 2021.
\newblock {Challenges and Opportunities in NLP Benchmarking}.
\newblock \url{http://ruder.io/nlp-benchmarking}.

\bibitem[{Ruder et~al.(2023)Ruder, Clark, Gutkin, Kale, Ma, Nicosia, Rijhwani, Riley, Sarr, Wang, Wieting, Gupta, Katanova, Kirov, Dickinson, Roark, Samanta, Tao, Adelani, Axelrod, Caswell, Cherry, Garrette, Ingle, Johnson, Panteleev, and Talukdar}]{Ruder2023XTREMEUPAU}
Sebastian Ruder, J.~Clark, Alexander Gutkin, Mihir Kale, Min Ma, Massimo Nicosia, Shruti Rijhwani, Parker Riley, Jean Michel~A. Sarr, Xinyi Wang, John Wieting, Nitish Gupta, Anna Katanova, Christo Kirov, Dana~L. Dickinson, Brian Roark, Bidisha Samanta, Connie Tao, David~Ifeoluwa Adelani, Vera Axelrod, Isaac Caswell, Colin Cherry, Dan Garrette, R.~Reeve Ingle, Melvin Johnson, Dmitry Panteleev, and Partha~Pratim Talukdar. 2023.
\newblock \href {https://api.semanticscholar.org/CorpusID:258833298} {Xtreme-up: A user-centric scarce-data benchmark for under-represented languages}.
\newblock \emph{ArXiv}, abs/2305.11938.

\bibitem[{Shi et~al.(2022)Shi, Suzgun, Freitag, Wang, Srivats, Vosoughi, Chung, Tay, Ruder, Zhou, Das, and Wei}]{wei-multilingual}
Freda Shi, Mirac Suzgun, Markus Freitag, Xuezhi Wang, Suraj Srivats, Soroush Vosoughi, Hyung~Won Chung, Yi~Tay, Sebastian Ruder, Denny Zhou, Dipanjan Das, and Jason Wei. 2022.
\newblock \href {https://doi.org/10.48550/ARXIV.2210.03057} {Language models are multilingual chain-of-thought reasoners}.
\newblock \emph{arXiv preprint}.

\bibitem[{Shode et~al.(2023)Shode, Adelani, Peng, and Feldman}]{shode-etal-2023-nollysenti}
Iyanuoluwa Shode, David~Ifeoluwa Adelani, JIng Peng, and Anna Feldman. 2023.
\newblock \href {https://doi.org/10.18653/v1/2023.acl-short.85} {{N}olly{S}enti: Leveraging transfer learning and machine translation for {N}igerian movie sentiment classification}.
\newblock In \emph{Proceedings of the 61st Annual Meeting of the Association for Computational Linguistics (Volume 2: Short Papers)}, pages 986--998, Toronto, Canada. Association for Computational Linguistics.

\bibitem[{Singh et~al.(2024)Singh, Romanou, Fourrier, Adelani, Ngui, Vila-Suero, Limkonchotiwat, Marchisio, Leong, Susanto, Ng, Longpre, Ko, Smith, Bosselut, Oh, Martins, Choshen, Ippolito, Ferrante, Fadaee, Ermiş, and Hooker}]{Singh2024GlobalMU}
Shivalika Singh, Angelika Romanou, Cl{\'e}mentine Fourrier, David~Ifeoluwa Adelani, Jian~Gang Ngui, Daniel Vila-Suero, Peerat Limkonchotiwat, Kelly Marchisio, Wei~Qi Leong, Yosephine Susanto, Raymond Ng, Shayne Longpre, Wei-Yin Ko, Madeline Smith, Antoine Bosselut, Alice Oh, Andr{\'e} F.~T. Martins, Leshem Choshen, Daphne Ippolito, Enzo Ferrante, Marzieh Fadaee, Beyza~Hilal Ermiş, and Sara Hooker. 2024.
\newblock \href {https://api.semanticscholar.org/CorpusID:274464561} {Global mmlu: Understanding and addressing cultural and linguistic biases in multilingual evaluation}.
\newblock \emph{ArXiv}, abs/2412.03304.

\bibitem[{Team et~al.(2023)Team, Anil, Borgeaud, Alayrac, Yu, Soricut, Schalkwyk, Dai, Hauth, Millican et~al.}]{team2023gemini}
Gemini Team, Rohan Anil, Sebastian Borgeaud, Jean-Baptiste Alayrac, Jiahui Yu, Radu Soricut, Johan Schalkwyk, Andrew~M Dai, Anja Hauth, Katie Millican, et~al. 2023.
\newblock Gemini: a family of highly capable multimodal models.
\newblock \emph{arXiv preprint arXiv:2312.11805}.

\bibitem[{Touvron et~al.(2023)Touvron, Martin, Stone, Albert, Almahairi, Babaei, Bashlykov, Batra, Bhargava, Bhosale, Bikel, Blecher, Ferrer, Chen, Cucurull, Esiobu, Fernandes, Fu, Fu, Fuller, Gao, Goswami, Goyal, Hartshorn, Hosseini, Hou, Inan, Kardas, Kerkez, Khabsa, Kloumann, Korenev, Koura, Lachaux, Lavril, Lee, Liskovich, Lu, Mao, Martinet, Mihaylov, Mishra, Molybog, Nie, Poulton, Reizenstein, Rungta, Saladi, Schelten, Silva, Smith, Subramanian, Tan, Tang, Taylor, Williams, Kuan, Xu, Yan, Zarov, Zhang, Fan, Kambadur, Narang, Rodriguez, Stojnic, Edunov, and Scialom}]{Touvron2023Llama2O}
Hugo Touvron, Louis Martin, Kevin~R. Stone, Peter Albert, Amjad Almahairi, Yasmine Babaei, Nikolay Bashlykov, Soumya Batra, Prajjwal Bhargava, Shruti Bhosale, Daniel~M. Bikel, Lukas Blecher, Cristian~Cant{\'o}n Ferrer, Moya Chen, Guillem Cucurull, David Esiobu, Jude Fernandes, Jeremy Fu, Wenyin Fu, Brian Fuller, Cynthia Gao, Vedanuj Goswami, Naman Goyal, Anthony~S. Hartshorn, Saghar Hosseini, Rui Hou, Hakan Inan, Marcin Kardas, Viktor Kerkez, Madian Khabsa, Isabel~M. Kloumann, A.~V. Korenev, Punit~Singh Koura, Marie-Anne Lachaux, Thibaut Lavril, Jenya Lee, Diana Liskovich, Yinghai Lu, Yuning Mao, Xavier Martinet, Todor Mihaylov, Pushkar Mishra, Igor Molybog, Yixin Nie, Andrew Poulton, Jeremy Reizenstein, Rashi Rungta, Kalyan Saladi, Alan Schelten, Ruan Silva, Eric~Michael Smith, R.~Subramanian, Xia Tan, Binh Tang, Ross Taylor, Adina Williams, Jian~Xiang Kuan, Puxin Xu, Zhengxu Yan, Iliyan Zarov, Yuchen Zhang, Angela Fan, Melanie Kambadur, Sharan Narang, Aurelien Rodriguez, Robert Stojnic, Sergey Edunov, and
  Thomas Scialom. 2023.
\newblock \href {https://api.semanticscholar.org/CorpusID:259950998} {Llama 2: Open foundation and fine-tuned chat models}.
\newblock \emph{ArXiv}, abs/2307.09288.

\bibitem[{Wang et~al.(2024)Wang, Adelani, Agrawal, Masiak, Rei, Briakou, Carpuat, He, Bourhim, Bukula, Mohamed, Olatoye, Adewumi, Mokayed, Mwase, Kimotho, Yuehgoh, Aremu, Ojo, Muhammad, Osei, Omotayo, Chukwuneke, Ogayo, Hourrane, El~Anigri, Ndolela, Mangwana, Mohamed, Ayinde, Awoyomi, Alkhaled, Al-azzawi, Etori, Ochieng, Siro, Kiragu, Muchiri, Kimotho, Wamba~Momo, Abolade, Ajao, Shode, Macharm, Iro, Abdullahi, Moore, Opoku, Akinjobi, Afolabi, Obiefuna, Ogbu, Ochieng{'}, Otiende, Mbonu, Toadoum~Sari, Lu, and Stenetorp}]{wang-etal-2024-afrimte}
Jiayi Wang, David~Ifeoluwa Adelani, Sweta Agrawal, Marek Masiak, Ricardo Rei, Eleftheria Briakou, Marine Carpuat, Xuanli He, Sofia Bourhim, Andiswa Bukula, Muhidin Mohamed, Temitayo Olatoye, Tosin Adewumi, Hamam Mokayed, Christine Mwase, Wangui Kimotho, Foutse Yuehgoh, Anuoluwapo Aremu, Jessica Ojo, Shamsuddeen~Hassan Muhammad, Salomey Osei, Abdul-Hakeem Omotayo, Chiamaka Chukwuneke, Perez Ogayo, Oumaima Hourrane, Salma El~Anigri, Lolwethu Ndolela, Thabiso Mangwana, Shafie~Abdi Mohamed, Hassan Ayinde, Oluwabusayo~Olufunke Awoyomi, Lama Alkhaled, Sana Al-azzawi, Naome~A. Etori, Millicent Ochieng, Clemencia Siro, Njoroge Kiragu, Eric Muchiri, Wangari Kimotho, Lyse~Naomi Wamba~Momo, Daud Abolade, Simbiat Ajao, Iyanuoluwa Shode, Ricky Macharm, Ruqayya~Nasir Iro, Saheed~S. Abdullahi, Stephen~E. Moore, Bernard Opoku, Zainab Akinjobi, Abeeb Afolabi, Nnaemeka Obiefuna, Onyekachi~Raphael Ogbu, Sam Ochieng{'}, Verrah~Akinyi Otiende, Chinedu~Emmanuel Mbonu, Sakayo Toadoum~Sari, Yao Lu, and Pontus Stenetorp. 2024.
\newblock \href {https://doi.org/10.18653/v1/2024.naacl-long.334} {{A}fri{MTE} and {A}fri{COMET}: Enhancing {COMET} to embrace under-resourced {A}frican languages}.
\newblock In \emph{Proceedings of the 2024 Conference of the North American Chapter of the Association for Computational Linguistics: Human Language Technologies (Volume 1: Long Papers)}, pages 5997--6023, Mexico City, Mexico. Association for Computational Linguistics.

\bibitem[{Xue et~al.(2021)Xue, Constant, Roberts, Kale, Al-Rfou, Siddhant, Barua, and Raffel}]{xue-etal-2021-mt5}
Linting Xue, Noah Constant, Adam Roberts, Mihir Kale, Rami Al-Rfou, Aditya Siddhant, Aditya Barua, and Colin Raffel. 2021.
\newblock \href {https://doi.org/10.18653/v1/2021.naacl-main.41} {m{T}5: A massively multilingual pre-trained text-to-text transformer}.
\newblock In \emph{Proceedings of the 2021 Conference of the North American Chapter of the Association for Computational Linguistics: Human Language Technologies}, pages 483--498, Online. Association for Computational Linguistics.

\bibitem[{Yu et~al.(2025)Yu, Alabi, Bukula, Jian, Lee, Guge, Azime, Buzaaba, Sibanda, Kalipe, Mukiibi, Kabenamualu, Setaka, Ndolela, Odu, Mabuya, Muhammad, Osei, Samb, Murage, Klakow, and Adelani}]{Yu2025INJONGOAM}
Hao Yu, Jesujoba~Oluwadara Alabi, Andiswa Bukula, Zhuang~Yun Jian, En-Shiun~Annie Lee, Tadesse~Kebede Guge, Israel~Abebe Azime, Happy Buzaaba, Blessing~K. Sibanda, Godson Kalipe, Jonathan Mukiibi, Salomon~Kabongo Kabenamualu, Mmasibidi Setaka, Lolwethu Ndolela, Nkiruka~Bridget Odu, Rooweither Mabuya, Shamsuddeen~Hassan Muhammad, Salomey Osei, Sokhar Samb, Juliet~W. Murage, Dietrich Klakow, and David~Ifeoluwa Adelani. 2025.
\newblock \href {https://api.semanticscholar.org/CorpusID:276394493} {Injongo: A multicultural intent detection and slot-filling dataset for 16 african languages}.
\newblock \emph{ArXiv}, abs/2502.09814.

\bibitem[{Zhang* et~al.(2020)Zhang*, Kishore*, Wu*, Weinberger, and Artzi}]{Zhang*2020BERTScore:}
Tianyi Zhang*, Varsha Kishore*, Felix Wu*, Kilian~Q. Weinberger, and Yoav Artzi. 2020.
\newblock \href {https://openreview.net/forum?id=SkeHuCVFDr} {Bertscore: Evaluating text generation with bert}.
\newblock In \emph{International Conference on Learning Representations}.

\bibitem[{Zhou et~al.(2023)Zhou, Lu, Mishra, Brahma, Basu, Luan, Zhou, and Hou}]{zhou2023instructionfollowingevaluationlargelanguage}
Jeffrey Zhou, Tianjian Lu, Swaroop Mishra, Siddhartha Brahma, Sujoy Basu, Yi~Luan, Denny Zhou, and Le~Hou. 2023.
\newblock \href {https://arxiv.org/abs/2311.07911} {Instruction-following evaluation for large language models}.
\newblock \emph{Preprint}, arXiv:2311.07911.

\bibitem[{Üstün et~al.(2024)Üstün, Aryabumi, Yong, Ko, D'souza, Onilude, Bhandari, Singh, Ooi, Kayid, Vargus, Blunsom, Longpre, Muennighoff, Fadaee, Kreutzer, and Hooker}]{ustün2024aya}
Ahmet Üstün, Viraat Aryabumi, Zheng-Xin Yong, Wei-Yin Ko, Daniel D'souza, Gbemileke Onilude, Neel Bhandari, Shivalika Singh, Hui-Lee Ooi, Amr Kayid, Freddie Vargus, Phil Blunsom, Shayne Longpre, Niklas Muennighoff, Marzieh Fadaee, Julia Kreutzer, and Sara Hooker. 2024.
\newblock \href {https://arxiv.org/abs/2402.07827} {Aya model: An instruction finetuned open-access multilingual language model}.
\newblock \emph{Preprint}, arXiv:2402.07827.

\end{thebibliography}
\appendix
\section{Task Based Results}
\label{sec:task_results}
We group tasks using similar evaluation metrics to analyze model performance systematically.
\begin{figure*}
    \centering
    \includegraphics[width=0.95\linewidth]{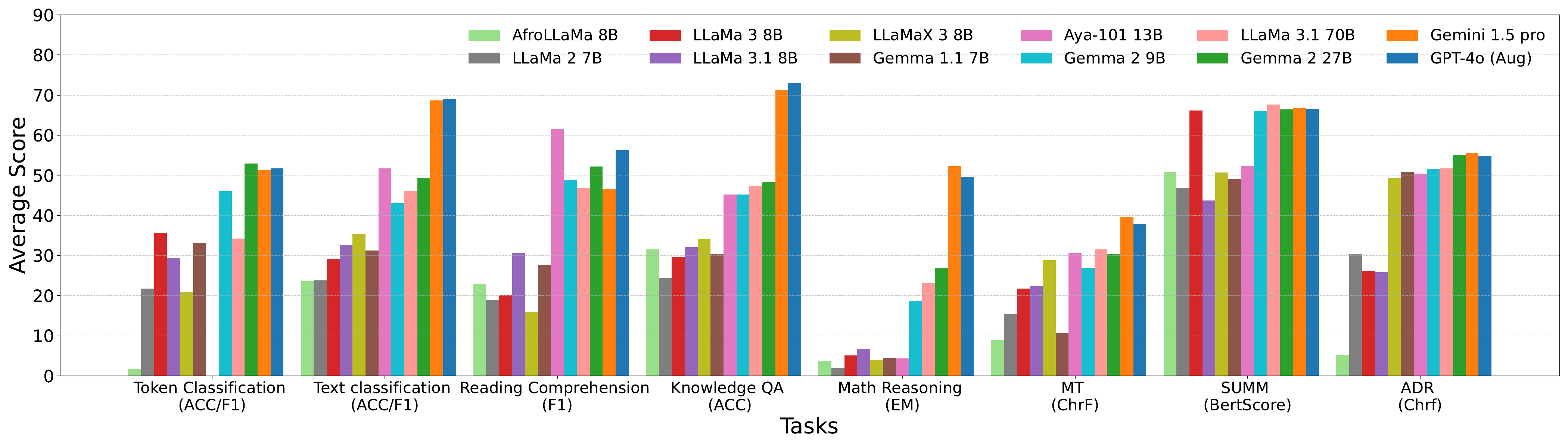}
     \vspace{-4mm}
    \caption{Performance of models across various NLP tasks, grouped by metric-based evaluation categories. Tasks include Token Classification, Text classification, Reading Comprehension QA, Knowledge QA, Math Reasoning, Machine Translation (MT), and Summarization (SUMM) and Diacritics Restoration (ADR).}
    \label{fig:afrobench_tasks_results}
     \vspace{-2mm}
\end{figure*}

\section{LLMs evaluated}
\label{sec:llm_evaluated}
Models are selected to cover a range of open and closed-source LLMs with diverse parameter sizes, multilingual capabilities, and recent advancements. We prioritize models with strong multilingual support, accessibility for research, and relevance to African languages.
\subsubsection{Open Models}
These are LLMs whose architectures, weights, and often training datasets are publicly available, allowing researchers and practitioners to fine-tune or adapt them to specific use cases. These models promote transparency, replicability, and accessibility, particularly for low-resource language tasks. 

\paragraph{Aya-101.} Aya-101~\citep{ustün2024aya} is a T5-style encoder-decoder model specifically fine-tuned for low-resource multilingual applications, including African languages. It was fine-tuned on a curated dataset, consisting of public multilingual corpora, and machine \& human translated datasets from more than 100 languages. The model adopts a text-to-text paradigm and emphasizes cross-lingual transfer learning, allowing for robust generalization across various multilingual text-based tasks

\paragraph{LLaMa 2 7B Chat.} LLaMa 2~\citep{Touvron2023Llama2O} is a collection of open-source pretrained and fine-tuned generative text models developed by Meta, ranging from 7 billion to 70 billion parameters. The 7B Chat variant allows for dialogue use cases. It employs an auto-regressive transformer architecture and has been fine-tuned using supervised fine-tuning (SFT) and reinforcement learning with human feedback (RLHF). They are pretrained on multiple languages, but has limited coverage of African languages.

\paragraph{LLaMa 3 8B Instruct} Llama 3~\citep{Dubey2024TheL3} is an updated variant of Llama 2 ~\citep{Touvron2023Llama2O} series. They are instruction-fine-tuned to handle a wide range of text-based tasks. Similar to LLaMa 2, it also supports multiple languages but coverage of African languages remains limited.  The number of parameters ranges from 8B to 70B; we make use of the 8B for this evaluation.

\paragraph{LLaMa 3.1 Instruct (8B, 70B)} LLaMa 3.1~\citep{Llama3.1Meta2024} is an updated variant of the LLaMa 3 series. Compared to LLaMa 3 \citep{Dubey2024TheL3}, LLaMa 3.1~\citep{Llama3.1Meta2024} introduces improvements in multilingual capabilities and general instruction-following. We use the instruction-tuned variants, fine-tuned for a broad range of NLP tasks. While it supports multiple languages, coverage of African languages remains limited. The model is available in parameter sizes ranging from 8B to 405B; due to computational cost, we evaluate only the 8B and 70B variants.

\paragraph{Gemma 1.1 7B IT.} ~\citep{Mesnard2024GemmaOM} is a lightweight open model from Google, built from the same research and technology used to create the Gemini models. They are text-to-text, decoder-only large language models, available in English, with open weights, pre-trained variants, and instruction-tuned variants. However, it does not have strong multilingual support. We evaluate the 7B instruction-finetuned variant of this model.

\paragraph{Gemma 2 IT (9B, 27B).} Gemma 2 \citep{Riviere2024Gemma2I} is an improved iteration of the Gemma model series optimized for efficiency. Comapred to Gemma 1, Gemma 2 incorporates enhanced instruction-following capabilities and more robust parameter scaling. We evaluate the instruction-tuned variants of Gemma 2 at 9B and 27B parameter scales.

\paragraph{AfroLlama-V1.}~\citep{AfroLlamaV1} is a decoder-only transformer model, optimized for African language applications. It leverages proprietary datasets, including text from social media, newspapers, and government publications in African languages. Its architecture is based on LLaMa 3 8B ~\citep{Dubey2024TheL3}, but it incorporates additional pretraining on African-centric text. 

\subsubsection{Proprietary Models}
These are proprietary systems developed and maintained by organizations. Their training data and architectures are typically undisclosed.

\paragraph{GPT-4o (Aug)} GPT-4o~\citep{openai2024gpt4o} is an optimized version of OpenAI's GPT-4 model~\citep{OpenAI2023GPT4TR}. It is an autoregressive omni model, trained end-to-end across text, vision, and audio on both public and proprietary data. While specific details about its architecture and datasets are not publicly disclosed, the GPT series is designed to adapt effectively to various language tasks, making it suitable for applications involving African languages. We evaluated the August 2024 version of this model

\paragraph{Gemini 1.5 Pro 002.} Gemini~\citep{Reid2024Gemini1U} s a cutting-edge proprietary model with strong multilingual capacity. Its a compute-efficient multimodal mobel withtraining data are tailored for diverse linguistic contexts, including low-resource languages. While specific details about its architecture and datasets are not publicly disclosed, Gemini is designed to adapt effectively to various language tasks, making it suitable for applications involving African languages.

\section{Evaluation Tools and Framework}
\label{sec:tool_integration}
AfroBench and AfroBench-Lite is fully integrated with Eleuther LM Evaluation Harness \citep{eval-harness} for open models, with sample run scripts and instructions on how to run the benchmark. We chose Eleuther LM Evaluation Harness due to their open-source and reproducible nature and widespread adoption within the industry. The evaluation methodology varies by task type: text classification and multiple-choice tasks are assessed using log-likelihood evaluation, which measures the probability of a prompt-generated continuation containing the expected response, while all other tasks utilize free-form generation approaches.

For proprietary models accessed through their API, we built a custom framework to prompt and evaluate these models. This framework is also open-sourced with sample run scripts and instructions on how to reproduce the benchmarks. The same prompt and evaluation methodology per task is used in both the LM Evaluation Harness and our custom API framework.  

\section{AfroBench Evaluation with Confidence Scores}
\label{sec:afrobench_ci}
We computed 95\% confidence intervals AfroBench results to quantify statistical significance. The calculation was based on the results of 5 prompts for each task (3 prompts for NLG tasks).  \autoref{tab:confidence_intervals} presents the average performance and confidence intervals accross prompts to assess variability and significance.

\section{Newer LLM evaluation on AfroBench-Lite}
\label{sec:afrobench_lite}
We extended our evaluation for \afrolite to includes newer LLMs such as Lugha-LLaMa (an African-centric LLM)~\citep{Buzaaba2025LughaLlamaAL}, GPT-4.1, Gemini-2.0-Flash, and LLaMa 4 400B (Maverick) in \autoref{tab:afrobench-lite-new}.

\insertafrobenchlitenew

\begin{table*}[t]
\centering
\scriptsize
\setlength{\tabcolsep}{2pt}
\resizebox{\textwidth}{!}{%
\begin{tabular}{lccccccccccc}
\toprule
\textbf{Task} & \textbf{LLaMa2 7B} & \textbf{LLaMa3 8B} & \textbf{LLaMaX 8B} & \textbf{LLaMa3.1 8B} & \textbf{AfroLLaMa 8B} & \textbf{Gemma2 9B} & \textbf{Aya-101 13B} & \textbf{Gemma2 27B} & \textbf{LLaMa3.1 70B} & \textbf{Gemini1.5 Pro} & \textbf{GPT-4o (Aug)} \\
\midrule
POS & 22.6±13.6 & 45.8±4.4 & 38.7±4.4 & 42.9±6.5 & 0.0±0.0 & 47.9±7.9 & 0.0±0.0 & 53.6±3.1 & 52.0±4.7 & 59.5±3.0 & 60.1±5.9 \\
NER & 11.1±10.7 & 17.3±8.3 & 0.0±0.0 & 7.7±5.6 & 2.9±2.2 & 25.9±30.8 & 0.0±0.0 & 43.1±11.4 & 12.9±5.8 & 40.6±3.6 & 37.1±6.8 \\
SA & 37.5±17.0 & 39.7±16.3 & 44.5±17.1 & 45.7±18.4 & 39.8±25.2 & 48.3±29.0 & 60.0±9.8 & 58.4±17.3 & 43.4±18.3 & 65.4±15.2 & 64.6±17.7 \\
TC & 15.3±14.5 & 24.6±26.9 & 23.5±32.0 & 37.5±26.4 & 16.9±22.1 & 51.6±15.9 & 68.9±4.4 & 59.4±8.7 & 47.0±17.5 & 73.5±10.2 & 73.3±4.9 \\
Intent & 0.8±1.5 & 0.9±2.3 & 3.1±3.8 & 4.0±5.0 & 0.3±1.0 & 29.2±5.6 & 42.4±4.6 & 33.0±4.9 & 31.8±7.4 & 68.4±12.2 & 70.4±6.6 \\
Hate & 16.8±10.8 & 21.8±11.0 & 23.0±12.5 & 19.3±5.9 & 15.2±8.1 & 21.3±13.0 & 28.7±— & 36.6±15.0 & 36.5±29.3 & 49.7±33.5 & 49.5±37.6 \\
NLI & 33.4±1.5 & 33.7±2.7 & 35.0±6.8 & 34.3±3.8 & 34.2±4.4 & 36.3±6.6 & 48.3±5.3 & 37.3±7.3 & 35.2±5.4 & 56.1±15.9 & 58.4±11.4 \\
XQA & 10.4±5.7 & 9.6±5.6 & 2.0±0.5 & 14.1±14.3 & 19.2±5.2 & 39.3±13.8 & 61.9±1.6 & 47.7±7.6 & 37.1±8.7 & 34.8±11.8 & 31.6±25.0 \\
RC & 24.3±3.5 & 28.0±8.2 & 24.6±5.7 & 36.2±16.2 & 24.4±2.5 & 47.7±26.2 & 55.2±29.4 & 47.6±28.8 & 44.5±16.8 & 52.7±7.6 & 71.4±3.2 \\
Arc-E & 21.0±4.3 & 30.8±3.8 & 39.3±2.6 & 31.7±3.0 & 35.8±2.8 & 52.9±1.8 & 59.3±1.4 & 55.5±1.9 & 55.4±4.3 & 83.8±2.1 & 85.2±1.4 \\
MMLU & 24.5±2.4 & 26.7±2.2 & 28.0±1.4 & 30.3±4.5 & 25.1±1.9 & 34.8±8.8 & 30.4±4.0 & 38.9±9.6 & 37.9±8.6 & 50.7±12.2 & 55.3±15.5 \\
Math & 1.8±1.3 & 4.2±3.2 & 3.7±2.5 & 5.5±3.4 & 0.1±0.4 & 14.1±8.0 & 4.3±1.6 & 25.4±4.8 & 20.3±5.4 & 46.6±20.3 & 48.7±4.2 \\
MT (en-xx) & 7.9±7.1 & 15.0±4.7 & 21.9±4.6 & 16.1±2.5 & 7.4±3.2 & 24.5±1.2 & 23.0±2.9 & 27.5±2.8 & 24.7±5.2 & 37.6±1.9 & 34.4±2.9 \\
MT (xx-en) & 17.8±7.5 & 23.1±11.0 & 34.0±5.0 & 27.7±3.8 & 8.3±3.0 & 28.8±0.8 & 36.9±4.1 & 32.7±1.5 & 35.8±8.8 & 41.7±0.8 & 40.5±1.5 \\
ADR & 22.8±19.2 & 24.1±7.4 & 47.2±6.6 & 23.1±6.8 & 4.3±2.4 & 50.3±4.4 & 49.8±1.8 & 53.5±4.4 & 48.2±16.2 & 54.5±4.2 & 52.9±5.0 \\
\bottomrule
\end{tabular}
}
\caption{Model performance based on average with standard deviation at 95\% confidence intervals}
\label{tab:confidence_intervals}
\end{table*}

\section{Languages covered in the evaluation}
\label{appendix:languages}
\autoref{tab:languages_covered} shows the languages and tasks we evaluated on.


\insertlanguagescovered

\inserttaskscovered

\section{Best Performing Prompt}
Below details which prompt performed best per model per dataset. Actual prompt can be retrieved from \ref{sec:prompt-bank}.
\insertbestprompts

\clearpage

\section{Prompt Bank}
\label{sec:prompt-bank}
In this section, we list all prompts used in our experiments. 
We use zero-shot cross-lingual prompts, where the context and query are in English, while the input text is in the target African language. This approach leverages LLMs' stronger instruction-following in English \citep{xglm, wei-multilingual}.
We display the prompts grouped by the task category shown in Figure~\ref{fig:afrobench-overview}. 
\subsection{Natural Language Understanding}
\label{sec:nlu}
\noindent\textbf{POS prompts:}
\begin{lstlisting}[
  caption=Listing 1: MasakhaPOS Prompt 1,
  basicstyle=\scriptsize\ttfamily,
  breaklines=true
]
Please provide the POS tags for each word in the input sentence. The input will be a list of words in the sentence. The output format should be a list of tuples, where each tuple consists of a word from the input text and its corresponding POS tag label from the tag label set: ['ADJ', 'ADP', 'ADV', 'AUX', 'CCONJ, 'DET', 'INTJ', 'NOUN', 'NUM', 'PART', 'PRON', 'PROPN', 'PUNCT', 'SCONJ', 'SYM', 'VERB', 'X']. 
Your response should include only a list of tuples, in the order that the words appear in the input sentence, including punctuations, with each tuple containing the corresponding POS tag label for a word. 

Sentence: {{text}}
Output: 
\end{lstlisting}
\begin{lstlisting}[
  caption=Listing 2: MasakhaPOS Prompt 2,
  basicstyle=\scriptsize\ttfamily,
  breaklines=true
]
You are an expert in tagging words and sentences in {{language}} with the right POS tag. 

Please provide the POS tags for each word in the {{language}} sentence. The input is a list of words in the sentence. POS tag label set: [ 'ADJ', 'ADP', 'ADV', 'AUX', 'CCONJ, 'DET', 'INTJ', 'NOUN', 'NUM', 'PART', 'PRON', 'PROPN', 'PUNCT', 'SCONJ', 'SYM', 'VERB', 'X' ]. The output format should be a list of tuples, where each tuple consists of a word from the input text and its corresponding POS tag label from the POS tag label set provided.
Your response should include only a list of tuples, in the order that the words appear in the input sentence, including punctuations, with each tuple containing the corresponding POS tag label for a word. 

Sentence: {{text}}
Output: 
\end{lstlisting}
\begin{lstlisting}[
  caption=Listing 3: MasakhaPOS Prompt 3,
  basicstyle=\scriptsize\ttfamily,
  breaklines=true
]
Acting as a {{language}} linguist and without making any corrections or changes to the text, perform a part of speech (POS) analysis of the sentences using the following POS tag label annotation ['ADJ', 'ADP', 'ADV', 'AUX', 'CCONJ, 'DET', 'INTJ', 'NOUN', 'NUM', 'PART', 'PRON', 'PROPN', 'PUNCT', 'SCONJ', 'SYM', 'VERB', 'X']. The input will be a list of words in the sentence. The output format should be a list of tuples, where each tuple consists of a word from the input text and its corresponding POS tag label from the POS tag label set provided.
Your response should include only a list of tuples, in the order that the words appear in the input sentence, including punctuations, with each tuple containing the corresponding POS tag label for a word.

Sentence: {{text}}
Output: 
\end{lstlisting}
\begin{lstlisting}[
  caption=Listing 4: MasakhaPOS Prompt 4,
  basicstyle=\scriptsize\ttfamily,
  breaklines=true
]
Annotate each word in the provided sentence with the appropriate POS tag. The annotation list is given as: ['ADJ', 'ADP', 'ADV', 'AUX', 'CCONJ, 'DET', 'INTJ', 'NOUN', 'NUM', 'PART', 'PRON', 'PROPN', 'PUNCT', 'SCONJ', 'SYM', 'VERB', 'X']. The input sentence will be a list of words in the sentence. The output format should be a list of tuples, where each tuple consists of a word from the input text and its corresponding POS tag label from the POS tag label set provided\nYour response should include only a list of tuples, in the order that the words appear in the input sentence, including punctuations, with each tuple containing the corresponding POS tag label for a word.

Sentence: {{text}}
Output: 
\end{lstlisting}
\begin{lstlisting}[
  caption=Listing 5: MasakhaPOS Prompt 5,
  basicstyle=\scriptsize\ttfamily,
  breaklines=true
]
Given the following sentence, identify the part of speech (POS) for each word. Use the following POS tag set: 
NOUN: Noun (person, place, thing), 
VERB: Verb (action, state), 
ADJ: Adjective (describes a noun), 
ADV: Adverb (modifies a verb, adjective, or adverb), 
PRON: Pronoun (replaces a noun), 
DET: Determiner (introduces a noun), 
ADP: Adposition (preposition or postposition), 
CCONJ: Conjunction (connects words, phrases, clauses) 
PUNCT: Punctuation, 
PROPN: Proper Noun, 
AUX: Auxiliary verb (helper verb), \nSCONJ: Subordinating conjunction 
PART: Particle, 
SYM: Symbol, 
INTJ: Interjection, 
NUM: Numeral, 
X: others. The output format should be a list of tuples, where each tuple consists of a word from the input text and its corresponding POS tag label key only from the POS tag set provided
Your response should include only a list of tuples, in the order that the words appear in the input sentence, including punctuations, with each tuple containing the corresponding POS tag label for a word.

Sentence: {{text}}
Output: 
\end{lstlisting}
\noindent\textbf{NER prompts:}
\begin{lstlisting}[
  caption=Listing 1: MasakhaNER Prompt 1,
  basicstyle=\scriptsize\ttfamily,
  breaklines=true
]
Named entities refers to names of location, organisation and personal name. 
For example, 'David is an employee of Amazon and he is visiting New York next week to see Esther' will be 
PERSON: David $ ORGANIZATION: Amazon $ LOCATION: New York $ PERSON: Esther 

Ensure the output strictly follows the format: label: entity $ label: entity, with each unique entity on a separate label line, avoiding grouped entities (e.g., avoid LOC: entity, entity) or irrelevant entries like none. 

Text: {{text}} 
Return only the output
\end{lstlisting}
\begin{lstlisting}[
  caption=Listing 2: MasakhaNER Prompt 2,
  basicstyle=\scriptsize\ttfamily,
  breaklines=true
]
You are working as a named entity recognition expert and your task is to label a given text with named entity labels. Your task is to identify and label any named entities present in the text. The named entity labels that you will be using are PER (person), LOC (location), ORG (organization) and DATE (date). Label multi-word entities as a single named entity. For words which are not part of any named entity, do not return any value for it. 
Ensure the output strictly follows the format: label: entity $$ label: entity, with each unique entity on a separate label line, avoiding grouped entities (e.g., avoid LOC: entity, entity) or irrelevant entries like none. Return only the output

Text: {{text}} 
\end{lstlisting}
\begin{lstlisting}[
  caption=Listing 3: MasakhaNER Prompt 3,
  basicstyle=\scriptsize\ttfamily,
  breaklines=true
]
You are a Named Entity Recognition expert in {{language}} language. 
Extract all named entities from the following {{language}} text and categorize them into PERSON, LOCATION, ORGANIZATION, or DATE. 
Ensure the output strictly follows the format; label: entity $$ label: entity, with each unique entity on a separate label line, avoiding grouped entities (e.g., avoid LOC: entity, entity) or irrelevant entries like none. Return only the output

Text: {{text}} 
Return only the output
\end{lstlisting}
\begin{lstlisting}[
  caption=Listing 4: MasakhaNER Prompt 4,
  basicstyle=\scriptsize\ttfamily,
  breaklines=true
]
As a {{language}} linguist, label all named entities in the {{language}} text below with the categories: PERSON, LOCATION, ORGANIZATION, and DATE. Ensure the output strictly follows the format; label: entity $$ label: entity, with each unique entity on a separate label line, avoiding grouped entities (e.g., avoid LOC: entity, entity) or irrelevant entries like none. Return only the output.

Text: {{text}} 
Return only the output
\end{lstlisting}
\begin{lstlisting}[
  caption=Listing 5: MasakhaNER Prompt 5,
  basicstyle=\scriptsize\ttfamily,
  breaklines=true
]
Provide a concise list of named entities in the text below. Use the following labels: PERSON, LOCATION, ORGANIZATION, and DATE. Ensure the output strictly follows the format; label: entity $$ label: entity, with each unique entity on a separate label line, avoiding grouped entities (e.g., avoid LOC: entity, entity) or irrelevant entries like none. Return only the output.  

Text: {{text}} 
Return only the output
\end{lstlisting}
\noindent\textbf{Sentiment prompts:}
\begin{lstlisting}[
  caption=Listing 1: AfriSenti Prompt 1,
  basicstyle=\scriptsize\ttfamily,
  breaklines=true
]
Does this statement; "{{tweet}}" have a Neutral, Positive or Negative sentiment? Labels only
\end{lstlisting}
\begin{lstlisting}[
  caption=Listing 2: AfriSenti Prompt 2,
  basicstyle=\scriptsize\ttfamily,
  breaklines=true
]
Does this {{language}} statement; "{{tweet}}" have a Neutral, Positive or Negative sentiment? Labels only
\end{lstlisting}
\begin{lstlisting}[
  caption=Listing 3: AfriSenti Prompt 3,
  basicstyle=\scriptsize\ttfamily,
  breaklines=true
]
You are an assistant able to detect sentiments in tweets. 

Given the sentiment labels Neutral, Positive or Negative; what is the sentiment of the {{language}} statement below? Return only the labels. 

text: {{tweet}} 
label:
\end{lstlisting}
\begin{lstlisting}[
  caption=Listing 4: AfriSenti Prompt 4,
  basicstyle=\scriptsize\ttfamily,
  breaklines=true
]
Label the following text as Neutral, Positive, or Negative. Provide only the label as your response. 

text: {{tweet}} 
label: 
\end{lstlisting}
\begin{lstlisting}[
  caption=Listing 5: AfriSenti Prompt 5,
  basicstyle=\scriptsize\ttfamily,
  breaklines=true
]
You are tasked with performing sentiment classification on the following {{language}} text. For each input, classify the sentiment as positive, negative, or neutral. Use the following guidelines: 

Positive: The text expresses happiness, satisfaction, or optimism. 
Negative: The text conveys disappointment, dissatisfaction, or pessimism. 
Neutral: The text is factual, objective, or without strong emotional undertones. 
If the text contains both positive and negative sentiments, choose the dominant sentiment. For ambiguous or unclear sentiments, select the label that best reflects the overall tone. Please provide a single classification for each input.

text: {{tweet}} 
label: 
\end{lstlisting}
\begin{lstlisting}[
  caption=Listing 6: NollySenti Prompt 1,
  basicstyle=\scriptsize\ttfamily,
  breaklines=true
]
Does this movie description "{{review}}" have a Positive or Negative sentiment? Labels only
\end{lstlisting}
\begin{lstlisting}[
  caption=Listing 7: NollySenti Prompt 2,
  basicstyle=\scriptsize\ttfamily,
  breaklines=true
]
Does this {{language} movie description; "{{review}}" have a Positive or Negative sentiment? Labels only
\end{lstlisting}
\begin{lstlisting}[
  caption=Listing 8: NollySenti Prompt 3,
  basicstyle=\scriptsize\ttfamily,
  breaklines=true
]
You are an assistant able to detect sentiment in movie reviews. 

Given the sentiment labels Positive or Negative; what is the sentiment of the English statement below? Return only the labels

Review: {{review}}"
\end{lstlisting}
\begin{lstlisting}[
  caption=Listing 9: NollySenti Prompt 4,
  basicstyle=\scriptsize\ttfamily,
  breaklines=true
]
Label the following text as Positive, or Negative. Provide only the label as your response. 

text: {{review}} 
label: 
\end{lstlisting}
\begin{lstlisting}[
  caption=Listing 10: NollySenti Prompt 5,
  basicstyle=\scriptsize\ttfamily,
  breaklines=true
]
You are tasked with performing sentiment classification on the following English text. For each input, classify the sentiment as positive, negative. Use the following guidelines: 
Positive: The text expresses happiness, satisfaction, or optimism. 
Negative: The text conveys disappointment, dissatisfaction, or pessimism. 
If the text contains both positive and negative sentiments, choose the dominant sentiment. For ambiguous or unclear sentiments, select the label that best reflects the overall tone. Please provide a single classification for each input.

text: {{review}}
label: 
\end{lstlisting}
\noindent\textbf{Topic Classification prompts:}
\begin{lstlisting}[
  caption=Listing 1: SIB Prompt 1,
  basicstyle=\scriptsize\ttfamily,
  breaklines=true
]
Given the categories science/technology, travel, politics, sports, health, entertainment, or geography; what category does the text: '{{text}}' belong to:
\end{lstlisting}
\begin{lstlisting}[
  caption=Listing 2: SIB Prompt 2,
  basicstyle=\scriptsize\ttfamily,
  breaklines=true
]
Does this {{language}} topic; '{{text}}' belong to one of the following categories: science/technology, travel, politics, sports, health, entertainment, or geography? category only
\end{lstlisting}
\begin{lstlisting}[
  caption=Listing 3: SIB Prompt 3,
  basicstyle=\scriptsize\ttfamily,
  breaklines=true
]
You are an assistant able to classify topics in texts. 

Given the categories science/technology, travel, politics, sports, health, entertainment, or geography; what is the topic of the {{language}} statement below? Return only the category. 

text: {{text}} 
category: "
\end{lstlisting}
\begin{lstlisting}[
  caption=Listing 4: SIB Prompt 4,
  basicstyle=\scriptsize\ttfamily,
  breaklines=true
]
Label the following text as science/technology, travel, politics, sports, health, entertainment, or geography. Provide only the category as your response. 

text: {{text}} 
category:
\end{lstlisting}
\begin{lstlisting}[
  caption=Listing 5: SIB Prompt 5,
  basicstyle=\scriptsize\ttfamily,
  breaklines=true
]
You are tasked with performing topic classification on the following {{language}} text. For each input, classify the topic as science/technology, travel, politics, sports, health, entertainment, or geography. Use the following guidelines: 

science/technology: The text discusses scientific discoveries, technological advancements, or related topics. 
travel: The text describes travel experiences, destinations, or related topics. 
politics: The text covers political events, policies, or related topics. 
sports: The text talks about sports events, athletes, or related topics. 
health: The text addresses health issues, medical advancements, or related topics. 
entertainment: The text pertains to movies, music, celebrities, or related topics. 
geography: The text involves geographical information, locations, or related topics. 

If the text contains multiple topics, choose the dominant topic. For ambiguous or unclear topics, select the category that best reflects the overall content. Please provide a single classification for each input.

text: {{text}} 
category: 
\end{lstlisting}
\begin{lstlisting}[
  caption=Listing 6: MasakhaNEWS Prompt 1,
  basicstyle=\scriptsize\ttfamily,
  breaklines=true
]
Given the categories technology, business, politics, sports, health, entertainment, or religion; what category does the text: '{{headline}}' belong to: 

Return only the one category
\end{lstlisting}
\begin{lstlisting}[
  caption=Listing 7: MasakhaNEWS Prompt 2,
  basicstyle=\scriptsize\ttfamily,
  breaklines=true
]
Does this {{language}} topic; '{{headline}}' belong to one of the following categories: technology, business, politics, sports, health, entertainment, or religion? category only
\end{lstlisting}
\begin{lstlisting}[
  caption=Listing 8: MasakhaNEWS Prompt 3,
  basicstyle=\scriptsize\ttfamily,
  breaklines=true
]
You are an assistant able to classify topics in texts. 

Given the categories technology, religion, politics, sports, health, entertainment, or business; what is 

text: {{headline}} 
category: 
\end{lstlisting}
\begin{lstlisting}[
  caption=Listing 9: MasakhaNEWS Prompt 4,
  basicstyle=\scriptsize\ttfamily,
  breaklines=true
]
Label the following text as technology, religion, politics, sports, health, entertainment, or geography. Provide only the category as your response. 

text: {{headline}} 
category: 
\end{lstlisting}
\begin{lstlisting}[
  caption=Listing 10: MasakhaNEWS Prompt 5,
  basicstyle=\scriptsize\ttfamily,
  breaklines=true
]
You are tasked with performing topic classification on the following {{language}} text. For each input, classify the topic as technology, business, politics, sports, health, entertainment, or religion. Use the following guidelines: 

technology: The text discusses scientific discoveries, technological advancements, or related topics. 
politics: The text covers political events, policies, or related topics. 
sports: The text talks about sports events, athletes, or related topics. 
health: The text addresses health issues, medical advancements, or related topics. 
entertainment: The text pertains to movies, music, celebrities, or related topics. 
religion: The text talks about relgions, religious institutions and beliefs or related topics. 
business: The text covers economy, business, or related topics. 

If the text contains multiple topics, choose the dominant topic. For ambiguous or unclear topics, select the category that best reflects the overall content. Please provide a single classification for each input.

text: {{headline}} 
category:
\end{lstlisting}
\noindent\textbf{Intent Detection prompts:}
\begin{lstlisting}[
  caption=Listing 1: IngongoIntent Prompt 1,
  basicstyle=\scriptsize\ttfamily,
  breaklines=true
]
Given the text: '{{text}}', classify it into one of these intents: [alarm, balance, bill_balance, book_flight, book_hotel, calendar_update, cancel_reservation, car_rental, confirm_reservation, cook_time, exchange_rate, food_last, freeze_account, ingredients_list, interest_rate, international_visa, make_call, meal_suggestion, min_payment, pay_bill, pin_change, play_music, plug_type, recipe, restaurant_reservation, restaurant_reviews, restaurant_suggestion, share_location, shopping_list_update, spending_history, text, time, timezone, transactions, transfer, translate, travel_notification, travel_suggestion, update_playlist, weather]. Only output one intent from the list.
\end{lstlisting}
\begin{lstlisting}[
  caption=Listing 2: IngongoIntent Prompt 2,
  basicstyle=\scriptsize\ttfamily,
  breaklines=true
]
Analyze the text: '{{text}}'. Choose the most appropriate intent from these options: [alarm, balance, bill_balance, book_flight, book_hotel, calendar_update, cancel_reservation, car_rental, confirm_reservation, cook_time, exchange_rate, food_last, freeze_account, ingredients_list, interest_rate, international_visa, make_call, meal_suggestion, min_payment, pay_bill, pin_change, play_music, plug_type, recipe, restaurant_reservation, restaurant_reviews, restaurant_suggestion, share_location, shopping_list_update, spending_history, text, time, timezone, transactions, transfer, translate, travel_notification, travel_suggestion, update_playlist, weather]. Respond with only the selected intent.
\end{lstlisting}
\begin{lstlisting}[
  caption=Listing 3: IngongoIntent Prompt 3,
  basicstyle=\scriptsize\ttfamily,
  breaklines=true
]
You are a linguistic analyst trained to understand user intent. Based on the text: '{{text}}', choose the intent that best matches from this list: [alarm, balance, bill_balance, book_flight, book_hotel, calendar_update, cancel_reservation, car_rental, confirm_reservation, cook_time, exchange_rate, food_last, freeze_account, ingredients_list, interest_rate, international_visa, make_call, meal_suggestion, min_payment, pay_bill, pin_change, play_music, plug_type, recipe, restaurant_reservation, restaurant_reviews, restaurant_suggestion, share_location, shopping_list_update, spending_history, text, time, timezone, transactions, transfer, translate, travel_notification, travel_suggestion, update_playlist, weather]. Return only the intent.
\end{lstlisting}
\begin{lstlisting}[
  caption=Listing 4: IngongoIntent Prompt 4,
  basicstyle=\scriptsize\ttfamily,
  breaklines=true
]
You are a English linguistic analyst trained to understand {{language}} user intent. Based on the {{language}} text: "{{text}}", choose the intent that best matches from this list: [alarm, balance, bill_balance, book_flight, book_hotel, calendar_update, cancel_reservation, car_rental, confirm_reservation, cook_time, exchange_rate, food_last, freeze_account, ingredients_list, interest_rate, international_visa, make_call, meal_suggestion, min_payment, pay_bill, pin_change, play_music, plug_type, recipe, restaurant_reservation, restaurant_reviews, restaurant_suggestion, share_location, shopping_list_update, spending_history, text, time, timezone, transactions, transfer, translate, travel_notification, travel_suggestion, update_playlist, weather]. Return only the intent.
\end{lstlisting}
\begin{lstlisting}[
  caption=Listing 5: IngongoIntent Prompt 5,
  basicstyle=\scriptsize\ttfamily,
  breaklines=true
]
The following text is in {{language}}: '{{text}}'. Given the list of intents: [alarm, balance, bill_balance, book_flight, book_hotel, calendar_update, cancel_reservation, car_rental, confirm_reservation, cook_time, exchange_rate, food_last, freeze_account, ingredients_list, interest_rate, international_visa, make_call, meal_suggestion, min_payment, pay_bill, pin_change, play_music, plug_type, recipe, restaurant_reservation, restaurant_reviews, restaurant_suggestion, share_location, shopping_list_update, spending_history, text, time, timezone, transactions, transfer, translate, travel_notification, travel_suggestion, update_playlist, weather], identify the intent expressed in the text. Return only the identified intent.
\end{lstlisting}
\noindent\textbf{Hate Speech prompts:}
\begin{lstlisting}[
  caption=Listing 1: AfriHate Prompt 1,
  basicstyle=\scriptsize\ttfamily,
  breaklines=true
]
I am providing you with the definition Hate speech, Abusive language and Normal tweets. 
Hate speech is a language content that expresses hatred towards a particular group or individual based on their political affiliation, race, ethnicity, religion, gender, sexual orientation, or other characteristics. It also includes threats of violence 
Abusive language is any form of bad language expressions including rude, impolite, insulting or belittling utterance intended to offend or harm an individual. 
Normal does not contain any bad language. 

Tweet: {{tweet}} 

Which category does the tweet above belong to: 'Hate', 'Abuse' or 'Normal'. Pick exactly one category. Return only the label
\end{lstlisting}
\begin{lstlisting}[
  caption=Listing 2: AfriHate Prompt 2,
  basicstyle=\scriptsize\ttfamily,
  breaklines=true
]
Read the following label definitions and provide a label without any explanations. 

Hate: Hate speech is public speech that expresses hate or encourages violence towards a person or group based on something such as race, religion, gender, ethnicity, sexual orientation or other characteristics. 

Abusive: Abusive and offensive language means verbal messages that use words in an inappropriate way and may include but is not limited to swearing, name-calling, or profanity. Offensive language may upset or embarrass people because it is rude or insulting. 

Normal: Normal language is neither hateful nor abusive or offensive. It does not contain any bad language. 

Text: {{tweet}} 
Label: 
\end{lstlisting}
\begin{lstlisting}[
  caption=Listing 3: AfriHate Prompt 3,
  basicstyle=\scriptsize\ttfamily,
  breaklines=true
]
Read the following text and definitions: 

Text: {{tweet}}. 

Definitions: 
Hate: Hate speech is public speech that expresses hate or encourages violence towards a person or group based on something such as race, religion, gender, ethnicity, sexual orientation or other characteristics. 

Abuse: Abusive and offensive language means verbal messages that use words in an inappropriate way and may include but is not limited to swearing, name-calling, or profanity. Offensive language may upset or embarrass people because it is rude or insulting. 

Normal: Normal language is neither hateful nor abusive or offensive. It does not contain any bad language. 

Which of these definitions (hate, abuse, normal) apply to this tweet?, return only the label
\end{lstlisting}
\begin{lstlisting}[
  caption=Listing 4: AfriHate Prompt 4,
  basicstyle=\scriptsize\ttfamily,
  breaklines=true
]
Read the following definitions and text to categorize: 

Definitions:
Hate: Hate speech is public speech that expresses hate or encourages violence towards a person or group based on something such as race, religion, gender, ethnicity, sexual orientation or other characteristics. 

Abuse: Abusive and offensive language means verbal messages that use words in an inappropriate way and may include but is not limited to swearing, name-calling, or profanity. Offensive language may upset or embarrass people because it is rude or insulting. 

Normal: Normal language is neither hateful nor abusive or offensive. It does not contain any bad language. 

Text: {{tweet}}. 

Which of these definitions (hate, abuse, normal) apply to this tweet? Return only the label
\end{lstlisting}
\begin{lstlisting}[
  caption=Listing 5: AfriHate Prompt 5,
  basicstyle=\scriptsize\ttfamily,
  breaklines=true
]
You will be given a text snippet and 3 category definitions. 
Your task is to choose which category applies to this text. 

Your text snippet is: {{tweet}}. 

Your category definitions are: 
HATE category definition: Hate speech is public speech that expresses hate or encourages violence towards a person or group based on something such as race, religion, gender, ethnicity, sexual orientation or other characteristics. 

ABUSE category definition: Abusive and offensive language means verbal messages that use words in an inappropriate way and may include but is not limited to swearing, name-calling, or profanity. Offensive language may upset or embarrass people because it is rude or insulting.

NORMAL category definition: Normal language is neither hateful nor abusive or offensive. It does not contain any bad language. 

Does the text snippet belong to the HATE, ABUSIVE, or the NORMAL category? Thinking step by step answer HATE, ABUSIVE, or NORMAL capitalizing all the letters. 
Explain your reasoning FIRST, then output HATE, ABUSIVE, or NORMAL. Clearly return the label in capital letters.
\end{lstlisting}
\noindent\textbf{Natural Language Inference prompts:}
\begin{lstlisting}[
  caption=Listing 1: AfriXNLI Prompt 1,
  basicstyle=\scriptsize\ttfamily,
  breaklines=true
]
Please identify whether the premise entails or contradicts the hypothesis in the following premise and hypothesis. The answer should be exact entailment, contradiction, or neutral. 

Premise: {{premise}} 
Hypothesis: {{hypothesis}}. 

Is it entailment, contradiction, or neutral?
\end{lstlisting}
\begin{lstlisting}[
  caption=Listing 2: AfriXNLI Prompt 2,
  basicstyle=\scriptsize\ttfamily,
  breaklines=true
]
{{premise}}
Question: {{hypothesis}} True, False, or Neither?
Answer: 
\end{lstlisting}
\begin{lstlisting}[
  caption=Listing 3: AfriXNLI Prompt 3,
  basicstyle=\scriptsize\ttfamily,
  breaklines=true
]
Given the following premise and hypothesis in {{language}}, identify if the premise entails, contradicts, or is neutral towards the hypothesis. Please respond with exact 'entailment', 'contradiction', or 'neutral'. 

Premise: {{premise}} 
Hypothesis: {{hypothesis}}
\end{lstlisting}
\begin{lstlisting}[
  caption=Listing 4: AfriXNLI Prompt 4,
  basicstyle=\scriptsize\ttfamily,
  breaklines=true
]
You are an expert in Natural Language Inference (NLI) specializing in {{language}} language.
Analyze the premise and hypothesis given in {{language}}, and determine the relationship between them.
Respond with one of the following options: 'entailment', 'contradiction', or 'neutral'. 

Premise: {{premise}}
Hypothesis: {{hypothesis}}
\end{lstlisting}
\begin{lstlisting}[
  caption=Listing 5: AfriXNLI Prompt 5,
  basicstyle=\scriptsize\ttfamily,
  breaklines=true
]
Based on the given statement, is the following claim 'true', 'false', or 'inconclusive'. 

Statement: {{premise}} 
Claim: {{hypothesis}}
\end{lstlisting}

\subsection{Question Answering}
\label{sec:QA}
\noindent\textbf{CrosslingualQA prompts:}
\begin{lstlisting}[
  caption=Listing 1: AfriQA Prompt 1,
  basicstyle=\scriptsize\ttfamily,
  breaklines=true
]
Your task is to answer a question given a context. 
Make sure you respond with the shortest span containing the answer in the context.
Question: {{question_lang}}
Context: {{context}}
Answer:
\end{lstlisting}
\begin{lstlisting}[
  caption=Listing 2: AfriQA Prompt 2,
  basicstyle=\scriptsize\ttfamily,
  breaklines=true
]
Your task is to answer a question given a context. The question is in {{language}}, while the context is in English or French. 
Make sure you respond with the shortest span in the context that contains the answer.
Question: {{question_lang}}
Context: {{context}}
Answer:
\end{lstlisting}
\begin{lstlisting}[
  caption=Listing 3: AfriQA Prompt 3,
  basicstyle=\scriptsize\ttfamily,
  breaklines=true
]
Given the context, provide the answer to the following question. 
Ensure your response is concise and directly from the context.
Question: {{question_lang}}
Context: {{context}}
Answer:
\end{lstlisting}
\begin{lstlisting}[
  caption=Listing 4: AfriQA Prompt 4,
  basicstyle=\scriptsize\ttfamily,
  breaklines=true
]
You are an AI assistant and your task is to answer the question based on the provided context. 
Your answer should be the shortest span that contains the answer within the context.
Question: {{question_lang}}
Context: {{context}}
Answer:
\end{lstlisting}
\begin{lstlisting}[
  caption=Listing 5: AfriQA Prompt 5,
  basicstyle=\scriptsize\ttfamily,
  breaklines=true
]
Using the context, find the answer to the question.
Respond with the briefest span that includes the answer from the context.
Question: {{question_lang}}
Context: {{context}}
Answer:
\end{lstlisting}
\noindent\textbf{Reading Comprehension prompts:}
\begin{lstlisting}[
  caption=Listing 1: Belebele Prompt 1,
  basicstyle=\scriptsize\ttfamily,
  breaklines=true
]
P: {{passage}}
Q: {{question}}
A: {{option_1}}
B: {{option_2}}
C: {{option_3}}
D: {{option_4}}
Please choose the correct answer from the options above:
\end{lstlisting}
\begin{lstlisting}[
  caption=Listing 2: Belebele Prompt 2,
  basicstyle=\scriptsize\ttfamily,
  breaklines=true
]
Passage: {{passage}}
Question: {{question}}
1: {{option_1}}
2: {{option_2}}
3: {{option_3}}
4: {{option_4}}
Please select the correct answer from the given choices
\end{lstlisting}
\begin{lstlisting}[
  caption=Listing 3: Belebele Prompt 3,
  basicstyle=\scriptsize\ttfamily,
  breaklines=true
]
Context: {{passage}}
Query: {{question}}
Option A: {{option_1}}
Option B: {{option_2}}
Option C: {{option_3}}
Option D: {{option_4}}
Please indicate the correct option from the list above:
\end{lstlisting}
\begin{lstlisting}[
  caption=Listing 4: Belebele Prompt 4,
  basicstyle=\scriptsize\ttfamily,
  breaklines=true
]
{{passage}}
Based on the above passage, answer the following question:
{{question}}
Choices:
A) {{option_1}}
B) {{option_2}}
C) {{option_3}}
D) {{option_4}}
Please provide the correct answer from the choices given
\end{lstlisting}
\begin{lstlisting}[
  caption=Listing 5: Belebele Prompt 5,
  basicstyle=\scriptsize\ttfamily,
  breaklines=true
]
Read the passage: {{passage}}
Then answer the question: {{question}}
Options:
A. {{option_1}}
B. {{option_2}}
C. {{option_3}}
D. {{option_4}}
Please choose the correct option from the above list
\end{lstlisting}
\begin{lstlisting}[
  caption=Listing 6: NaijaRC Prompt 1,
  basicstyle=\scriptsize\ttfamily,
  breaklines=true
]
P: {{story}}
Q: {{question}}
A: {{options_A}}
B: {{options_B}}
C: {{options_C}}
D: {{options_D}}
Please choose the correct answer from the options above
\end{lstlisting}
\begin{lstlisting}[
  caption=Listing 7: NaijaRC Prompt 2,
  basicstyle=\scriptsize\ttfamily,
  breaklines=true
]
Passage: {{story}}
Question: {{question}}
1: {{options_A}}
2: {{options_B}}
3: {{options_C}}
4: {{options_D}}
Please select the correct answer from the given choices
\end{lstlisting}
\begin{lstlisting}[
  caption=Listing 8: NaijaRC Prompt 3,
  basicstyle=\scriptsize\ttfamily,
  breaklines=true
]
Context: {{story}}
Query: {{question}}
Option A: {{options_A}}
Option B: {{options_B}}
Option C: {{options_C}}
Option D: {{options_D}}
Please indicate the correct option from the list above
\end{lstlisting}
\begin{lstlisting}[
  caption=Listing 9: NaijaRC Prompt 4,
  basicstyle=\scriptsize\ttfamily,
  breaklines=true
]
{{story}}
Based on the above passage, answer the following question
{{question}}
Choices:
A) {{options_A}}
B) {{options_B}}
C) {{options_C}}
D) {{options_D}}
Please provide the correct answer from the choices given
\end{lstlisting}
\begin{lstlisting}[
  caption=Listing 10: NaijaRC Prompt 5,
  basicstyle=\scriptsize\ttfamily,
  breaklines=true
]
Read the passage: {{story}}
Then answer the question: {{question}}
Options:
A. {{options_A}}
B. {{options_B}}
C. {{options_C}}
D. {{options_D}}
Please choose the correct option from the above list
\end{lstlisting}

\subsection{Knowledge}
\label{sec:QA}
\noindent\textbf{Arc-E prompts:}
\begin{lstlisting}[
  caption=Listing 1: UHURA Prompt 1,
  basicstyle=\scriptsize\ttfamily,
  breaklines=true
]
You are a virtual assistant that answers multiple-choice questions with the correct option only. 

Question: {{question}}

Choices: 
A. {{options_A}}
B. {{options_B}}
C. {{options_C}}
D. {{options_D}}
Answer: 
\end{lstlisting}
\begin{lstlisting}[
  caption=Listing 2: UHURA Prompt 2,
  basicstyle=\scriptsize\ttfamily,
  breaklines=true
]
Choose the correct option that answers the question below:

Question: {{question}}

Choices:
A. {{options_A}}
B. {{options_B}}
C. {{options_C}}
D. {{options_D}}
Answer: . 
\end{lstlisting}
\begin{lstlisting}[
  caption=Listing 3: UHURA Prompt 3,
  basicstyle=\scriptsize\ttfamily,
  breaklines=true
]
Answer the following multiple-choice question by picking 'A', 'B', 'C', or 'D' 

Question: {{question}}

Options: 
A. {{options_A}}
B. {{options_B}}
C. {{options_C}}
D. {{options_D}}
Answer: 
\end{lstlisting}
\begin{lstlisting}[
  caption=Listing 4: UHURA Prompt 4,
  basicstyle=\scriptsize\ttfamily,
  breaklines=true
]
Question: {{question}}

Options: 
A. {{options_A}}
B. {{options_B}}
C. {{options_C}}
D. {{options_D}}
Answer: 
\end{lstlisting}
\begin{lstlisting}[
  caption=Listing 5: UHURA Prompt 5,
  basicstyle=\scriptsize\ttfamily,
  breaklines=true
]
Which of the following options answers this question: {{question}}

Options: 
A. {{options_A}}
B. {{options_B}}
C. {{options_C}}
D. {{options_D}}
Answer: 
\end{lstlisting}
\noindent\textbf{MMLU prompts:}
\begin{lstlisting}[
  caption=Listing 1: OpenAIMMLU Prompt 1,
  basicstyle=\scriptsize\ttfamily,
  breaklines=true
]
Q: {{Question}}
A: {{A}}
B: {{B}}
C: {{C}}
D: {{D}}
Please choose the correct answer from the options above
\end{lstlisting}
\begin{lstlisting}[
  caption=Listing 2: OpenAIMMLU Prompt 2,
  basicstyle=\scriptsize\ttfamily,
  breaklines=true
]
Question: {{Question}}
1: {{A}}
2: {{B}}
3: {{C}}
4: {{D}}
Please select the correct answer from the given choices
\end{lstlisting}
\begin{lstlisting}[
  caption=Listing 3: OpenAIMMLU Prompt 3,
  basicstyle=\scriptsize\ttfamily,
  breaklines=true
]
Input Question: {{Question}}
Option A: {{A}}
Option B: {{B}}
Option C: {{C}}
Option D: {{D}}
Please indicate the correct option from the list above
\end{lstlisting}
\begin{lstlisting}[
  caption=Listing 4: OpenAIMMLU Prompt 4,
  basicstyle=\scriptsize\ttfamily,
  breaklines=true
]
Critically analyze the question and select the most probable answer from the list:
{{Question}}
Choices:
A) {{A}}
B) {{B}}
C) {{C}}
D) {{D}}
\end{lstlisting}
\begin{lstlisting}[
  caption=Listing 5: OpenAIMMLU Prompt 5,
  basicstyle=\scriptsize\ttfamily,
  breaklines=true
]
Answer the question and pick the correct answer from the options:
{{Question}}
Options:
A. {{A}}
B. {{B}}
C. {{C}}
D. {{D}}
Please choose the correct option from the above list
\end{lstlisting}
\begin{lstlisting}[
  caption=Listing 6: AfriMMLU Prompt 1,
  basicstyle=\scriptsize\ttfamily,
  breaklines=true
]
You are a highly knowledgeable and intelligent artificial intelligence model answers multiple-choice questions about {{subject}}. 

Question: {{question}} 
Choices: 
A: {{options_A}}
B: {{options_B}} 
C: {{options_C}} 
D: {{options_D}} 

Answer:
\end{lstlisting}
\begin{lstlisting}[
  caption=Listing 7: AfriMMLU Prompt 2,
  basicstyle=\scriptsize\ttfamily,
  breaklines=true
]
As an expert in {{subject}}, choose the most accurate answer to the question below. Your goal is to select the correct option 'A', 'B', 'C', or 'D' by understanding the nuances of the topic.

Question: {{question}}
Choices: 
A: {{options_A}}
B: {{options_B}} 
C: {{options_C}} 
D: {{options_D}} 

Answer: 
\end{lstlisting}
\begin{lstlisting}[
  caption=Listing 8: AfriMMLU Prompt 3,
  basicstyle=\scriptsize\ttfamily,
  breaklines=true
]
You are a subject matter expert in {{subject}}. Utilizing your expertise in {{subject}}, answer the following multiple-choice question by picking 'A', 'B', 'C', or 'D'.

Question: {{question}} 
Choices: 
A: {{options_A}}
B: {{options_B}} 
C: {{options_C}} 
D: {{options_D}} 

Answer:
\end{lstlisting}
\begin{lstlisting}[
  caption=Listing 9: AfriMMLU Prompt 4,
  basicstyle=\scriptsize\ttfamily,
  breaklines=true
]
Analyze each question critically and determine the most correct option based on your understanding of the subject matter
Question: {{question}} 
Choices: 
A: {{options_A}}
B: {{options_B}} 
C: {{options_C}} 
D: {{options_D}} 

Answer:
\end{lstlisting}
\begin{lstlisting}[
  caption=Listing 10: AfriMMLU Prompt 5,
  basicstyle=\scriptsize\ttfamily,
  breaklines=true
]
Given your proficiency in {{subject}}, please answer the subsequent multiple-choice question 
Question: {{question}} 
Choices: 
A: {{options_A}}
B: {{options_B}} 
C: {{options_C}} 
D: {{options_D}} 

Answer:
\end{lstlisting}
\subsection{Reasoning}
\label{sec:QA}
\noindent\textbf{Math prompts: from \textsc{IrokoBench}~\cite{Adelani2024IrokoBenchAN}}
\begin{lstlisting}[
  caption=Listing 1: AfriMGSM Prompt 1,
  basicstyle=\scriptsize\ttfamily,
  breaklines=true
]
{{question}} 
Step-by-step Answer: 
\end{lstlisting}
\begin{lstlisting}[
  caption=Listing 2: AfriMGSM Prompt 2,
  basicstyle=\scriptsize\ttfamily,
  breaklines=true
]
Give direct numerical answers for the question provided.

Question: {{question}} 
Step-by-step Answer: 
\end{lstlisting}
\begin{lstlisting}[
  caption=Listing 3: AfriMGSM Prompt 3,
  basicstyle=\scriptsize\ttfamily,
  breaklines=true
]
Solve the following math question 

Question: {{question}} 
Step-by-step Answer: 
\end{lstlisting}
\begin{lstlisting}[
  caption=Listing 4: AfriMGSM Prompt 4,
  basicstyle=\scriptsize\ttfamily,
  breaklines=true
]
Answer the given question with the appropriate numerical value, ensuring that the response is clear and without any supplementary information. 

Question: {{question}} 
Step-by-step Answer: 
\end{lstlisting}
\begin{lstlisting}[
  caption=Listing 5: AfriMGSM Prompt 5,
  basicstyle=\scriptsize\ttfamily,
  breaklines=true
]
For mathematical questions provided in {{language}} language. Supply the accurate numeric step by step answer to the provided question. 

Question: {{question}} 
Step-by-step Answer: 
\end{lstlisting}
\subsection{Text Generation}
\label{sec:NLG}
\noindent\textbf{Machine Translation prompts}
\begin{lstlisting}[
  caption=Listing 1: Machine Translation Prompt 1,
  basicstyle=\scriptsize\ttfamily,
  breaklines=true
]
{{source_lang}} sentence: {{source_text}} 
{{arget_lang}} sentence: 
\end{lstlisting}
\begin{lstlisting}[
  caption=Listing 2: Machine Translation Prompt 2,
  basicstyle=\scriptsize\ttfamily,
  breaklines=true
]
You are a translation expert. Translate the following {{source_lang}} sentences to {{target_lang}}

{{source_lang}} sentence: {{source_text}} 
{{target_lang}} sentence: 
\end{lstlisting}
\begin{lstlisting}[
  caption=Listing 3: Machine Translation Prompt 3,
  basicstyle=\scriptsize\ttfamily,
  breaklines=true
]
As a {{source_lang}} and {{target_lang}} linguist, translate the following {{source_lang}} sentences to {{target_lang}}.

{{source_lang}} sentence: {{source_text}} 
{{target_lang}} sentence: 
\end{lstlisting}

\bigskip
\noindent\textbf{Summarization prompts}
\begin{lstlisting}[
  caption=Listing 1: XL-SUM Prompt 1,
  basicstyle=\scriptsize\ttfamily,
  breaklines=true
]
Provide a summary of the document written in {{language}}. Ensure that you provide the summary in {{language}} and nothing else. 

Document in {{language}}: {{text}} 

Summary: 
\end{lstlisting}
\begin{lstlisting}[
  caption=Listing 2: XL-SUM Prompt 2,
  basicstyle=\scriptsize\ttfamily,
  breaklines=true
]
Summarize the document below in triple backticks and return only the summary and nothing else. 

{{text}}
\end{lstlisting}
\begin{lstlisting}[
  caption=Listing 3: XL-SUM Prompt 3,
  basicstyle=\scriptsize\ttfamily,
  breaklines=true
]
You are an advanced Summarizer, a specialized assistant designed to summarize documents in {{language}}. Your main goal is to ensure summaries are concise and informative.
Ensure you return the summary only and nothing else. 

Document: {{text}} 

Summary:
\end{lstlisting}
\noindent\textbf{Diacritics Restoration prompts}
\begin{lstlisting}[
  caption=Listing 1: AFRIADR  Prompt 1,
  basicstyle=\scriptsize\ttfamily,
  breaklines=true
]
Please restore the missing diacritics in the following sentence: {{text}}. 
Return output sentence only
\end{lstlisting}
\begin{lstlisting}[
  caption=Listing 2: AFRIADR  Prompt 2,
  basicstyle=\scriptsize\ttfamily,
  breaklines=true
]
Given a sentence without diacritics, add the appropriate diacritics to make it grammatically and semantically correct. 
Sentence: {{text}}. 
Return output sentence only
\end{lstlisting}
\begin{lstlisting}[
  caption=Listing 3: AFRIADR  Prompt 3,
  basicstyle=\scriptsize\ttfamily,
  breaklines=true
]
This text is in {{language}}. Restore all diacritical marks to their proper places in the following sentence: {{text}}. Return output sentence only
\end{lstlisting}
\begin{lstlisting}[
  caption=Listing 4: AFRIADR  Prompt 4,
  basicstyle=\scriptsize\ttfamily,
  breaklines=true
]
You are a linguist specializing in diacritical marks for {{language}}. Add the appropriate diacritics to this {{language}} sentence: {{text}}. Return output sentence only
\end{lstlisting}
\begin{lstlisting}[
  caption=Listing 5: AFRIADR  Prompt 5,
  basicstyle=\scriptsize\ttfamily,
  breaklines=true
]
You are a linguist specializing in diacritical marks for {{language}}. Diacritics are essential for proper pronunciation and meaning in {{language}}. You are tasked with converting {{language}} sentences  without diacritics into their correctly accented forms. Here's the input: {{text}}. Return output sentence only
\end{lstlisting}

\clearpage
\onecolumn

\section{Detailed Results Per Language}
\label{sec:detailed_lang_results}
This appendix presents detailed per-language performance results for each dataset.  We group them by the task category shown in Figure~\ref{fig:afrobench-overview}. Each figure shows the model performance on the best prompt per language. 

\subsection{Natural Language Understanding (NLU)}
\subsubsection{POS}
\textbf{MasakhaPOS}
\begin{figure}[h]
    \centering
    \includegraphics[width=0.95\linewidth]{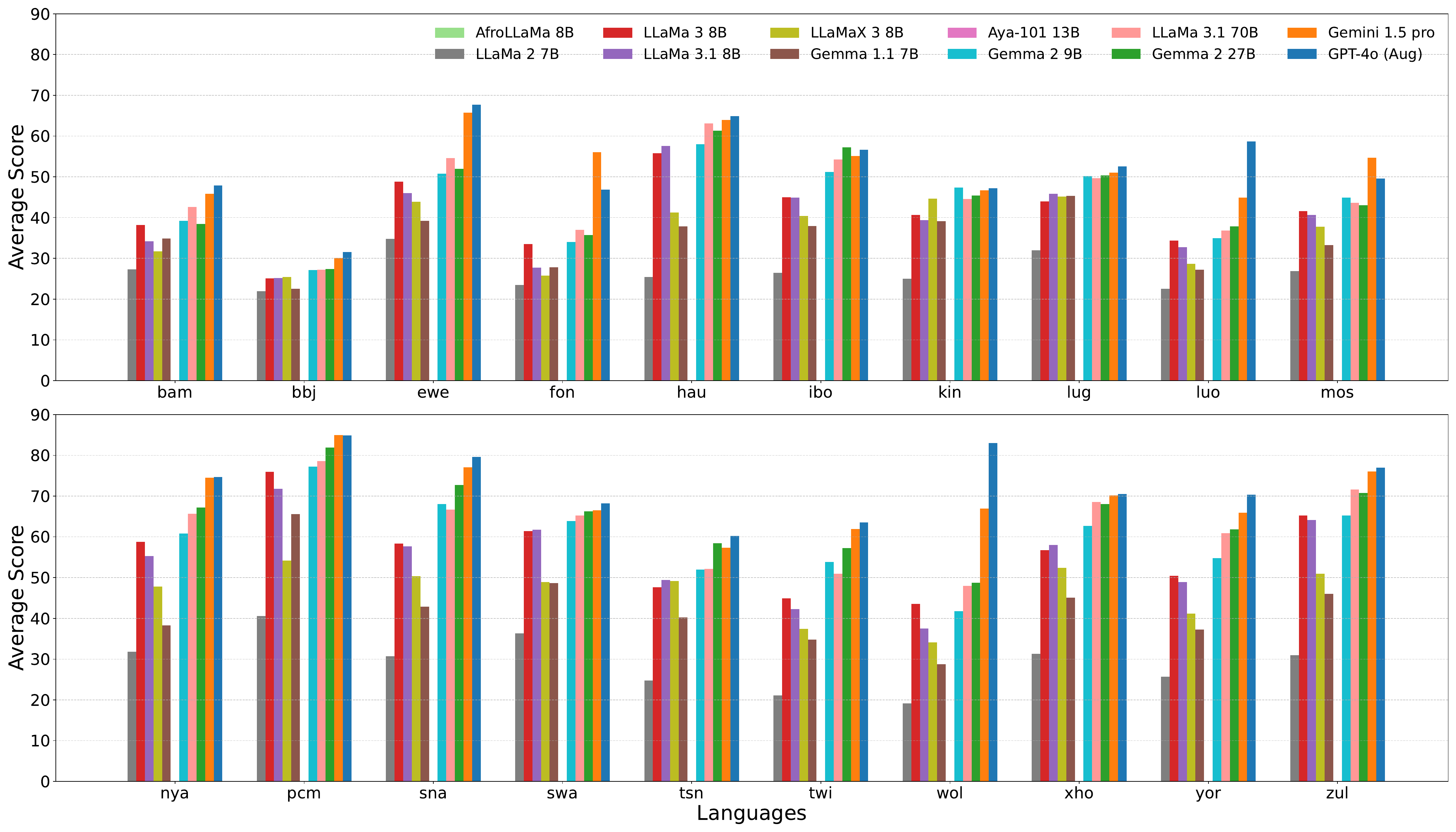}
     \vspace{-2mm}
    \caption{Per-language performance results for the MasakhaPOS dataset.}
    \label{fig:masakhapos_results}
     \vspace{-2mm}
\end{figure}

\subsubsection{NER}
\textbf{MasakhaNER}
\begin{figure}[h]
    \centering
    \includegraphics[width=0.95\linewidth]{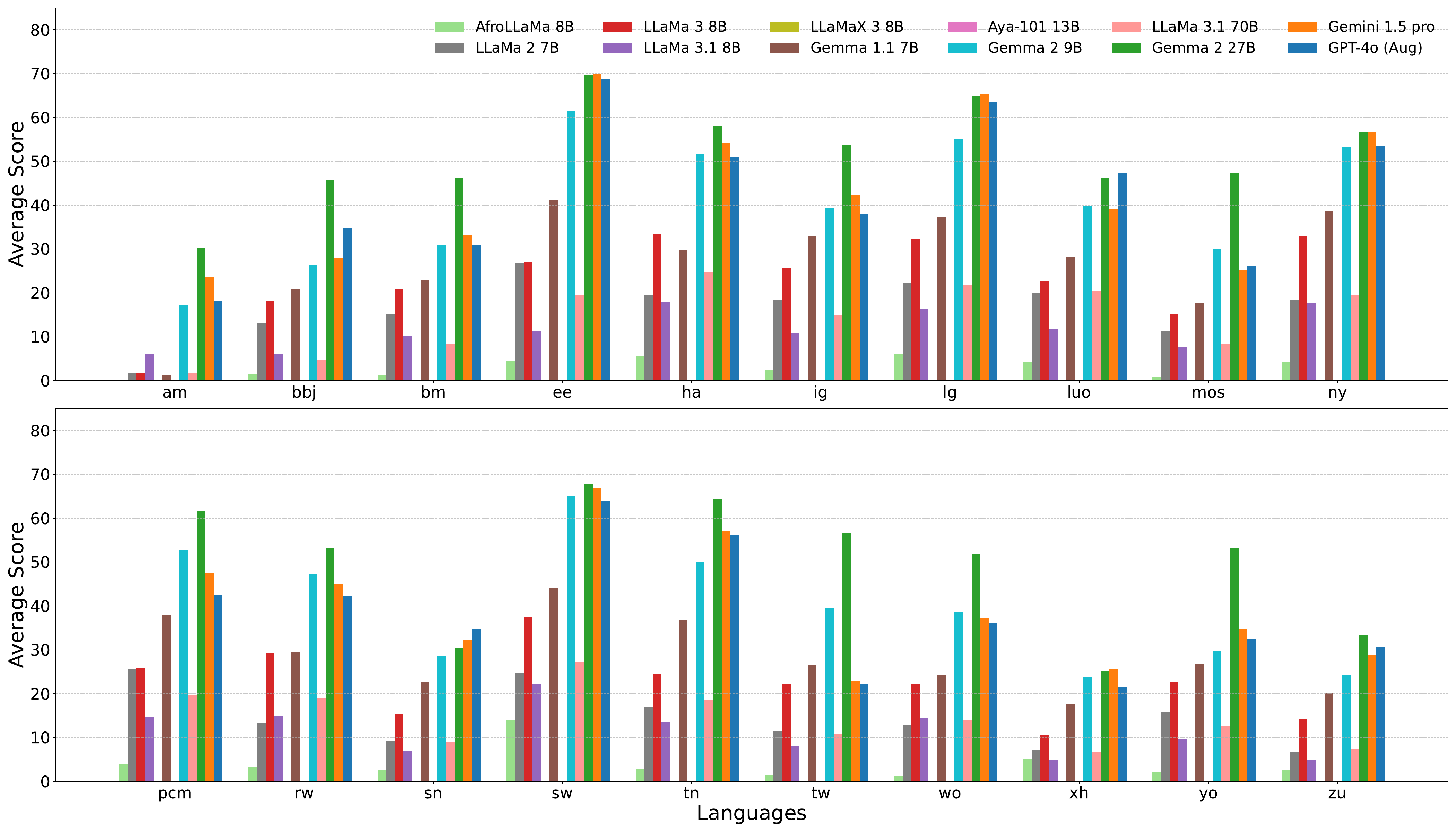}
     \vspace{-2mm}
    \caption{Per-language performance results for the MasakhaNER dataset.}
    \label{fig:masakhaner_results}
     \vspace{-2mm}
\end{figure}

\subsubsection{Sentiment Analysis}

\textbf{AfriSenti}
\begin{figure}[h]
    \centering
    \includegraphics[width=0.95\linewidth]{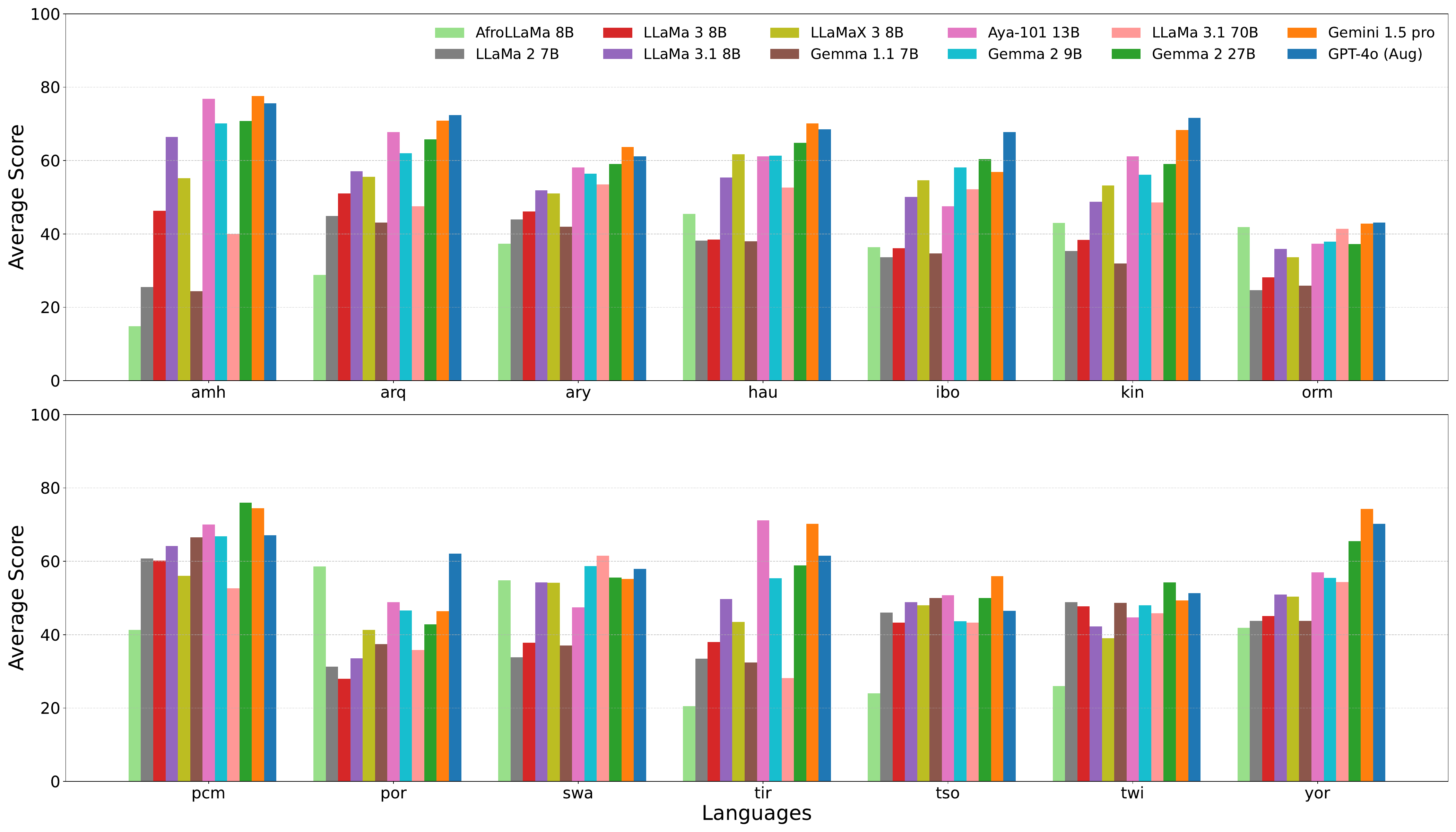}
     \vspace{-2mm}
    \caption{Per-language performance results for the AfriSenti dataset.}
    \label{fig:afrisenti_results}
     \vspace{-2mm}
\end{figure}

\textbf{NollySenti}

\begin{figure}[ht]
    \centering
    \includegraphics[width=0.95\linewidth]{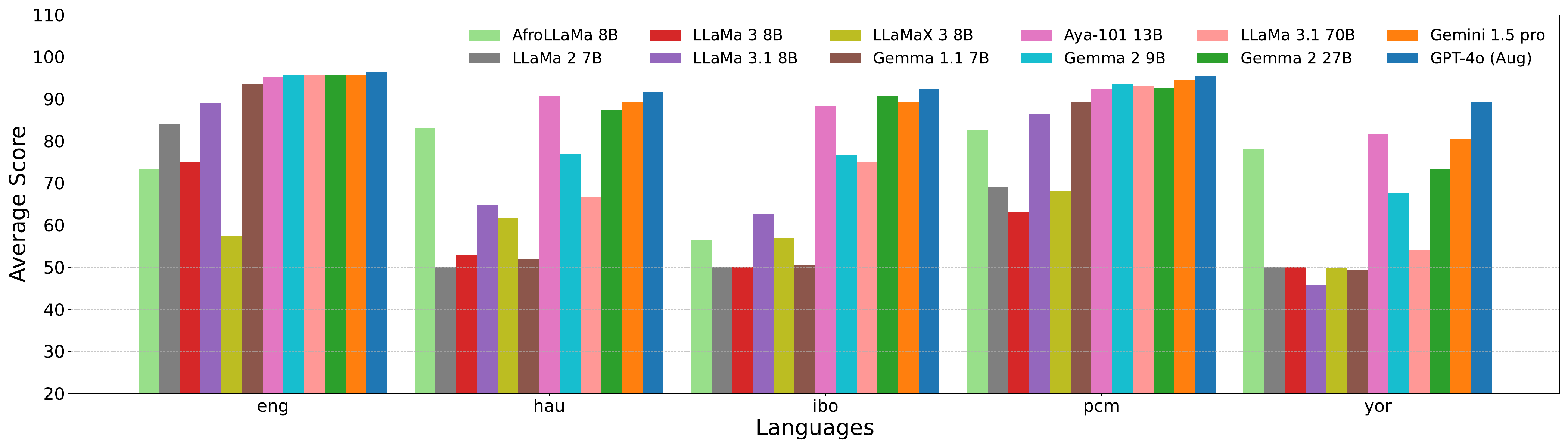}
     \vspace{-2mm}
    \caption{Per-language performance results for the NollySenti dataset.}
    \label{fig:nollysenti_results}
     \vspace{-2mm}
\end{figure}
\clearpage
\subsubsection{Intent Detection}
\textbf{Injongo Intent}

\begin{figure}[h]
    \centering
    \includegraphics[width=0.95\linewidth]{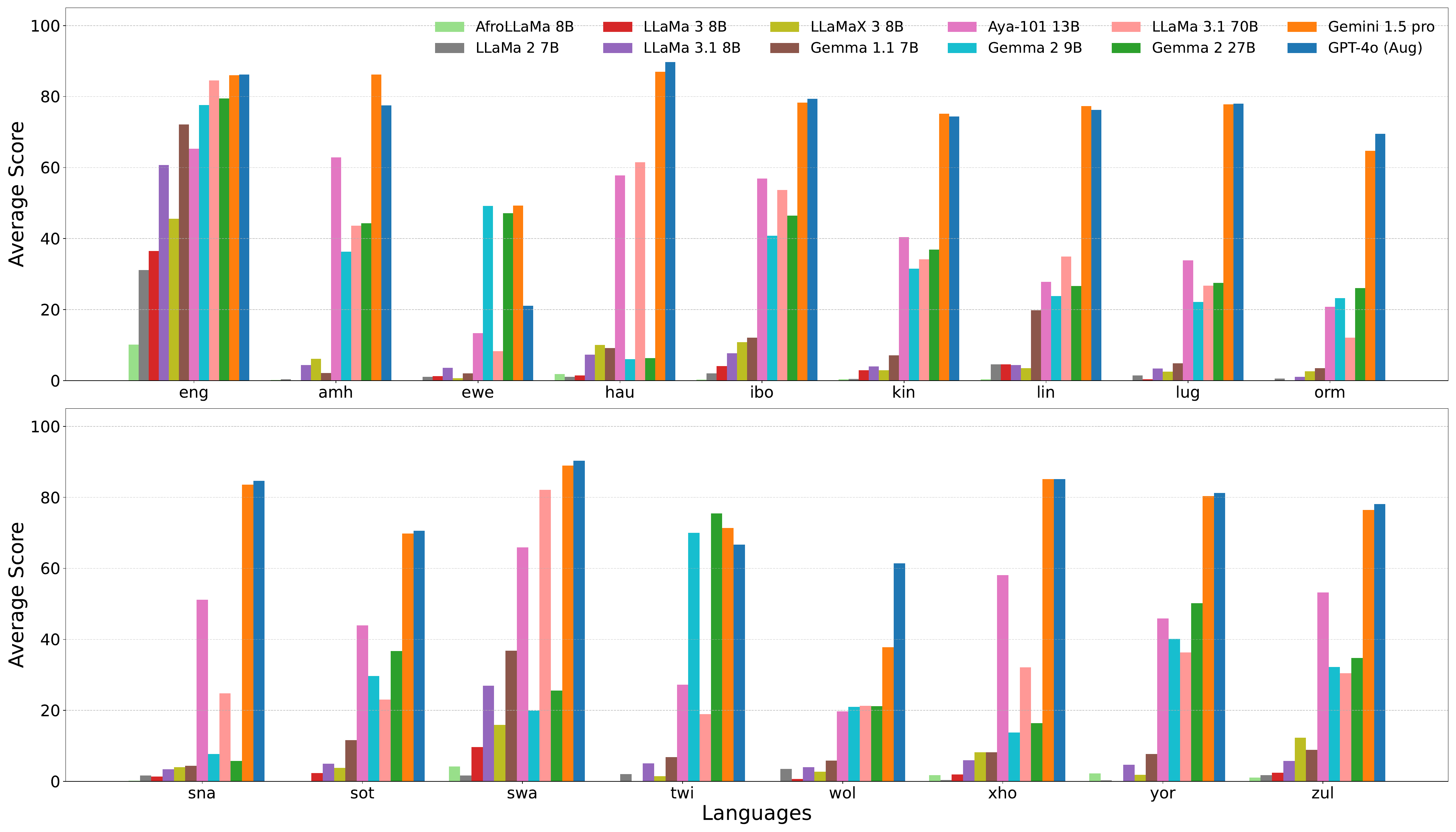}
     \vspace{-2mm}
    \caption{Per-language performance results for the InjongoIntent dataset.}
    \label{fig:injongo_results}
     \vspace{-2mm}
\end{figure}

\subsubsection{Topic Classification}
\textbf{MasakhaNEWS}

\begin{figure}[h]
    \centering
    \vspace{-4mm}
    \includegraphics[width=\linewidth]{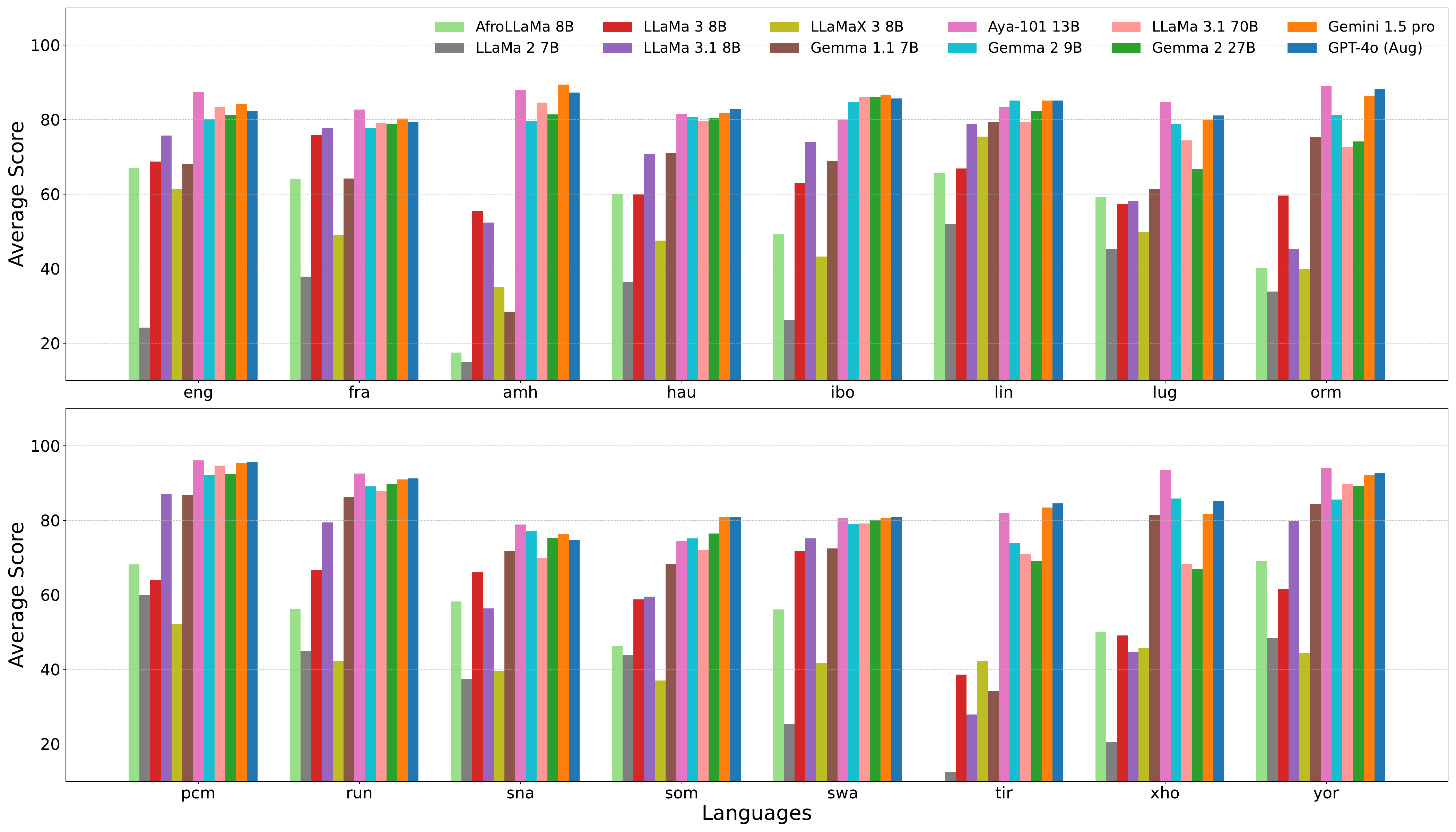}
     
    \caption{Per-language performance results for the MasakhaNEWS dataset.}
    \label{fig:masakhanews_results}
     \vspace{-2mm}
\end{figure}

\clearpage
\textbf{SIB}
\vspace{-2mm}
\begin{figure}[h]
    \centering
    \includegraphics[width=\linewidth]{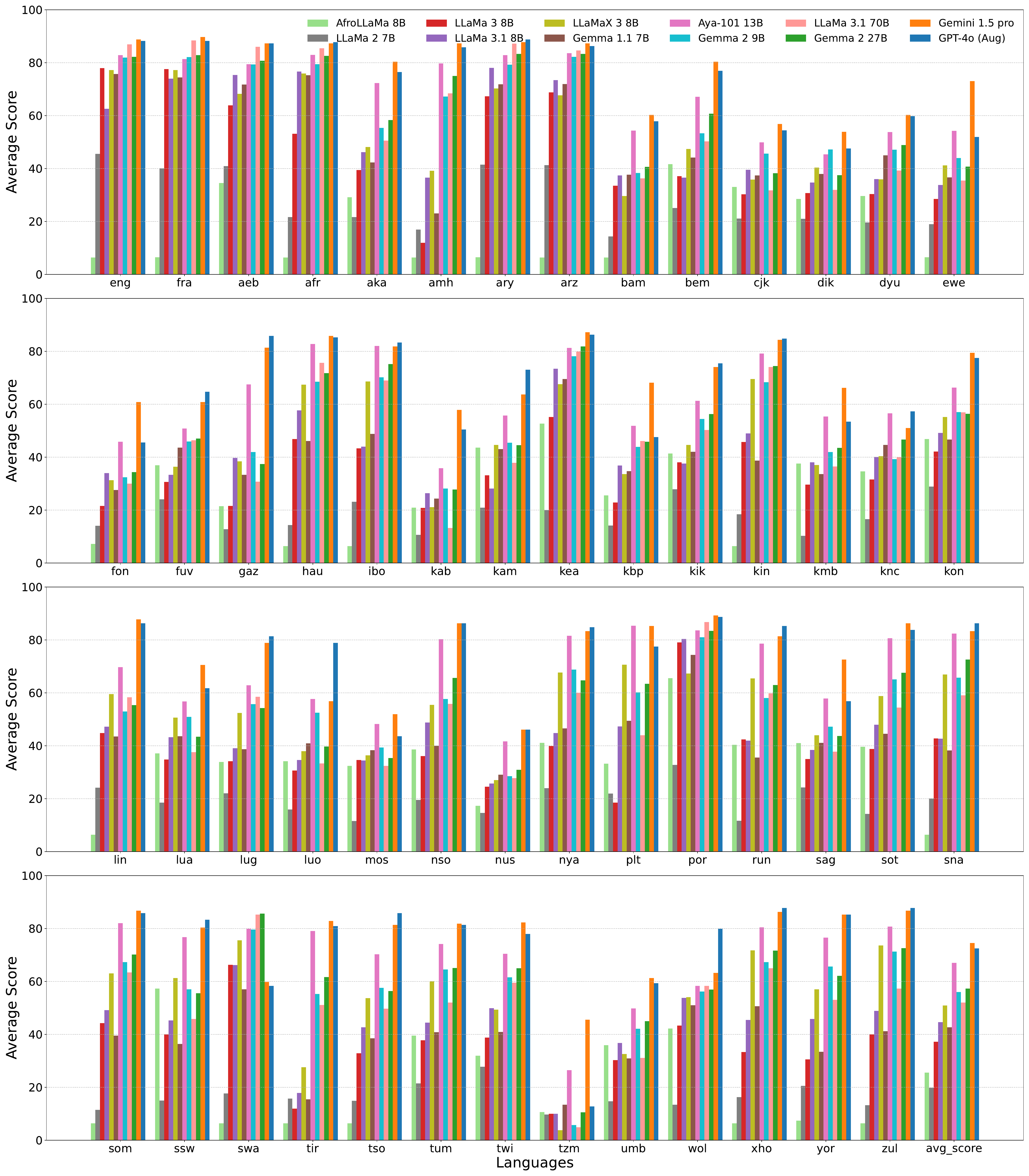}
     \vspace{-4mm}
    \caption{Per-language performance results for the SIB dataset.}
    \label{fig:sib_results}
     \vspace{-4mm}
\end{figure}

\clearpage

\subsubsection{Hate Speech:}

\textbf{AfriHate}

\begin{figure}[h]
    \centering
    \includegraphics[width=0.95\linewidth]{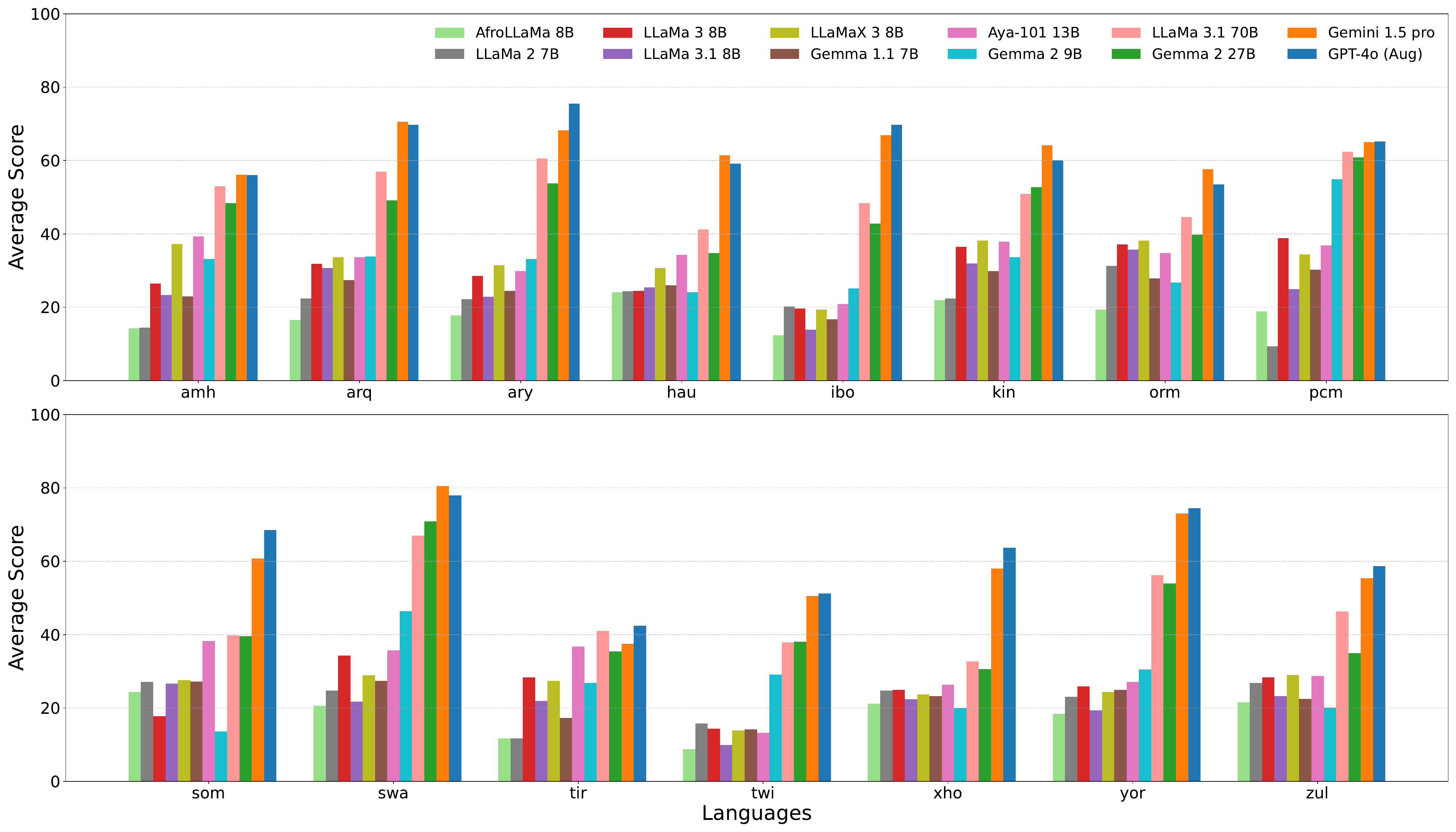}
     \vspace{-4mm}
    \caption{Per-language performance results for the AfriHate dataset.}
    \label{fig:afrihate_results}
     \vspace{-4mm}
\end{figure}

\subsection{Natural Language Inference}

\textbf{AfriXNLI}

\begin{figure}[h]
    \centering
    \includegraphics[width=0.95\linewidth]{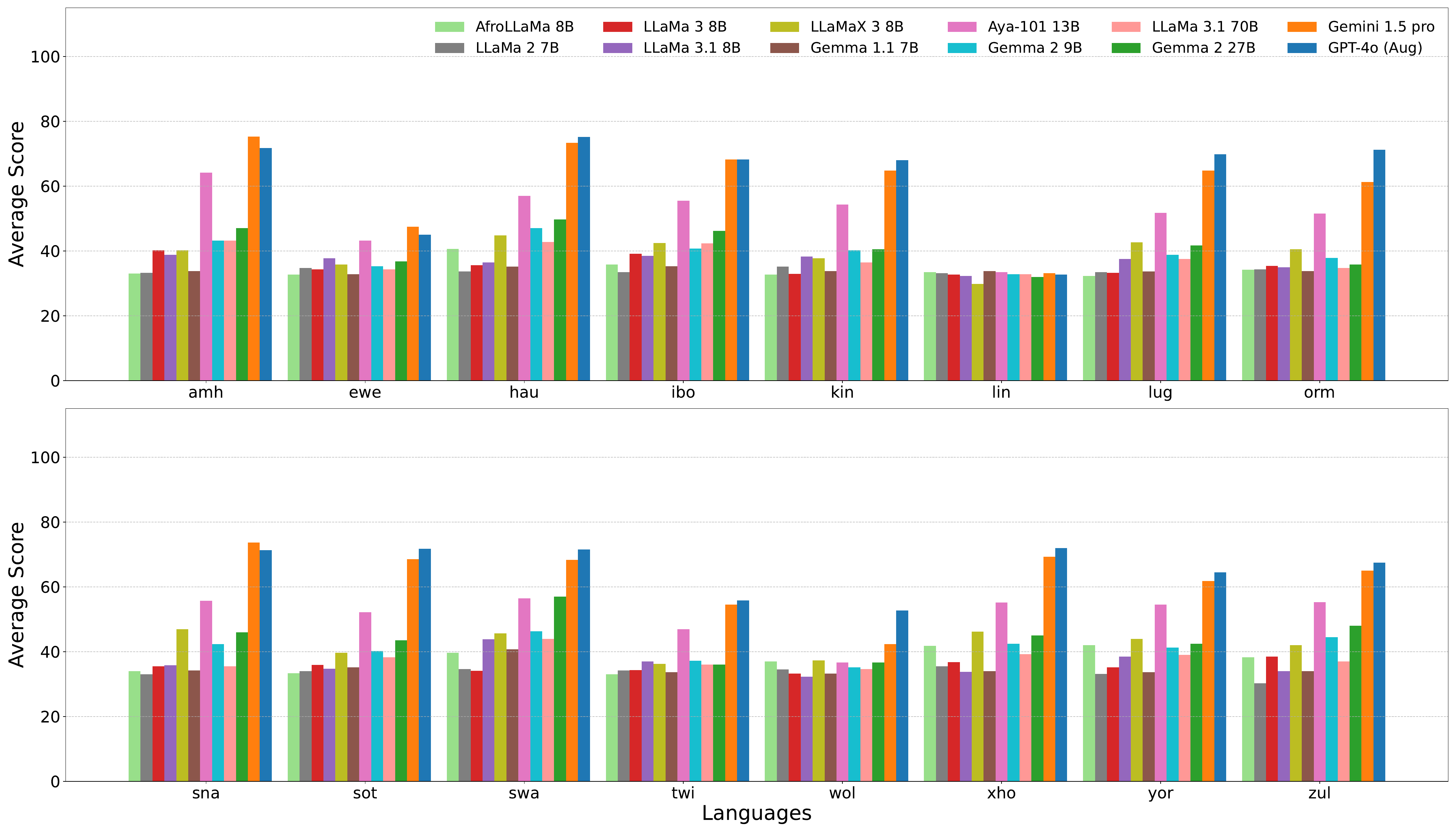}
     \vspace{-4mm}
    \caption{Per-language performance results for the \afrixnli dataset.}
    \label{fig:afrixnli_results}
     \vspace{-4mm}
\end{figure}

\clearpage
\subsection{Question Answering}

\subsubsection{Cross-lingual Question Answering}

\textbf{AfriQA}

\begin{figure}[h]
    \centering
    \includegraphics[width=0.95\linewidth]{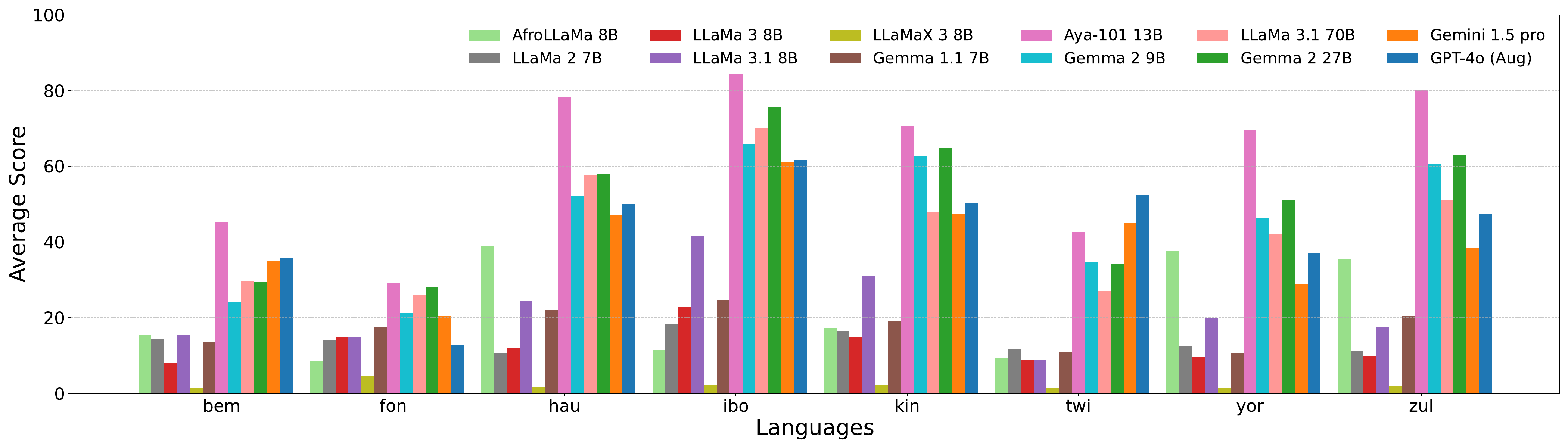}
     \vspace{-4mm}
    \caption{Per-language performance results for the \afriqa dataset.}
    \label{fig:afriqa_results}
     \vspace{-4mm}
\end{figure}

\subsubsection{Reading Comprehension}

\textbf{Belebele}

\begin{figure}[h]
    \centering
    \includegraphics[width=0.95\linewidth]{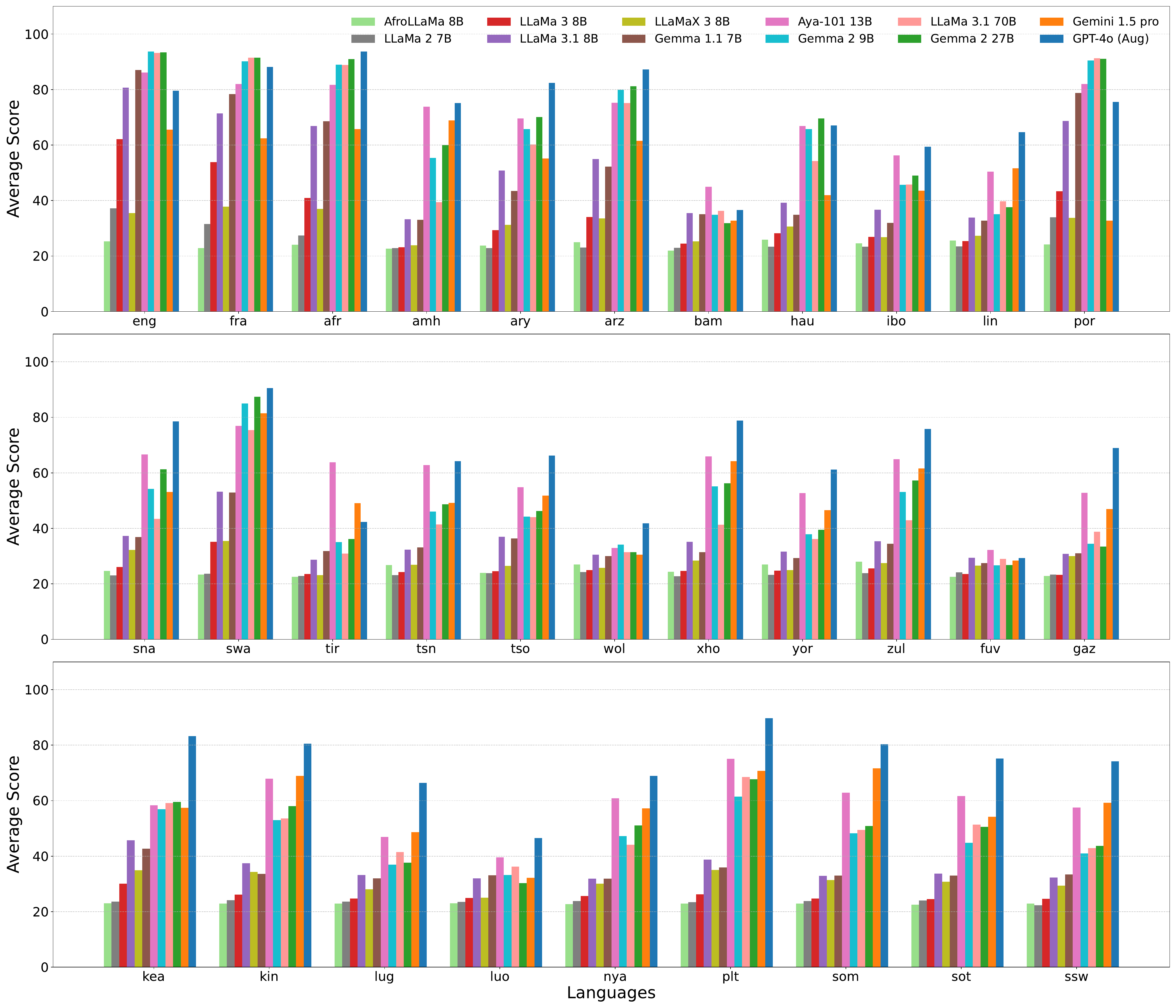}
     \vspace{-4mm}
    \caption{Per-language performance results for the \belebele dataset.}
    \label{fig:belebel_results}
     \vspace{-4mm}
\end{figure}
\clearpage
\textbf{NaijaRC}

\begin{figure}[h]
    \centering
    \includegraphics[width=0.95\linewidth]{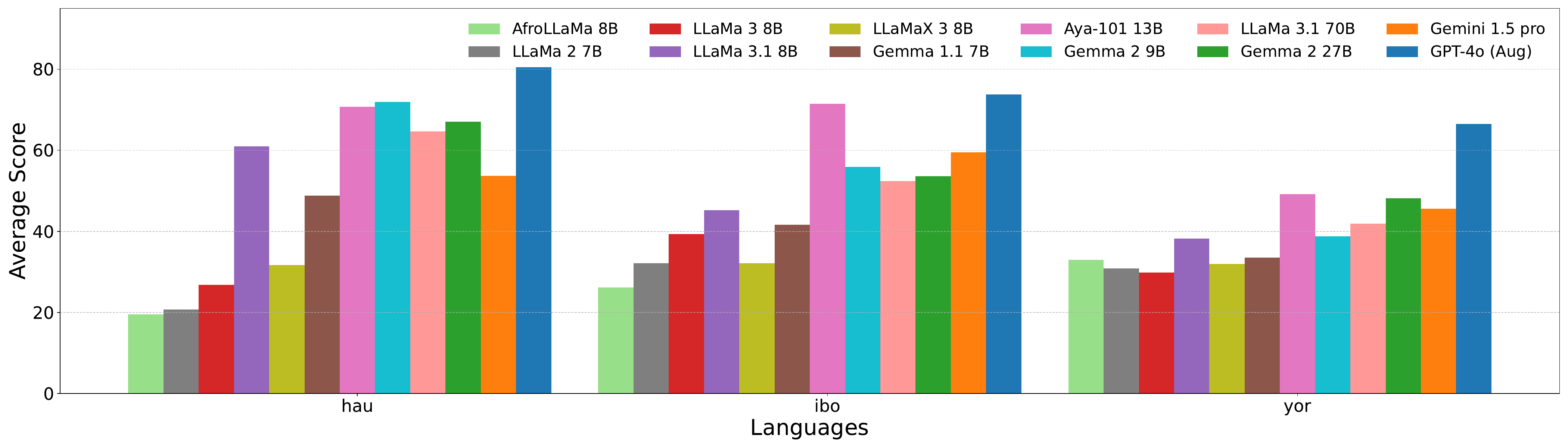}
     \vspace{-4mm}
    \caption{Per-language performance results for the \naijarc dataset.}
    \label{fig:naijarc_results}
     \vspace{-4mm}
\end{figure}

\subsection{Knowledge}
\subsubsection{Arc-E}
\textbf{\uhura}
\begin{figure}[h]
    \centering
    \includegraphics[width=0.95\linewidth]{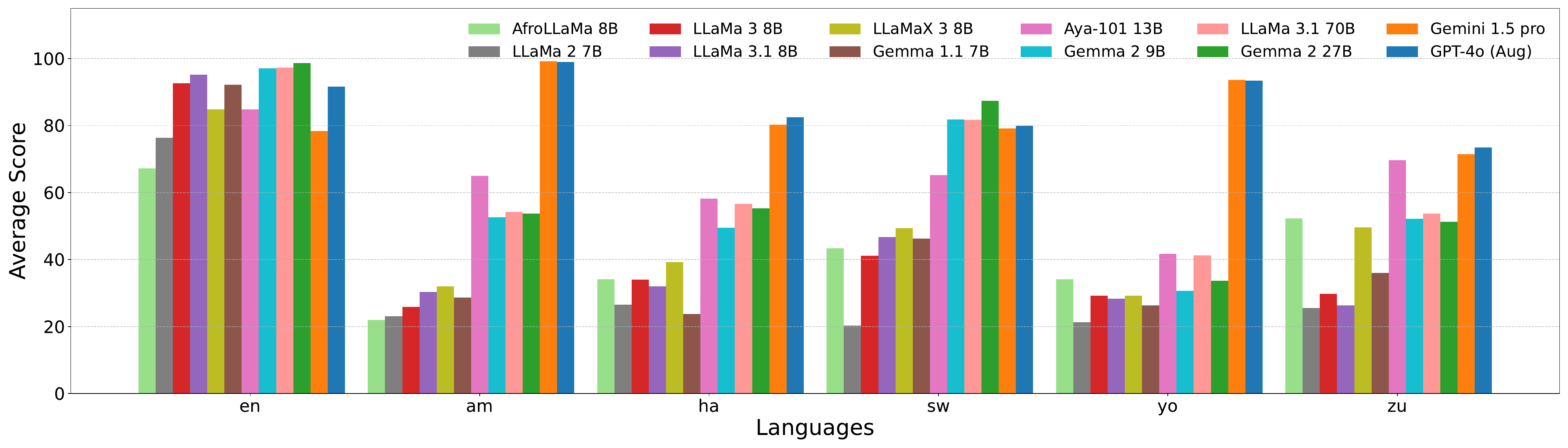}
     \vspace{-4mm}
    \caption{Per-language performance results for the \uhura dataset.}
    \label{fig:uhura_results}
     \vspace{-4mm}
\end{figure}

\subsubsection{MMLU}
\textbf{OpenAIMMLU}

\begin{figure}[h]
    \centering
    \includegraphics[width=0.95\linewidth]{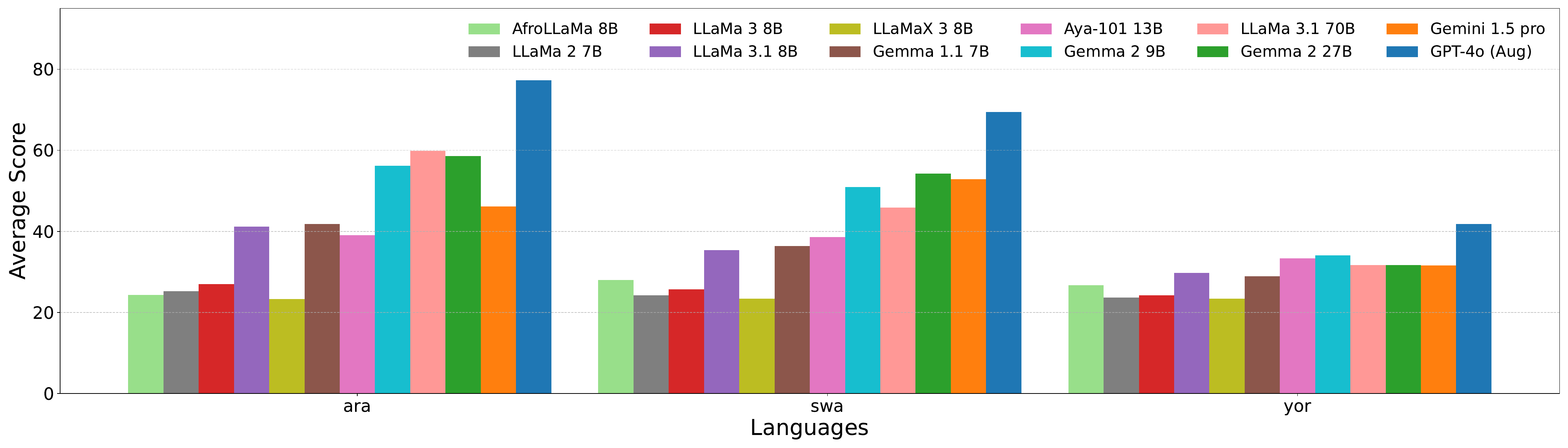}
     \vspace{-4mm}
    \caption{Per-language performance results for the \openaimmlu dataset.}
    \label{fig:openaimmlu_results}
     \vspace{-4mm}
\end{figure}

\clearpage
\textbf{AfriMMLU}

\begin{figure}[h]
    \centering
    \includegraphics[width=0.95\linewidth]{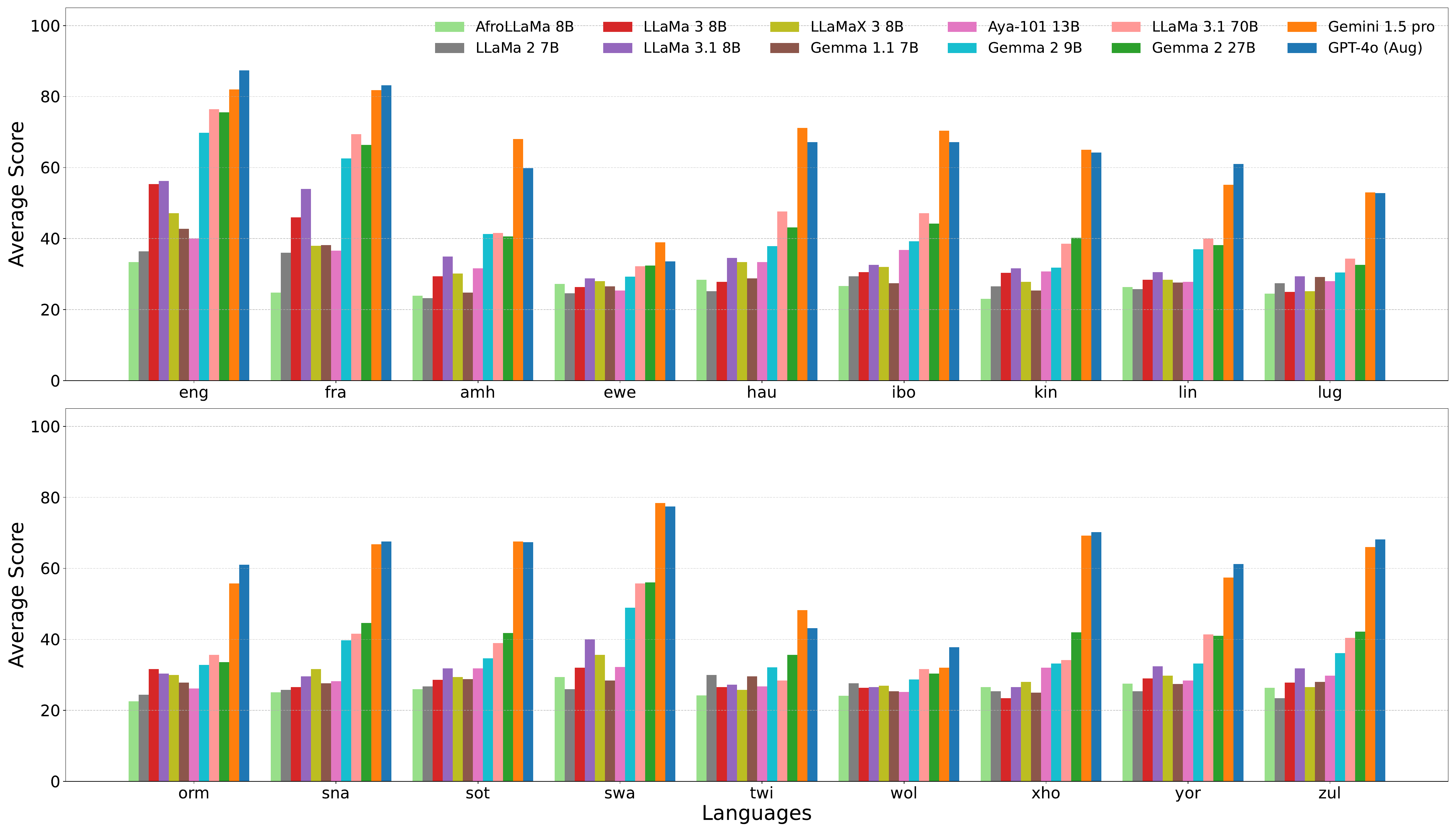}
     \vspace{-4mm}
    \caption{Per-language performance results for the \afrimmlu dataset.}
    \label{fig:afrimmlu_results}
     \vspace{-4mm}
\end{figure}

\subsection{Reasoning}

\textbf{AfriMGSM}

\begin{figure}[h]
    \centering
    \includegraphics[width=0.95\linewidth]{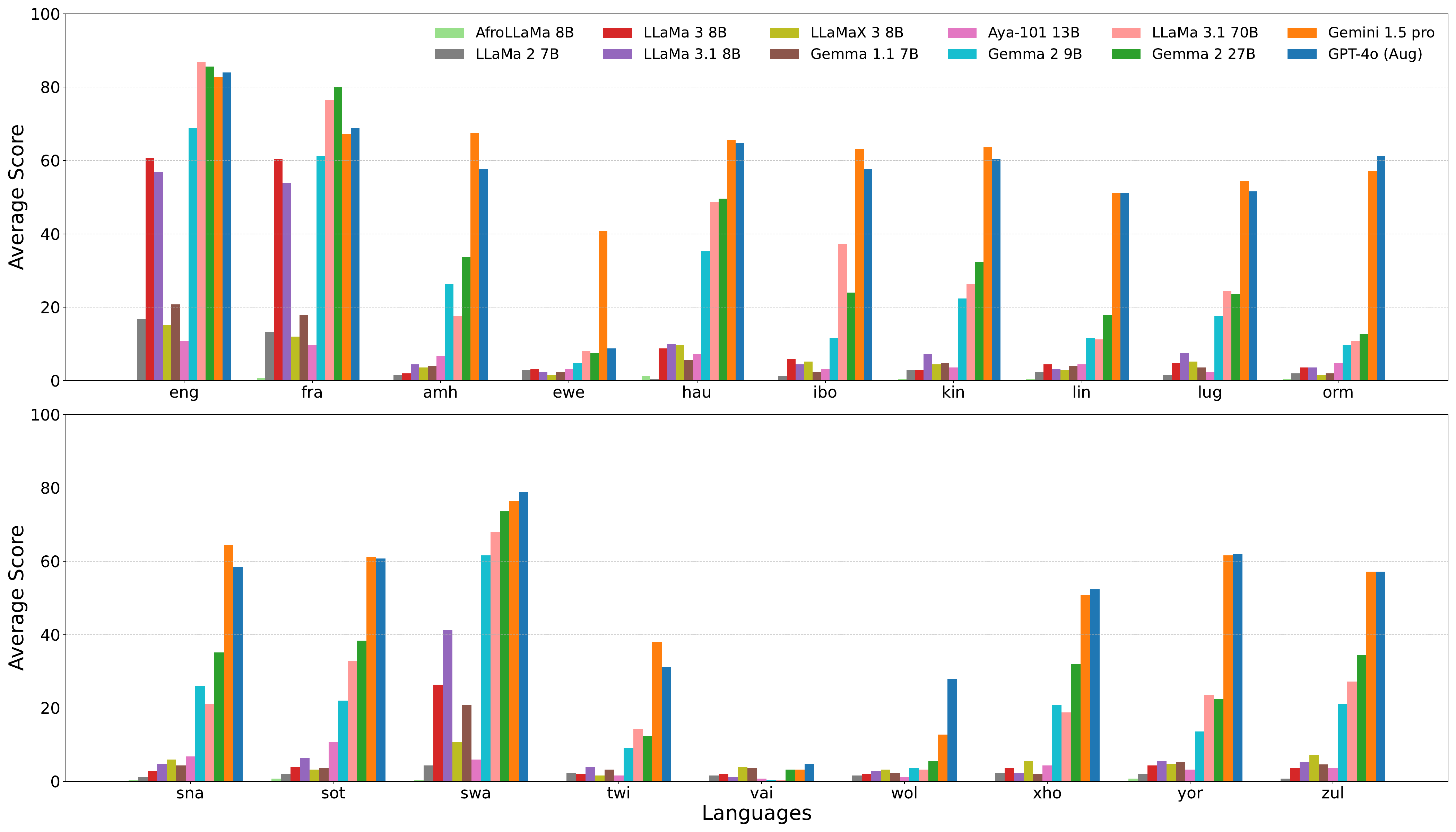}
     \vspace{-4mm}
    \caption{Per-language performance results for the \afrimgsm dataset.}
    \label{fig:afrimgsm_results}
     \vspace{-4mm}
\end{figure}
\clearpage

\subsection{Text Generation}
\subsubsection{Machine Translation}
\textbf{SALT (\textit{en/fr-xx)}}
\begin{figure}[h]
    \centering
    \includegraphics[width=0.95\linewidth]{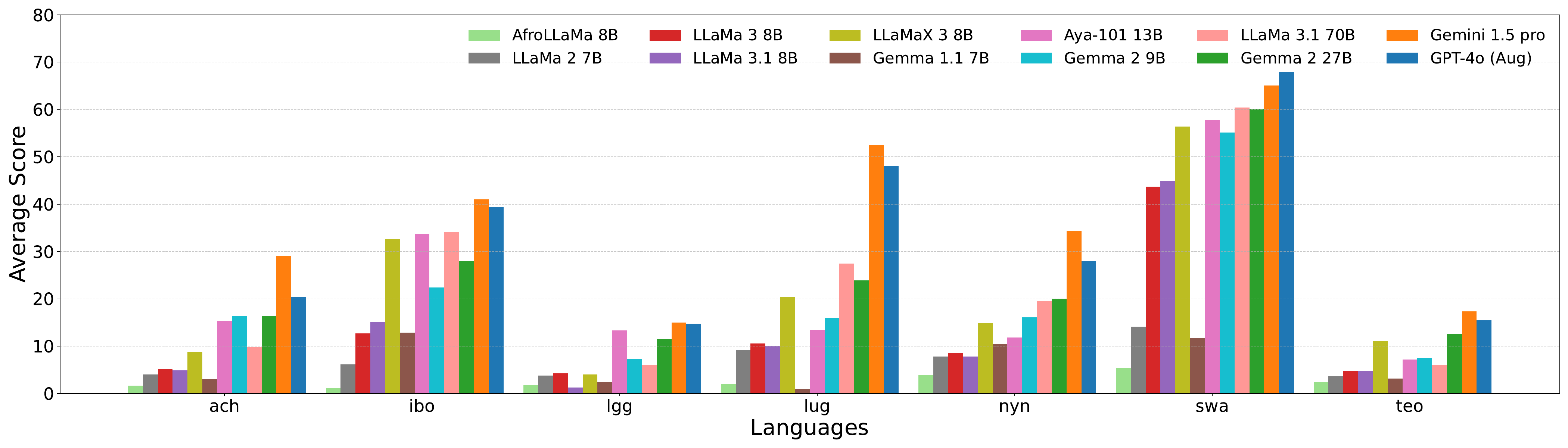}
     \vspace{-4mm}
    \caption{Per-language performance results for the \salt dataset (\textit{en/fr-xx}).}
    \label{fig:salt_enfr_xx_results}
     \vspace{-4mm}
\end{figure}

\textbf{SALT (\textit{xx-en/fr)}}

\begin{figure}[h]
    \centering
    \includegraphics[width=0.95\linewidth]{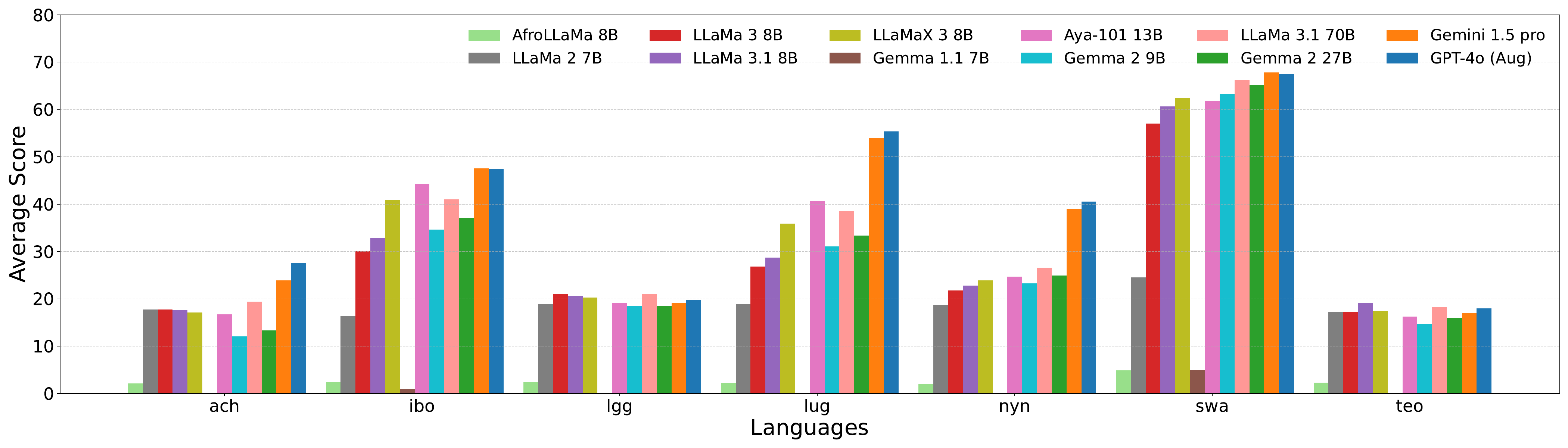}
     \vspace{-4mm}
    \caption{Per-language performance results for the \salt dataset (\textit{xx-en/fr}).}
    \label{fig:salt_xx_enfrresults}
     \vspace{-4mm}
\end{figure}

\textbf{MAFAND (\textit{en-xx/fr)}}

\begin{figure}[h]
    \centering
    \includegraphics[width=0.95\linewidth]{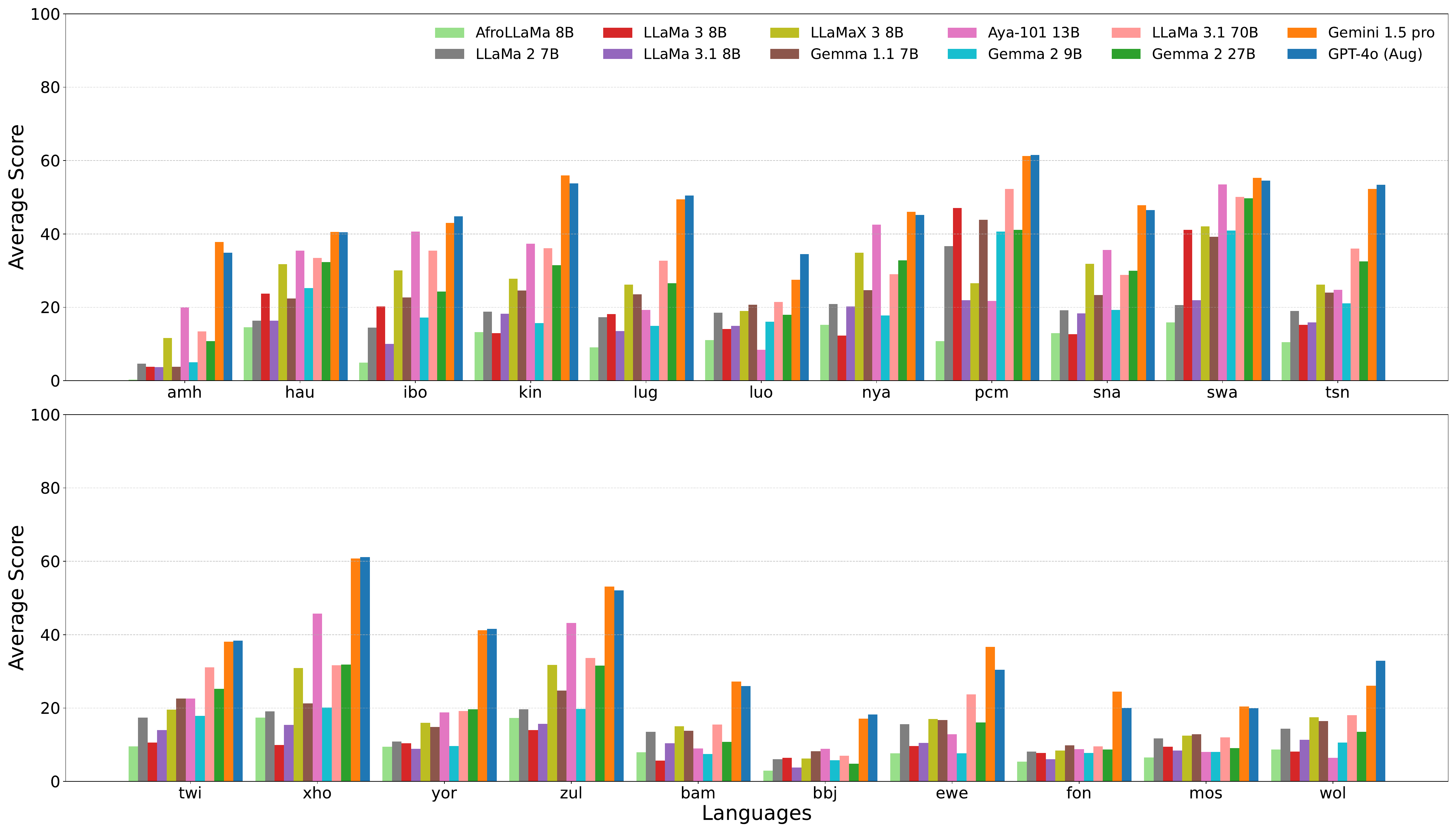}
     \vspace{-4mm}
    \caption{Per-language performance results for the \mafand dataset.}
    \label{fig:mafand_en-xxfrresults}
     \vspace{-4mm}
\end{figure}

\clearpage
\textbf{MAFAND (\textit{xx-en/fr)}}

\begin{figure}[h]
    \centering
    \includegraphics[width=0.95\linewidth]{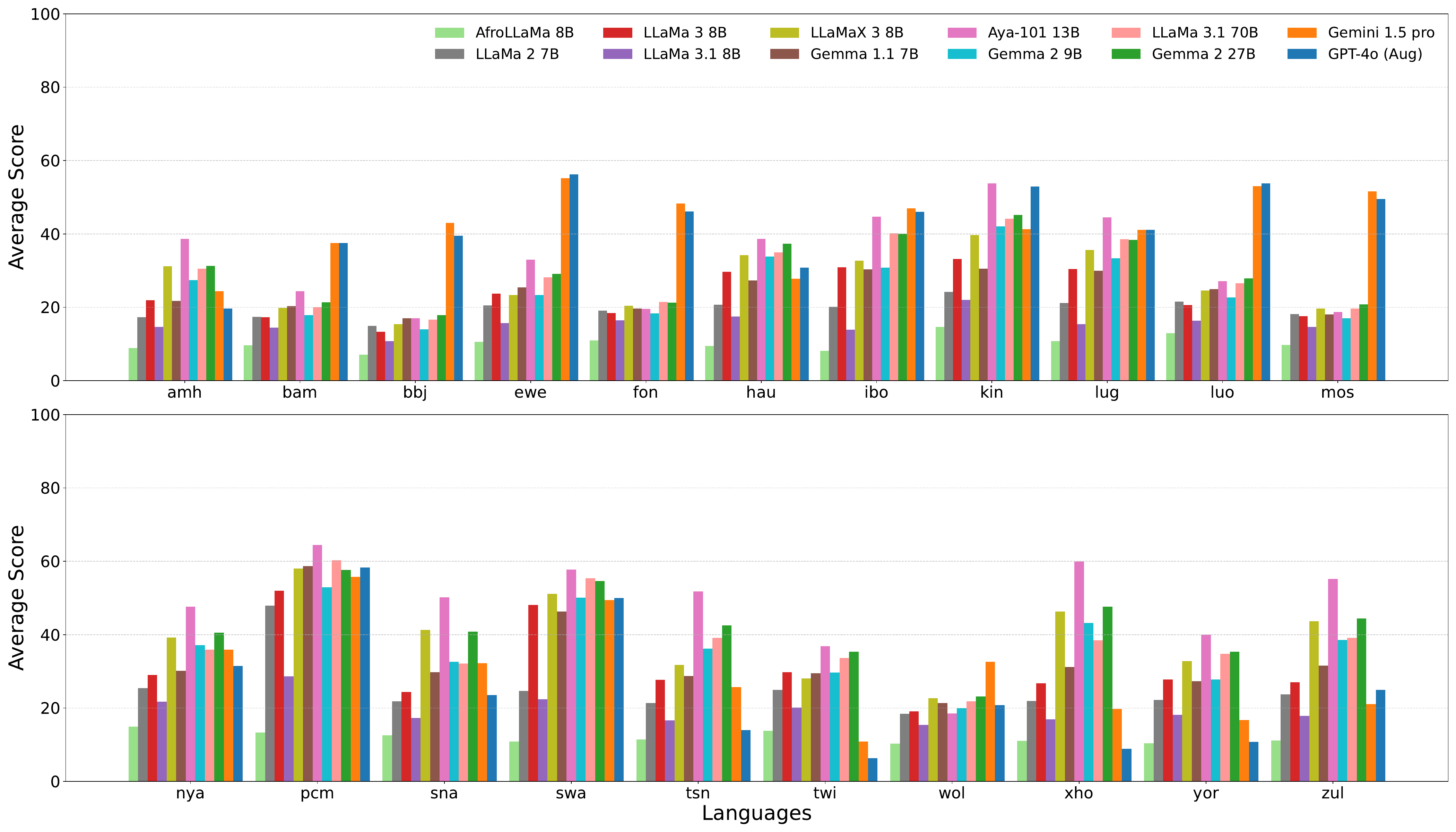}
     \vspace{-4mm}
    \caption{Per-language performance results for the \mafand dataset.}
    \label{fig:mafand_xx_enfrresults}
     \vspace{-4mm}
\end{figure}

\textbf{NTREX (\textit{en/fr-xx)}}

\begin{figure}[h]
    \centering
    \includegraphics[width=0.95\linewidth]{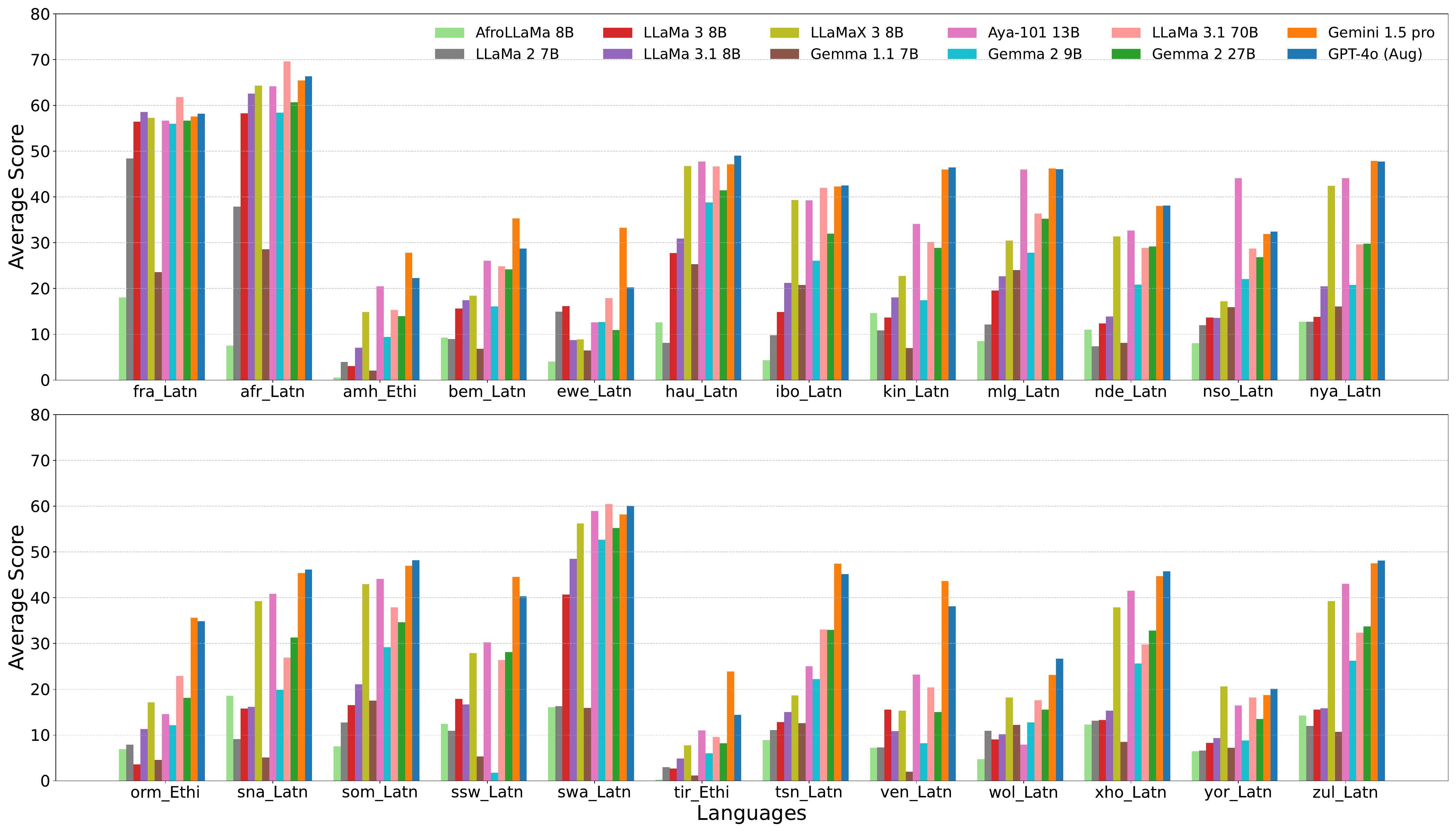}
     \vspace{-4mm}
    \caption{Per-language performance results for the \ntrex dataset (\textit{en/fr-xx}).}
    \label{fig:ntrex_enfr_xx_results}
     \vspace{-4mm}
\end{figure}

\clearpage

\textbf{NTREX (\textit{xx-en/fr)}}

\begin{figure}[h]
    \centering
    \includegraphics[width=0.95\linewidth]{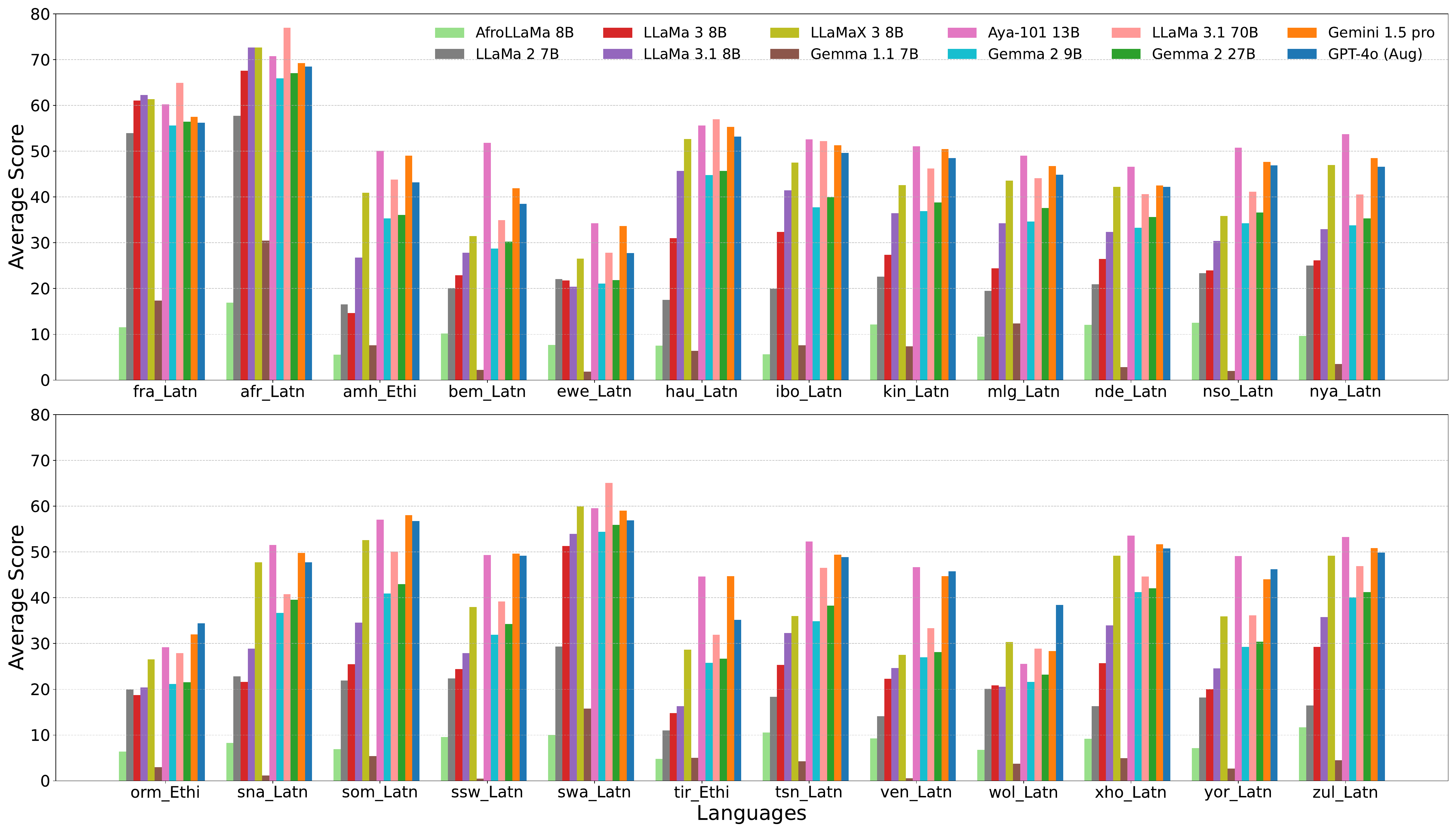}
     \vspace{-4mm}
    \caption{Per-language performance results for the \ntrex dataset (\textit{xx-en/fr}).}
    \label{fig:ntrex_xx_enfrresults}
     \vspace{-4mm}
\end{figure}

\clearpage

\textbf{Flores (African Languages only and French) (\textit{en/fr-xx)}}

\begin{figure}[h]
    \centering
    \includegraphics[width=0.95\linewidth]{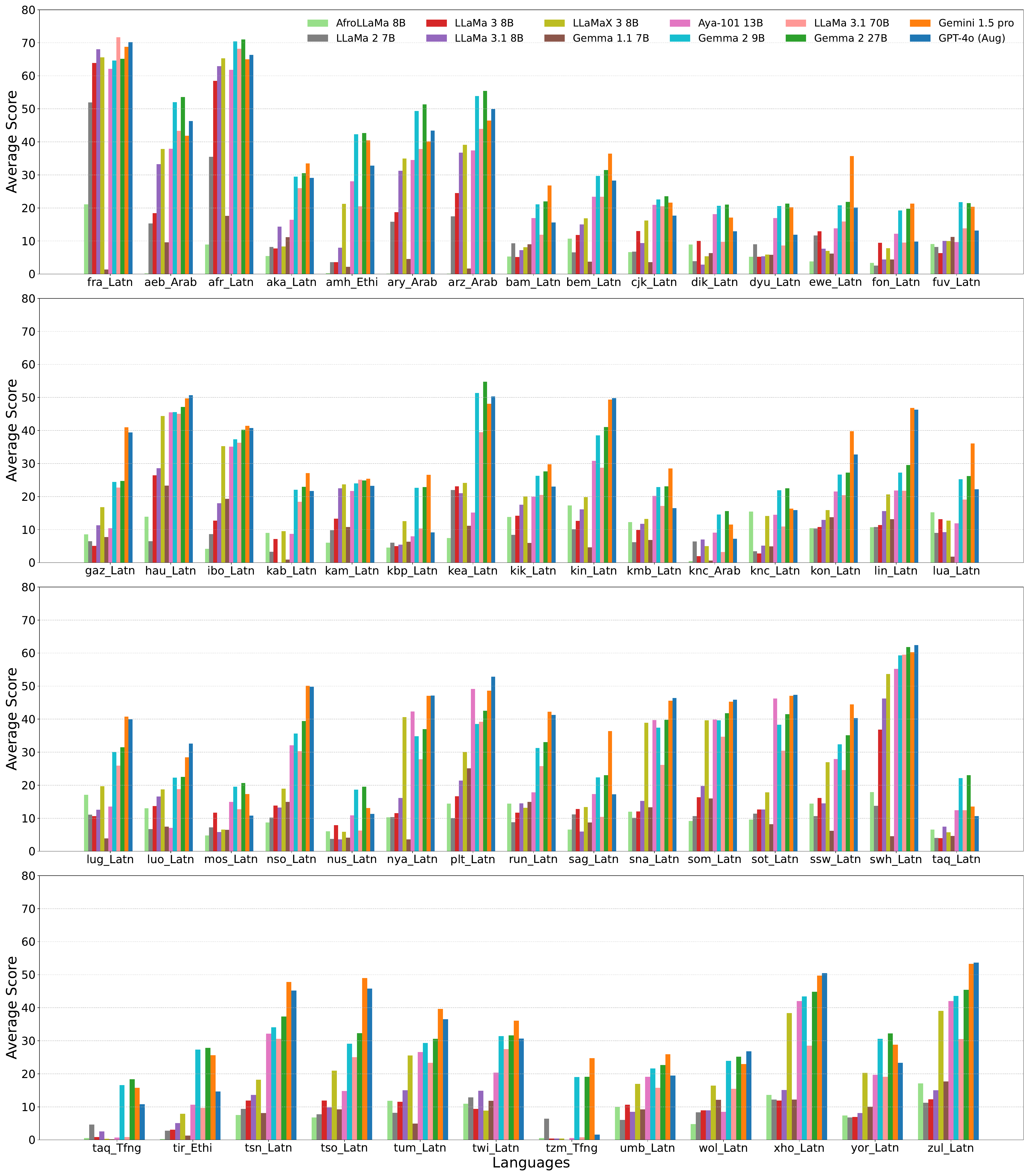 }
     \vspace{-4mm}
    \caption{Per-language performance results for the \flores dataset (\textit{en/fr-xx}).}
    \label{fig:flores_enfr_xx_results}
     \vspace{-4mm}
\end{figure}

\clearpage

\textbf{Flores (African Languages only and French) (\textit{xx-en/fr)}}
\begin{figure}[h]
    \centering
    \includegraphics[width=0.95\linewidth]{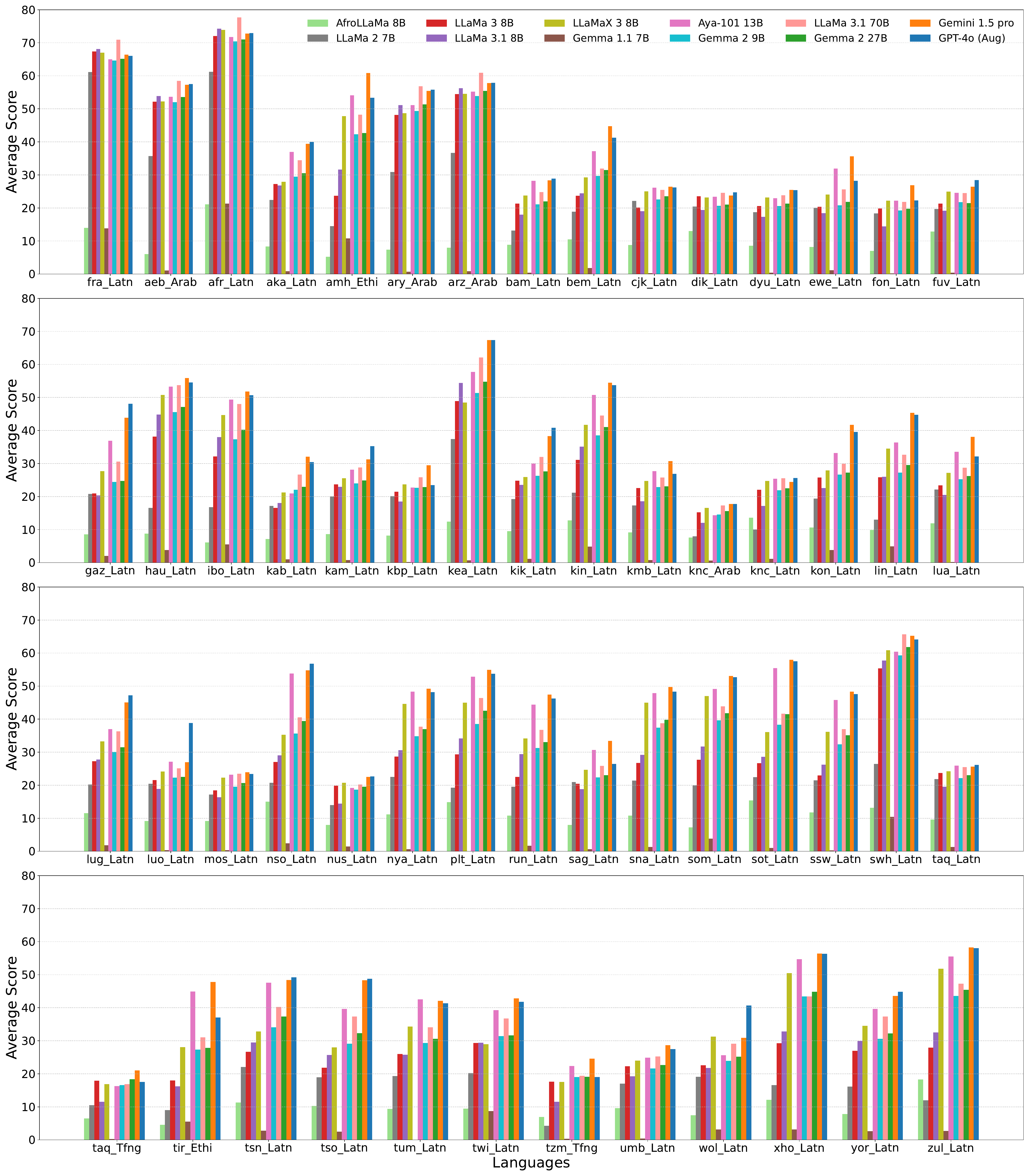}
     \vspace{-4mm}
    \caption{Per-language performance results for the \flores dataset (\textit{xx-en/fr}).}
    \label{fig:flores_xx_enfrresults}
     \vspace{-4mm}
\end{figure}

\clearpage

\subsubsection{Summarization}

\textbf{XL-SUM}
\begin{figure}[h]
    \centering
    \includegraphics[width=0.95\linewidth]{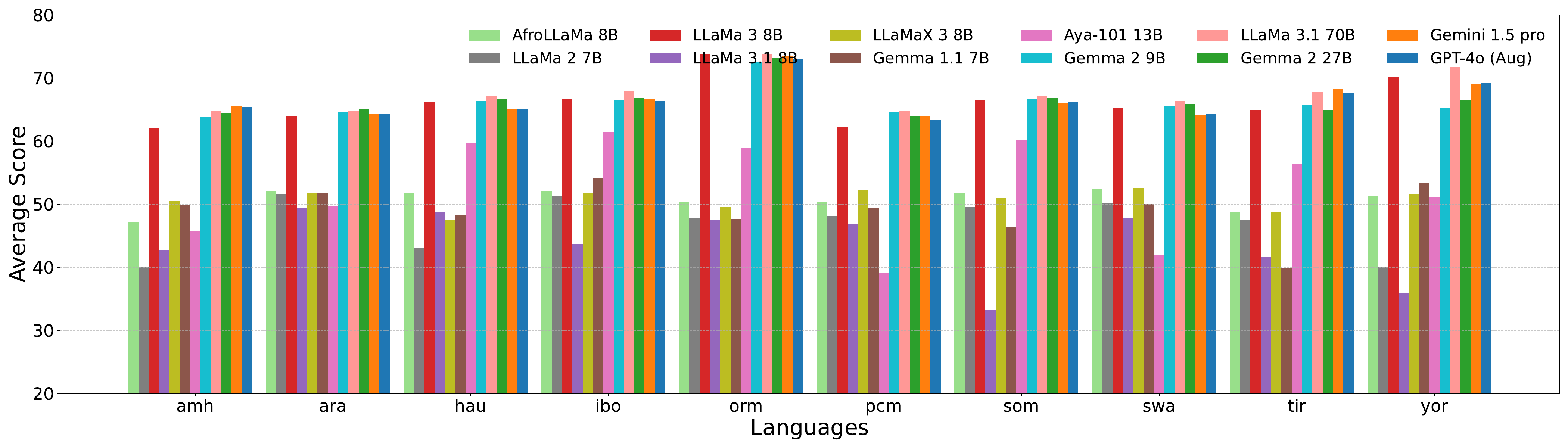}
     \vspace{-4mm}
    \caption{Per-language performance results for the \xlsum dataset.}
    \label{fig:xlsum_results}
     \vspace{-4mm}
\end{figure}

\subsubsection{Diacritics Restoration}

\textbf{\afriadr}

\begin{figure}[h]
    \centering
    \includegraphics[width=0.95\linewidth]{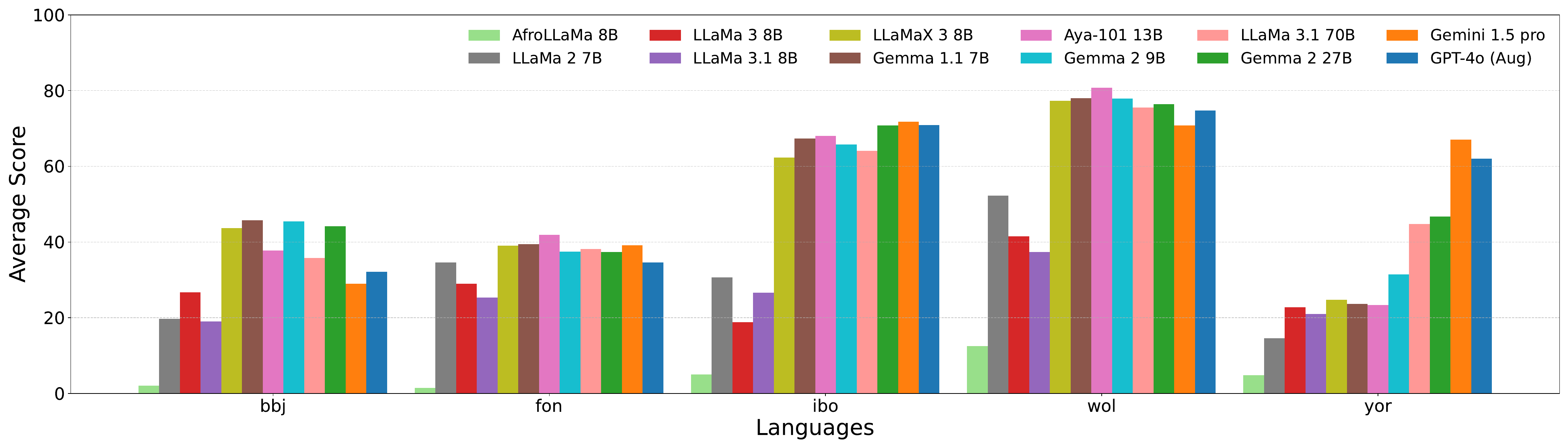}
     \vspace{-4mm}
    \caption{Per-language performance results for the \afriadr dataset.}
    \label{fig:adr_results}
     \vspace{-4mm}
\end{figure}

\end{document}